%% file: neurips_2026.tex
\documentclass{article}

\usepackage{xspace}

\usepackage[main, final]{neurips_2026}

\usepackage[utf8]{inputenc} 
\usepackage[T1]{fontenc}    
\usepackage{hyperref}       
\usepackage{url}            
\usepackage{booktabs}       
\usepackage{amsfonts}       
\usepackage{nicefrac}       
\usepackage{microtype}      
\usepackage{xcolor}         
\usepackage{graphicx}
\usepackage{subcaption}
\usepackage{enumitem}
\input{macros.tex}

\title{\tool: an alphabet of protein conformational ensembles}

\author{
  \textbf{Kaiwen Shi}$^{1,2}$ \qquad
  \textbf{Carlos Oliver}$^{1,2,3}$ \\
  $^{1}$Department of Computer Science, Vanderbilt University \\
  $^{2}$Center for AI in Protein Dynamics, Vanderbilt University \\
  $^{3}$Department of Molecular Physiology and Biophysics, Vanderbilt University \\
  \texttt{kaiwen.shi@vanderbilt.edu} \\
  \texttt{carlos.oliver@vanderbilt.edu}
}

\begin{document}

\maketitle

\begin{abstract}
Protein structure tokenizers (PSTs) are workhorses in protein language modeling, function prediction, and evolutionary analysis.
However, existing PSTs encode local geometry of \emph{static} structures and miss the correlated motions and alternative conformational states revealed by protein \emph{ensembles}.
Here we introduce \tool, the first learned tokenizer that maps a conformational ensemble to a single token per residue.
\tool addresses challenges inherent to tokenizing dynamics: deriving informative geometric descriptors across conformations, permutation-invariant encoding of variable-size ensembles, and conquering sparsity in dynamics data.
Trained with a Residual VQ-VAE using a frame distillation objective on a large molecular dynamics corpus,
\tool outperforms all tokenizer baselines on RMSF prediction and leads a token-conditioned ANOVA test on per-residue motion amplitude.
On EC, GO, binding site/affinity prediction and zero-shot mutation effect prediction \tool is competitive with existing static tokenizers despite using far less pretraining data than ESM3 and AminoAseed.
The distillation objective enables \tool to predict dynamics tokens from a single predicted structure, alleviating dynamics data sparsity.
As the field moves from static structure prediction toward ensemble generation, \tool offers the discrete vocabulary needed to bring dynamics into protein language modeling. Our code is available at: \href{https://github.com/OliverLaboratory/Ensembits}{https://github.com/OliverLaboratory/Ensembits}.
\end{abstract}

\input{0_introduction}

\input{1_method}
\input{2_results}
\input{3_conclusion}

\section{Acknowledgment}
We thank  Benjamin P. Brown and Hassane Mchaourab for constructive discussions and feedback. We also thank Asif Solanki for technical support and Jason Wang for codebase verification.

\newpage
\bibliographystyle{plainnat}   
\bibliography{references}

\newpage
\input{appendix.tex}

\end{document}

%% file: macros.tex
\usepackage{amsmath,amsfonts,bm}
\usepackage{booktabs}
\usepackage{tabularx}
\usepackage[normalem]{ulem}
\usepackage{algorithm}
\usepackage{algpseudocode}  
\usepackage{xcolor}
\usepackage{multirow}

\def\eqref#1{equation~\ref{#1}}

\def\1{\bm{1}}

\DeclareMathAlphabet{\mathsfit}{\encodingdefault}{\sfdefault}{m}{sl}
\SetMathAlphabet{\mathsfit}{bold}{\encodingdefault}{\sfdefault}{bx}{n}

\DeclareMathOperator*{\argmax}{arg\,max}
\DeclareMathOperator*{\argmin}{arg\,min}

\newcommand{\tool}{$\textsc{Ensembits}$\xspace}

%% file: 0_introduction.tex
\section{Introduction}

The effort to decode the language of protein function has progressed steadily
from mapping DNA sequences to solving 3D structure, and is currently facing
the challenge of predicting and capturing conformational dynamics. 
The ensemble of conformations
explored by a protein interfaces directly with biological function: catalytic
loops in enzymes such as adenylate kinase open and close to gate substrate
access, and G-protein coupled receptors transit between inactive and active
states to relay extracellular signals. 
In each case, function is encoded not
in any single snapshot but in the distribution of conformations a protein
samples and the correlated motions that connect them. It stands to reason that
such dynamical behaviours form the protein language that is shaped by evolution, akin to DNA codes, can be exploited by ML tools such as protein language
models (PLMs) if properly captured.
Grasping such a code will lead to improved protein structure and function
prediction, search and alignment, and ultimately protein design.

In ML terms, the objects we allude to are tokens.
Tokens are the
substrate upon which PLMs learn to reason, and as our grasp of protein
behaviour improves, these tokens have improved in granularity and information
richness from simple amino-acid sequences to structure-aware alphabets~\citep{Foldseek, ESM3, GeoBPE}, and to all-atom tokenization which enables modeling beyond proteins~\citep{bio2token}. Each step has unlocked downstream possibilities: 3Di
tokens drive Foldseek's four to five orders of magnitude speedup in structural
search~\citep{Foldseek}; ESM3's joint sequence-structure-function
vocabulary enables generative design~\citep{ESM3}; and discrete structural
tokens have been used as drop-in inputs for function prediction~\citep{SaProt}, remote
homology detection and inverse folding~\citep{Foldseek,ProstT5}.
A more expressive alphabet lets downstream models reason about a richer slice
of protein biology.

Access to rich dynamics data is growing rapidly. Large-scale MD repositories
such as mdCATH~\citep{mdCATH}, ATLAS~\citep{ATLAS}, and
MISATO~\citep{misato} now provide thousands of all-atom trajectories
spanning diverse fold space, replicas, and temperatures, offering an
unprecedented supervisory signal for learning dynamical priors directly from
physics. Concurrently, the next generation of protein structure prediction is
focusing on ensemble prediction: BioEmu~\citep{bioEmu} emulates
equilibrium ensembles at scale, AFsample2~\citep{AFSample} perturbs
AlphaFold's MSA~\citep{AF} and dropouts to sample alternative conformations, and methods
such as AlphaFlow, ESMDiff and Distributional
Graphormer~\citep{AFFlow, ESMDiff, DiffG} recast structure prediction as a generative problem over
conformational distributions.
Together, these advances mean that ensemble
data will not remain the bottleneck for long; instead, the bottleneck will shift to \emph{representation}.

Efforts to represent high-dimensional ensemble data in forms digestible by transformer architectures have recently begun.
Portal et al.~\citep{distogramRep} and Guo et al.~\citep{misatorgcn} show enhanced function prediction through
graph-based representations using distograms from Boltz2~\citep{boltz2} and MD-derived residue correlations respectively.
PETIMOT~\citep{PETIMOT} predicts per-residue principal motion directions from a single structure and protein language model embeddings.
ProtProfileMD~\citep{ProtProfileMD} encodes distributions of structural states as
histograms built on 3Di tokens~\citep{Foldseek} (thus not a tokenizer) to enhance language models, but doing so over
an alphabet trained on static structure forfeits joint modeling of correlated
motion across residues.
Several concurrent works also bridge discrete structure tokens and
conformational ensembles.
DynamicsPLM~\citep{DynamicsPLM} tokenizes each conformer of a
generated ensemble with the Foldseek/SaProt vocabulary and fuses
per-residue token frequency weights into a dynamics-aware
embedding, gaining on EC, metal-ion binding and localisation tasks.
RocketSHP~\citep{RocketSHP} builds a per-residue structural
heterogeneity profile over MD frames and predicts RMSF and
residue correlations from sequence.
Liu et~al.~\citep{harnessPST} show that structural-synonym
substitutions within the ESM3 codebook generate ensembles whose
flexibility correlates with MD.
MD-LLM-1~\citep{MDLLM} tokenizes MD frames with FoldToken and
trains a language model over the resulting per-frame token
sequences.
Like ProtProfileMD, these methods apply a static, single-frame
alphabet to each conformer and aggregate token counts or train over
token sequences; none learns a codebook from multi-frame input.

We propose \tool, the first learned tokenizer that maps a conformational \emph{ensemble} to a single token per residue.
\tool maps an unordered set of structural frames (protein conformers) to a
discrete token through three coupled design choices: (i) SE(3)-invariant descriptors computed
across frames whose neighbour identities are themselves frame-dependent, so
contact formation and breakage across the ensemble enter the descriptors
directly; (ii) a permutation-invariant
set encoder so that variable-size ensembles map deterministically to a single
latent regardless of frame order;  and (iii) a single-frame-to-token distillation objective that pulls
the embedding of a sub-sampled ensemble---down to a single frame---toward that
of the full ensemble, enabling the tokenizer to be queried at test time even
when only a static structure is available.
The result is a residual-VQ codebook
over local dynamical motifs that can be plugged into any downstream PLM in
place of, or alongside, existing static structural tokens. 

Our \textbf{contributions} are:
\begin{itemize}[itemsep=2pt, topsep=2pt, parsep=0pt, partopsep=0pt]
    \item \textbf{Formulation, model, and training pipeline.}
    We formulate ensemble tokenization as a multiset-to-token
    compression problem, design SE(3)-invariant descriptors of
    protein frames, and train an end-to-end tokenizer of protein
    dynamics.
    \item \textbf{Single-frame-to-token distillation (SFTD).}
    We devise a training objective that aligns sub-ensemble
    embeddings with their full-ensemble counterparts, letting one
    tokenizer serve both full or partial ensembles and single static frames
    at inference time.
    \item \textbf{Empirical validation across dynamics, function
    and mutation.} We empirically validate that \textsc{Ensembits}
    leads RMSF prediction and leads ANOVA tests on token-dynamics correlation. On EC, GO,
    binding site/affinity prediction and zero-shot mutation effect prediction, \textsc{Ensembits} is competitive with existing static structural tokenizers despite using far less pretraining data than ESM3 and AminoAseed.
\end{itemize}

%% file: 1_method.tex
\section{Method}

\subsection{Problem formulation and Ensembits}
Let $\mathcal{P} = \{x^1, \dots, x^P\}$ be an unordered multiset of
$P$ frames of a single protein, where each frame
$x^p \in \mathbb{R}^{L \times d}$ collects the (per-atom) coordinates
of the $L$ residues in conformation $p$. Here $d = 3 A$ encodes the
$A$ atoms retained per residue ($A = 3$ here: N, C$\alpha$, C). We write $x^p_r \in \mathbb{R}^{d}$ for the stacked
atom coordinates of residue $r$ in frame $p$. An SE(3)-invariant
descriptor function
\[
    \Phi \;:\; \mathbb{R}^{L \times d} \times \{1, \dots, L\}
    \;\longrightarrow\; \mathbb{R}^{D_f},
    \qquad
    \Phi(x^p, r) \;=\; \mathbf{f}^p_r
\]
maps a frame and a residue index to a per-residue, per-frame
descriptor of dimension $D_f$. A \emph{dynamics tokenizer} is then a
function
\[
    T \;:\; \bigcup_{P \ge 1}
    \bigl(\mathbb{R}^{D_f}\bigr)^{P}
    \;\longrightarrow\;
    \mathcal{C} \;=\; \{C_1, \dots, C_M\}
\]
that sends the per-residue descriptor multiset
$\{\mathbf{f}^1_r, \dots, \mathbf{f}^P_r\}$ to a token in a codebook
$\mathcal{C}$ of $M$ entries, where each token $C_m$ carries an
embedding $q_{C_m} \in \mathbb{R}^{D}$. Crucially, $T$ is
$P$-agnostic: the union over $P \ge 1$ in the domain expresses that
it accepts a multiset of any positive cardinality.

\begin{figure}[htbp]
    \centering
    \includegraphics[width=0.9\linewidth]{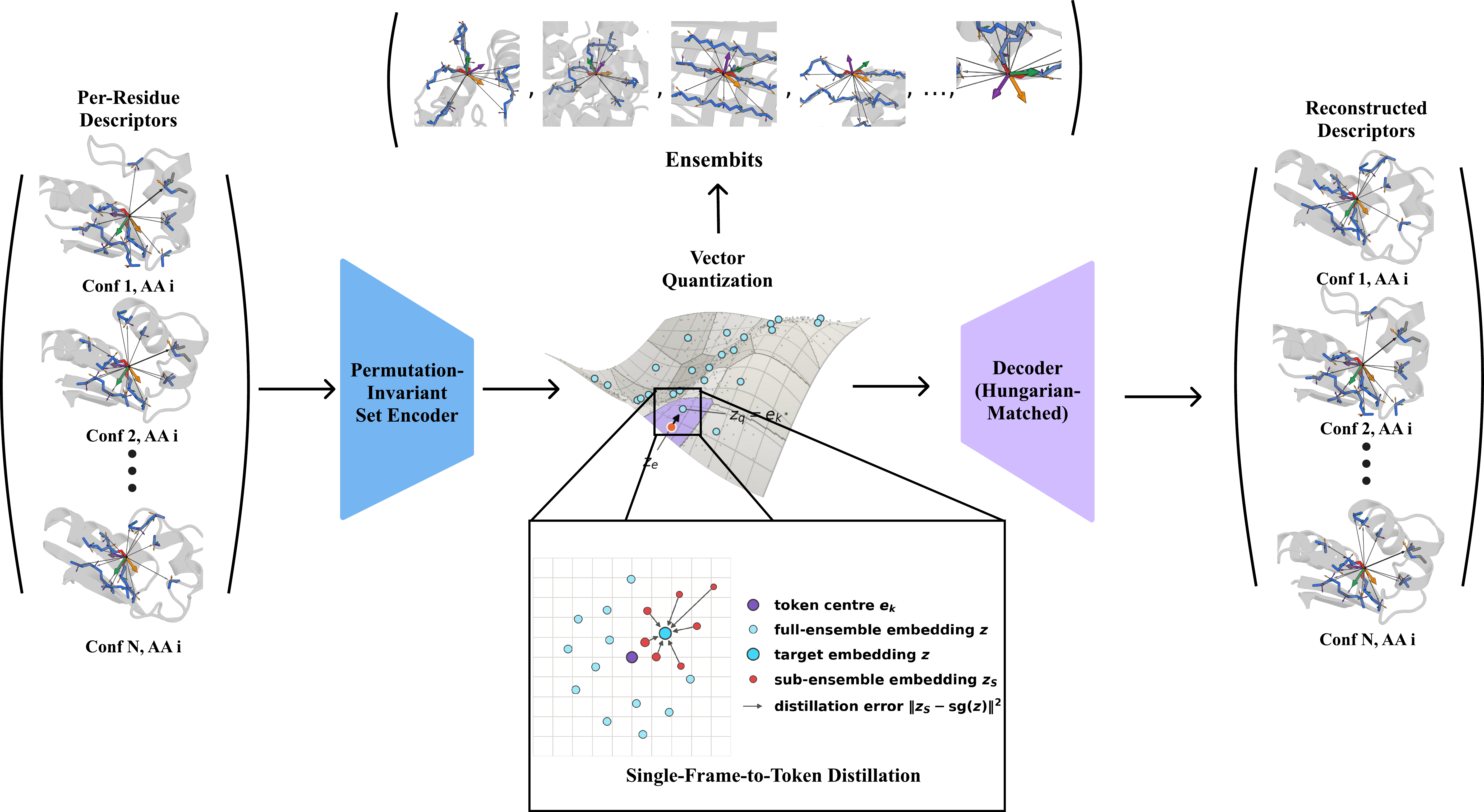}
    \caption{The Ensembits tokenization pipeline. The $P$-frame per-residue descriptor multiset is encoded by a permutation-invariant set encoder, residually quantized into a token tuple $(c_1, \dots, c_{N_q})$ with token embedding $q$, and decoded back to a fixed set of $P_{\max}$ descriptors under a Hungarian-matched FAPE reconstruction loss. Codebooks are updated by EMA. The box shows the mechanism of Single-Frame-to-Token Distillation: the embeddings of sub-ensembles are pulled towards the full-ensemble embedding.}
    \label{fig:arch}
\end{figure}

\tool realises this map with a Residual Vector-Quantized
Variational Autoencoder (RVQ-VAE)~\citep{RVQVAE,RQVAE}
(Figure~\ref{fig:arch}). The per-residue descriptor multiset
$\{\mathbf{f}^1_r, \dots, \mathbf{f}^P_r\}$ is passed through a
permutation-invariant set encoder, which produces a single latent
vector $z$. A residual vector quantizer with $N_q$ levels discretizes
$z$ into a tuple of codebook indices (the residue's dynamic
token) and outputs the corresponding quantized embedding $q$. A
decoder then maps $q$ back to $P$ descriptor vectors, supervised by a
permutation-invariant (Hungarian-matched~\citep{Hungarian}) reconstruction loss against
the input descriptor multiset: the decoder outputs a fixed set of $P_{\max} = 10$ descriptors and the loss is the Hungarian matched FAPE loss of \eqref{eq:fape}. Training jointly minimizes this
multiset-reconstruction loss and the standard VQ commitment loss.

\subsection{Descriptors}
\label{method:descriptors}
Ensembits captures protein dynamics by reconstructing per-residue
SE(3)-invariant descriptor multisets. Following the ESM3
construction~\citep{ESM3}, each residue is described by its
$K = 16$ nearest neighbors: each neighbor contributes its
$\mathrm{SE}(3)$ frame relative to the anchor's
$N/\mathrm{C\alpha}/C$ frame and its own backbone dihedrals
$\phi/\psi$ as sine and cosine, $256$ dimensions in total; \textbf{no sequence features are used}. The full
definition is in Appendix~\ref{app:descriptors}.

We compute the neighbors of each residue separately in every
frame (the \textbf{dynamical} mode), so neighbor identities can change
along the ensemble. This also enables the single-frame distillation of
Section~\ref{method:SFTD}. Two alternatives we considered, a fixed and a
fused neighbor set, called fixed and fused modes, are described in Appendix~\ref{app:descriptors-modes}
together with the reasons for this choice.

\paragraph{Dynamical mode.}
For each frame $p \in \{1,\dots,P\}$ and each residue
$r \in \{1,\dots,L\}$, we (i) compute residue $r$'s $K$ nearest
neighbors using the C$\alpha$ positions in frame $p$, obtaining a
frame-specific neighbor set
\[
\mathcal{N}^{p}_{r}
\;=\; \{n^{p}_{r,1}, \dots, n^{p}_{r,K}\}
\;\subseteq\; \{1,\dots,L\}, \qquad n^{p}_{r,1} = r,
\]
(residue $r$ is its own nearest neighbour, the self slot of Appendix~\ref{app:descriptor-esm3})
where $n^{p}_{r,i}$ is the index of residue $r$'s $i$-th nearest
neighbor in frame $p$ (sorted closest to farthest), and (ii) compute
the per-residue, per-frame feature vector
\[
\mathbf{f}^{p}_{r} \;=\; \Phi(x^{p}, r) \;\in\; \mathbb{R}^{D_f}
\]
from the per-pair feature block (Appendix~\ref{app:descriptors}). Each residue's descriptor is therefore
the multiset $\{\mathbf{f}^{1}_{r}, \dots, \mathbf{f}^{P}_{r}\}$ of $P$
per-frame feature vectors, which the set encoder ingests directly.
Because the neighbor identities $\mathcal{N}^{p}_{r}$ are allowed to
vary with $p$, the descriptor naturally encodes contact formation and
rupture across the ensemble. The per-frame dimension $D_f$ is
independent of $P$, so the same encoder ingests ensembles of any size,
which is critical for SFTD in Section~\ref{method:SFTD}.

\subsection{Tokenizer}
\tool relies on RVQ-VAE~\citep{RVQVAE} for quantization in the latent space. We detail the individual model components in the following section.

\paragraph{Set encoder.}
The set encoder maps the per-residue descriptor multiset
$\{\mathbf{f}^{1}_{r}, \dots, \mathbf{f}^{P}_{r}\} \subset \mathbb{R}^{D_f}$
to a single latent vector $z \in \mathbb{R}^{D}$. We adopt a
PerceiverIO-style design~\citep{PerceiverIO} that is
permutation-invariant by construction, removing any need to learn
invariance from data. A shared per-element MLP first embeds each
descriptor $\mathbf{f}^{p}_{r}$ independently, after which a small set
of learnable query tokens cross-attends to the $P$ embeddings, taking
a softmax-weighted average over them; the queries are then refined by
a stack of self-attention plus feed-forward blocks operating only
among themselves, and finally concatenated and linearly projected to
$z$. Permutation invariance follows because (i) the per-element MLP
is shared across descriptors, (ii) the softmax-weighted average in
cross-attention is invariant to the order of the key-value pairs, and
(iii) all subsequent operations act on the queries, whose ordering
is fixed by the architecture and independent of the input.

\paragraph{Residual quantizer.}
VQ-VAE~\citep{VQVAE} discretizes continuous embeddings via a learnable
neural net and has been applied in structural biology to tokenize protein
structures~\citep{Foldseek}. RVQ-VAE~\citep{RVQVAE} is a multistage
extension~\citep{VQVAEReview} in which a continuous embedding is
approximated by a sum of $N_q$ codewords drawn from $N_q$ independently
learned codebooks $\mathcal{C}_1, \dots, \mathcal{C}_{N_q}$. Given the
latent $z$ from the set encoder, the residual quantizer $Q$ outputs a token
index tuple $(c_1, \dots, c_{N_q})$ and the corresponding summed quantized
embedding
\[
q \;=\; \sum_{\ell=1}^{N_q} C^{\ell}_{c_\ell},
\qquad
c_\ell \;=\; \argmin_{i \in \{1, \dots, M_\ell\}} \bigl\lVert \rho_{\ell-1} - C^{\ell}_{i} \bigr\rVert,
\]
with the following terms:
\begin{itemize}
    \item $\mathcal{C}_\ell = \{C^{\ell}_{1}, \dots, C^{\ell}_{M_\ell}\} \subset \mathbb{R}^{D}$
    is the codebook of level $\ell$, a learned set of $M_\ell$ codewords
    in the latent space of the encoder.
    \item $c_\ell \in \{1, \dots, M_\ell\}$ is the codebook index selected
    at level $\ell$, the token of the residue at that level. Each $c_\ell$
    is the argmin of the distance between the current residual
    $\rho_{\ell-1}$ and the candidate codewords in $\mathcal{C}_\ell$.\footnote{The
    lowercase $c_\ell$ should be distinguished from the uppercase
    $C^{\ell}_{i}$, which denotes the $i$th \emph{codeword} (a continuous
    vector in $\mathbb{R}^{D}$), and from the calligraphic
    $\mathcal{C}_\ell$, which denotes the entire \emph{codebook} (the set of
    $M_\ell$ codewords at level $\ell$).}
    \item $C^{\ell}_{c_\ell} \in \mathbb{R}^{D}$ is the codeword at that
    index, the centroid embedding selected from $\mathcal{C}_\ell$ and added
    to $q$.
    \item $\rho_0 = z$ is the encoder latent, and
    $\rho_\ell = \rho_{\ell-1} - C^{\ell}_{c_\ell}$ is the residual at level
    $\ell$, the part of $z$ not yet explained by levels $1, \dots, \ell$.
\end{itemize}
We use $N_q = 3$ levels with $M_1 = 2048$ and $M_2 = M_3 = 128$, and
$D = 128$. To keep training stable, \tool uses EMA codebook
updates~\citep{VQVAE}, $k$-means codebook initialisation, dead code
revival~\citep{DeadRevival}, and a cosine learning rate
schedule~\citep{CosineAnnealing}. The full algorithm and training setup are
described in Appendix~\ref{app:RVQVAE}.

A practical concern with RVQ-VAE is that the total addressable codebook
grows multiplicatively: the token tuple $(c_1, \dots, c_{N_q})$ ranges over
$\prod_{\ell=1}^{N_q} M_\ell$ possibilities, which explodes with even two or
three levels when each $M_\ell$ is moderately large. In practice we found that training on all levels but using only the first
level tokens is sufficient for good downstream representations and works
better than a plain single codebook VQ-VAE with the same primary vocabulary
size $M_1 = 2048$ (Appendix~\ref{app:ablation-vq}). We suspect that the
later levels absorb fine grained, residue specific variation, freeing
$\mathcal{C}_1$ to encode the dominant dynamical mode in a denoised form.


\paragraph{Decoder and reconstruction loss.}
The decoder maps $q$ to a fixed set of $P_{\max} = 10$ descriptors
$\hat{F}_r = \{\hat{\mathbf{f}}^{1}_{r}, \dots, \hat{\mathbf{f}}^{P_{\max}}_{r}\}$.
We compare a decoded and a target descriptor with FAPE~\citep{AF}
on their $K$ relative frames $T$ (three pseudo atoms per frame, clamped at
$10$~\AA) plus a term on their dihedral values $\mathbf{d}$,
\[
\ell(\hat{\mathbf{f}}, \mathbf{f}) \;=\;
\mathrm{FAPE}(\hat{\mathbf{T}}, \mathbf{T})
\;+\; \frac{w_{\mathrm{dih}}}{4K} \bigl\lVert \hat{\mathbf{d}} - \mathbf{d} \bigr\rVert^{2}, \qquad f  = (T,d),
\qquad w_{\mathrm{dih}} = 3.
\]
Since frames have no order, we match the $P' \le P_{\max}$ encoded
descriptors $F_r = \{\mathbf{f}^{1}_{r}, \dots, \mathbf{f}^{P'}_{r}\}$ to
the decoder outputs with a Hungarian assignment~\citep{Hungarian} on this
same loss,
\begin{equation}
\sigma^{\star} \;=\; \argmin_{\sigma : \{1,\dots,P'\} \hookrightarrow \{1,\dots,P_{\max}\}}
\;\sum_{p=1}^{P'} \ell\bigl(\hat{\mathbf{f}}^{\sigma(p)}_{r}, \mathbf{f}^{p}_{r}\bigr),
\label{eq:hungarian}
\end{equation}
held fixed during backpropagation; unmatched outputs receive no loss. The
reconstruction loss is the matched loss,
\begin{equation}
\mathcal{L}_{\mathrm{recon}}\bigl(\hat{F}_r, F_r\bigr) \;=\;
\frac{1}{P'} \sum_{p=1}^{P'} \ell\bigl(\hat{\mathbf{f}}^{\sigma^{\star}(p)}_{r}, \mathbf{f}^{p}_{r}\bigr),
\label{eq:fape}
\end{equation}
averaged over all residues.

\subsection{Single-frame-to-token distillation}
\label{method:SFTD}
Next, we treat the question of how to handle situations where
conformation frames are not available. 
\textit{Can we still recover the
information learned from multi-conformation dynamics by just looking
at one frame or a few frames?} To address that, we propose \textbf{single-frame-to-token
distillation (SFTD)}, which biases single-frame inputs towards their
multi-frame counterparts.

Since the \textbf{dynamical} mode (Section~\ref{method:descriptors})
admits an arbitrary number of frames in arbitrary order, in each
training step we construct two ensembles $\mathcal{P}_1$ and
$\mathcal{P}_2$ from the same protein, where $\mathcal{P}_1$ contains
all $P_{\mathrm{max}}$ available frames and
$\mathcal{P}_2 \subseteq \mathcal{P}_1$ is a randomly sampled
sub-multiset of size
$|\mathcal{P}_2| = P^{\prime} \in \{1, \dots, P_{\mathrm{max}}\}$. The
two ensembles are encoded by the same set encoder to produce latent
vectors $z_1$ and $z_2$, respectively. We pull $z_2$ towards $z_1$
under a stop-gradient on the full-ensemble side:
\begin{equation}
\mathcal{L}_{\mathrm{total}}
\;=\;
\mathcal{L}_{\mathrm{recon}}
\;+\;
\beta \cdot \mathcal{L}_{\mathrm{commit}}
\;+\;
\frac{\lambda}{D} \cdot \bigl\lVert z_2 - \mathrm{sg}[z_1] \bigr\rVert_{2}^{2},
\label{eq:sftd}
\end{equation}
where $\mathrm{sg}[\cdot]$ is the stop-gradient operator,
$\mathcal{L}_{\mathrm{recon}}$ and $\mathcal{L}_{\mathrm{commit}}$ are
the mean of the per-branch RVQ-VAE reconstruction and commitment
losses across the two forward passes, $D$ is the latent
dimension, and $\beta = 0.5$, $\lambda = 10$ are
hyper-parameters. The stop-gradient ensures that $z_1$ acts as a
fixed teacher target while only the sub-ensemble branch $z_2$ receives
the distillation gradient. The full model has approximately $3.6$M parameters.
The effect of $\lambda$ is ablated in Appendix~\ref{app:ablation-sftd}; compute requirements are in Appendix~\ref{sec:compute}.

%% file: 2_results.tex
\section{\tool Validation \& Results}

\subsection{Datasets}
We train \tool on a combined MD corpus that fuses two all atom sources:
\textbf{BioEmu-CATH}, the MD dataset of CATH domains released with
BioEmu~\citep{bioEmu} ($1{,}057$ domains simulated at $300$~K, each in
many independent runs), and \textbf{MISATO}~\citep{misato}, an MD dataset
of $16{,}972$ protein--ligand complexes derived from PDBbind. For every
system we pool the frames of all runs and select $P_{\max} = 10$ diverse
frames by farthest point sampling on C$\alpha$ RMSD after optimal
superposition to maximize structural heterogeneity, and we split the domains and complexes $80/10/10$
into training, validation and test sets for the tokenizer. We evaluate
flexibility on mdCATH-div, a homology deduplicated derivative of
mdCATH~\citep{mdCATH} split at the CATH H superfamily level, and on
BioEmu-CATH and MISATO (Section~\ref{result:rmsf}), and remove from BioEmu-CATH
training every domain in an mdCATH-div test set. We defer dataset curation details to
Appendix~\ref{app:dataset}.

When MD is unavailable, one can use generative models such as
BioEmu~\citep{bioEmu} or CFRandom~\citep{CFRandom} to synthesize multiple
frames per protein and treat them as a surrogate ensemble. The quality of the learned tokens then becomes a
function of the generation quality. All reported results use MD only.

\subsection{Inspecting the learned alphabet}
\label{val:pics}

To test whether individual codebook entries capture coherent local
structural motifs, we visualize a few representative tokens by
selecting three distinct-protein exemplar residues per token and
overlaying them in a common reference frame. The detailed alignment
algorithms for exemplar selection and additional token representatives
can be found in Appendix~\ref{app:viz}, including the t-SNE analysis of
the codebook.

\begin{figure}[htbp]
    \centering
    \includegraphics[width=1\linewidth]{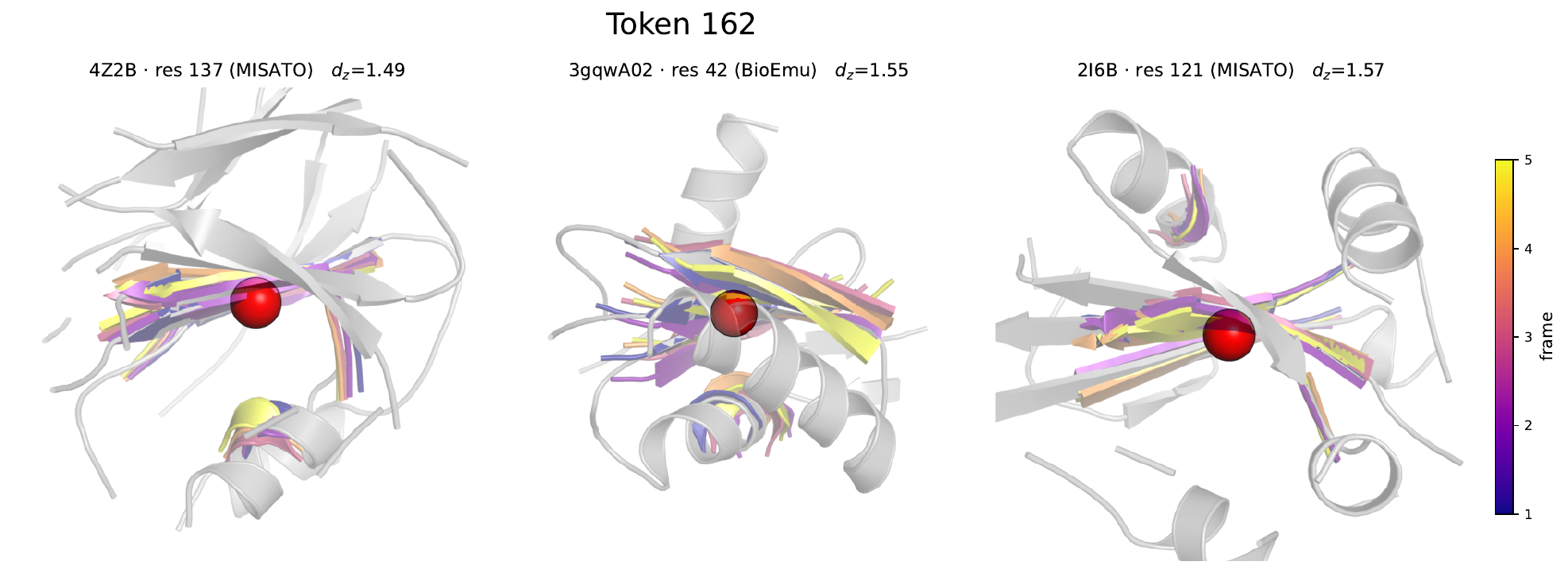}
    \caption{Token 162 ($\lambda = 10$ model, BioEmu-CATH + MISATO training corpus):
    a near-stationary local motif. Usage: 7{,}018 residues.
    Three distinct-protein exemplars ranked left-to-right by encoder-latent
    distance $d_z$ to the codebook centroid.
    Each panel overlays five descriptor frames (plasma colourmap)
    Kabsch-aligned on all C$\alpha$ atoms with the 3-mer residues of the
    16 canonical neighbours excluded.
    Mean motion amplitude $\langle s_1 \rangle = 0.75$, below the code median.}
    \label{fig:token1699}
\end{figure}

Token 162 (Fig.~\ref{fig:token1699}) illustrates a stationary motif:
the three highlighted regions superimpose tightly both across the five
frames within each panel and across the three protein exemplars,
evidence that this token represents a rigid, well-defined local
conformation. The exemplars are drawn from both training sources
(BioEmu-CATH and MISATO), demonstrating that the codebook generalizes the
same geometric pattern across heterogeneous corpora.

\begin{figure}[htbp]
    \centering
    \includegraphics[width=1\linewidth]{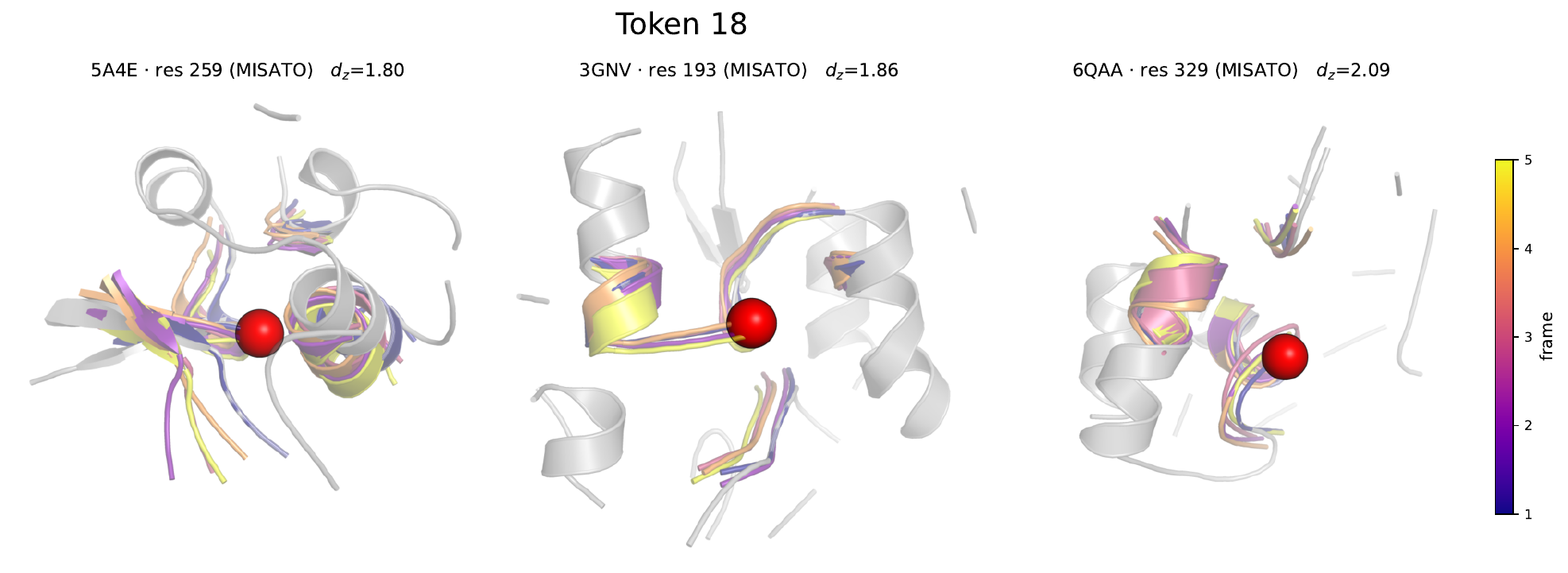}
    \caption{Token 18: a flexible local motif.
    Three exemplars ranked by increasing $d_z$. Same alignment and
    colouring as Fig.~\ref{fig:token1699}; mean motion amplitude
    $\langle s_1 \rangle = 1.27$, above the upper quartile of the codes with at least $500$ training residues.}
    \label{fig:token1063}
\end{figure}

Token 18 (Fig.~\ref{fig:token1063}) illustrates the opposite regime:
across the five overlaid frames the highlighted regions fan
out visibly, showing that this token encodes a flexible local
fragment whose conformation varies substantially across the ensemble
while its 3D-neighbour pattern is preserved.

\subsection{Does \tool encode flexibility?}
\label{result:rmsf}
We compare \tool against a set of tokenizer baselines on the task of predicting Root Mean Square Fluctuation (RMSF), the root mean square deviation of each residue's C$\alpha$ from its mean position across frames.
Specifically, SFTD empowers us to test against single-frame tokenizers, since oftentimes multi-frame ensembles are not available. 
We describe the implementations of each baseline in Appendix~\ref{app:baselines}, and test whether our tokenizer captures the dynamics through an RMSF probe test.

We probe the \emph{flexibility} signal carried by each token vocabulary
on three datasets: mdCATH-div (CATH-H structure split, $P=10$ frames),
BioEmu-CATH (CATH-H split, $P=10$) and
MISATO (sequence/structure split, $P=8$ frames)~\citep{misato}. For every protein, the label is
the per-residue C$\alpha$ root-mean-square fluctuation (RMSF) over all
frames of its MD simulation, superposed on their mean structure (iterated to
convergence). The tokenizers read only a few of these frames, and single-frame
methods the first of them, so the probe tests how well each tokenizer predicts the
flexibility of the whole simulation from a small ensemble or a single structure;
all methods are scored on the same residues (Appendix~\ref{app:baselines}).
Single-frame features are obtained
as described in Appendix~\ref{app:baselines}, and the same per-residue MLP head is
fit on the pooled training residues of each method. Performance is measured by the
Spearman rank correlation between predicted and ground-truth RMSF on
held-out test residues (rank rather than linear correlation,
because the relation of interest is monotonic: the token based probes
predict one flexibility level per token), with training stopped early on the
validation residues, averaged over $10$ probe seeds
(Table~\ref{tab:rmsf}). Because proteins differ in their overall
flexibility, Table~\ref{tab:rmsf_per_protein} in Appendix~\ref{app:downstreams} also reports the Spearman correlation
within each test protein, averaged over proteins.

\tool leads the RMSF benchmark among tokenizers in every setting. Among
multi-frame methods, \tool, $P{=}\text{full}$ is best on every column,
beating the strongest baseline (ProtProfileMD\_K) by $+0.031$ on
mdCATH-div, $+0.070$ on BioEmu-CATH, $+0.135$ on MISATO-seq and
$+0.139$ on MISATO-struct.
Among single-frame methods, \tool, $P{=}1$ leads on all four datasets,
ahead of AminoAseed by $+0.106$ on mdCATH-div and $+0.135$ on BioEmu-CATH.
The gain from $P{=}1$ to $P{=}\text{full}$ ($+0.114$ on mdCATH-div,
$+0.137$ on BioEmu-CATH) quantifies the marginal value of feeding the full
ensemble through the permutation-invariant set encoder at inference,
yet the strength of \tool, $P{=}1$ on its own confirms that SFTD
transfers dynamics signal into single-frame tokenization.
All $56$ protein level comparisons behind Table~\ref{tab:rmsf} ($P{=}1$ against each single-frame
baseline and $P{=}\text{full}$ against each multi-frame baseline, on the four datasets, for the pooled
and the per-protein Spearman correlation) favour \tool with Holm $p = 5.6\times10^{-4}$, the smallest
value attainable with $10^5$ bootstrap samples (Appendix~\ref{app:significance}).


We also perform a \textbf{token-conditioned ANOVA test }(Appendix~\ref{app:token_anova}) in which \tool tokens explain $\textbf{3.5}\times$ more motion variance than the best single-frame tokenizer and $\textbf{3}\times$ more than the best multi-frame baseline (\textbf{held-out $R^2$ on $s_1$: $0.308$ vs AminoAseed $0.087$ and ProtProfileMD\_K10 $0.102$}), with the gain persisting after controlling for confounds ($0.224$ vs $0.033$).
A complementary fold-pair test (Appendix~\ref{sec:fold_pairs}) finds that \tool ($P{=}\text{full}$) is \textbf{the only tokenizer whose token agreement is significantly higher on fold-matched pairs with similar dynamics than on pairs with divergent dynamics} ($\Delta = 0.056$, $p = 0.004$).

\begin{table}[htbp]
\centering
\small
\caption{RMSF Spearman (pooled over test residues) across four datasets, mean $\pm$ sample std over 10 probe seeds. mdCATH-div and BioEmu-CATH use CATH-H splits with $P=10$ frames; MISATO is reported on both the sequence (30\% identity clustering) and structure (CATH-H) splits with $P=8$ frames. Within the single-frame and multi-frame blocks, the best method is \textbf{bold} and the second best \underline{underlined}.}
\label{tab:rmsf}
\begin{tabular}{lcccc}
\toprule
Model & mdCATH-div & BioEmu-CATH & MISATO (seq) & MISATO (struct) \\
\midrule
AA              & 0.157 $\pm$ 0.000 & 0.178 $\pm$ 0.002 & 0.183 $\pm$ 0.000 & 0.205 $\pm$ 0.001 \\
3Di             & 0.361 $\pm$ 0.000 & 0.366 $\pm$ 0.004 & 0.322 $\pm$ 0.001 & 0.343 $\pm$ 0.001 \\
AminoAseed      & \underline{0.433 $\pm$ 0.001} & \underline{0.439 $\pm$ 0.004} & \underline{0.435 $\pm$ 0.001} & \underline{0.460 $\pm$ 0.001} \\
ProToken        & 0.032 $\pm$ 0.001 & 0.020 $\pm$ 0.007 & 0.044 $\pm$ 0.001 & 0.053 $\pm$ 0.004 \\
ESM3struct      & 0.390 $\pm$ 0.001 & 0.409 $\pm$ 0.001 & 0.407 $\pm$ 0.001 & 0.429 $\pm$ 0.001 \\
\tool, $P{=}1$  & \textbf{0.539 $\pm$ 0.002} & \textbf{0.574 $\pm$ 0.001} & \textbf{0.513 $\pm$ 0.001} & \textbf{0.535 $\pm$ 0.001} \\
\midrule
ProtProfileMD\_K & \underline{0.622 $\pm$ 0.002} & \underline{0.641 $\pm$ 0.006} & \underline{0.382 $\pm$ 0.001} & \underline{0.401 $\pm$ 0.001} \\
Vote\_3Di        & 0.452 $\pm$ 0.000 & 0.461 $\pm$ 0.001 & 0.333 $\pm$ 0.001 & 0.353 $\pm$ 0.001 \\
\tool, $P{=}\text{full}$    & \textbf{0.653 $\pm$ 0.001} & \textbf{0.711 $\pm$ 0.002} & \textbf{0.517 $\pm$ 0.003} & \textbf{0.540 $\pm$ 0.001} \\
\bottomrule
\end{tabular}
\end{table}

\subsection{Does the \tool alphabet carry functional information?}
\label{sec:downstreams}
The preceding experiments validated that \tool tokens carry quantized dynamics information and that SFTD enables predicting dynamics from a single frame. In this section we test the \tool tokens on multiple downstream tasks to investigate whether learned dynamics can help with function prediction, even when the pretraining data is sparse. Not to our surprise, the functional signals are noisy and mixed; on some of the tasks, \tool is better, while on some others, \tool is either competitive with other tokenizers or simply worse. We include our understandings of the test results, and leave further interpretations to the readers.

\paragraph{Mutation Effect Prediction}
We evaluate zero-shot mutational fitness prediction on the
\mbox{\textsc{ProteinGym}} substitution benchmark~\citep{ProteinGym},
restricted to the $96$ of $217$ deep mutational scanning assays whose
wild-type sequence length is at most $200$ residues ($64$ are
Tsuboyama stability assays; $88$ distinct UniProt proteins~\citep{UniProt};
$399{,}094$ variants total).
For each assay we fold the wild-type and every unique mutant
full-length sequence with ESMFold~\citep{ESMFold}
(built on ESM2-3B) and tokenize each predicted structure
at $P{=}1$ from backbone atoms (per-method atoms are listed in
Appendix~\ref{app:baselines}).
We score each variant with the negative sum, across all $L$ residue
positions, of the $\ell_2$ distance between the wild-type and mutant
level-0 codewords:
$s = -\sum_{i=1}^{L}
\bigl\lVert C^1_{c^{\,\mathrm{wt}}_{1,i}}
- C^1_{c^{\,\mathrm{mut}}_{1,i}} \bigr\rVert_2$.
Because no structural tokenizer is expected to beat ESM2
on a benchmark designed for sequence-fitness scoring, we test whether
the token-distance signal carries complementary information.
We combine each tokenizer with ESM2-650M~\citep{ESMFold} via the within-assay blend
$\alpha \cdot \tilde{s}_{\text{tok}} + (1{-}\alpha) \cdot \tilde{s}_{\textsc{ESM2}}$, where $\tilde{s}$ is a score
standardised within each assay (zero mean, unit variance), $s_{\text{tok}}$ is the token score $s$ above and
$s_{\textsc{ESM2}}$ the ESM2 score; $\alpha$ is chosen by leave-one-UniProt-out cross-validation
(grid $[0,1]$, step $0.05$). Significance is assessed by per-UniProt Wilcoxon signed-rank tests with Holm correction.

\tool + ESM2 significantly outperforms ESM2 alone
($\Delta = {+}0.025$, $95\%$ CI $[{+}0.017, {+}0.034]$, Holm $p = 4.8 \times 10^{-6}$)
and the $-n_\mathrm{mut}$ + ESM2 control
($\Delta = {+}0.018$, CI $[{+}0.009, {+}0.029]$, Holm $p = 0.0009$),
confirming that the token signal is complementary to ESM2 and goes beyond mutation count.
Combined with ESM2, \tool ($0.509$) is on par with AminoAseed, ProToken and 3Di ($0.507$--$0.510$)
and below ESM3struct ($0.515$; $\Delta = {-}0.006$, Holm $p = 0.005$).
With AminoAseed and ESM3struct retrained on our corpus (Appendix~\ref{app:static_retrain}), the order alone is
ESM3struct ($0.358$), \tool ($0.347$), AminoAseed ($0.337$).
Results are summarized in Table~\ref{tab:proteingym_ensemble}.

\begin{table}[htbp]
\centering
\small
\caption{Zero-shot mutational fitness on \textsc{ProteinGym}
($96$ DMS assays, wild-type length $\le 200$).
Mean per-assay Spearman $\rho$;
brackets show the mean over UniProt entries (each protein weighted once).
One ESMFold-predicted structure per sequence; tokenized at $P{=}1$.
The ``+ ESM2'' column uses a within-assay $z$-score blend with
$\alpha$ chosen by leave-one-UniProt-out CV.
$\Delta$ vs ESM2 with $95\%$ bootstrap CI;
Holm-adjusted per-UniProt Wilcoxon $p$.
Bold: highest value; Holm $p < 0.05$ where tested.
The Holm correction uses a pre-specified family of $14$ comparisons (the $10$ original plus $4$ baseline-vs-ESM2 tests added after $\lambda$ selection).}
\label{tab:proteingym_ensemble}
\begin{tabular}{lcccc}
\toprule
Predictor & Alone & + ESM2 (CV $\alpha$) & $\Delta$ vs ESM2 [95\% CI] & Holm $p$ \\
\midrule
$-n_\mathrm{mut}$ & 0.243 [0.257] & 0.491 [0.493] & $+$0.006 [$-$0.001, $+$0.015] & --- \\
3Di        & 0.309 [0.315] & 0.507 [0.511] & $+$0.023 [$+$0.015, $+$0.032] & \textbf{3.4e-05} \\
AminoAseed & 0.348 [0.359] & 0.510 [0.514] & $+$0.026 [$+$0.017, $+$0.035] & \textbf{5.3e-06} \\
ProToken   & 0.358 [0.369] & 0.508 [0.511] & $+$0.023 [$+$0.015, $+$0.032] & \textbf{1.7e-05} \\
ESM3struct & \textbf{0.384} [0.395] & \textbf{0.515} [0.520] & \textbf{$+$0.031} [$+$0.021, $+$0.042] & \textbf{5.2e-06} \\
\textbf{\tool} ($P{=}1$)
           & 0.347 [0.357] & 0.509 [0.513] & $+$0.025 [$+$0.017, $+$0.034] & \textbf{4.8e-06} \\
\midrule
ESM2-650M  & 0.484 [0.486] & --- & --- & --- \\
\bottomrule
\end{tabular}
\end{table}

\begin{figure}[htbp]
\centering
\includegraphics[width=\textwidth]{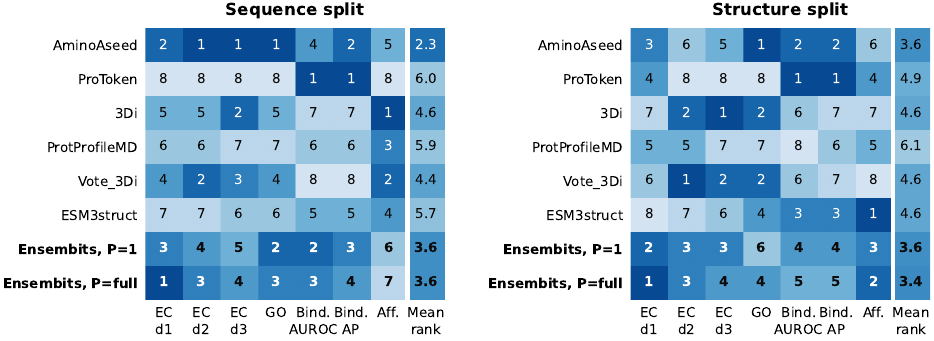}
\caption{Rank of each tokenizer setting on the MISATO function tasks (1 = best among the eight settings
shown; ties at printed precision share a rank) on the sequence (left) and structure (right) split, with
the mean rank over the seven metrics. Values are those of the function tables in
Appendix~\ref{app:downstreams}; amino acid identity and random tokens are reference rows there and are
not ranked.}
\label{fig:function_ranks}
\end{figure}

Figure~\ref{fig:function_ranks} ranks the eight tokenizer settings on every function metric.
On the sequence split, AminoAseed has the best mean rank ($2.3$) and the best EC and GO mAP among
single-frame methods; \tool follows (mean rank $3.6$ for both $P{=}1$ and $P{=}\text{full}$), with
$P{=}\text{full}$ first on EC depth~1 and $P{=}1$ second among single-frame methods on EC depths 1 and 2,
GO and binding AUROC. ProToken leads binding site prediction but ranks last on EC and GO.
On the structure split the methods are close: \tool, $P{=}\text{full}$ has the best mean rank ($3.4$) and
both \tool settings lead EC depth~1 among the tokenizers, 3Di leads EC depths 2 and 3 among single-frame
methods, AminoAseed leads GO and ProToken leads binding; the protein level tests find no significant
difference except ProToken over \tool on binding (Appendix~\ref{app:significance}).
On affinity, a model on the ligand fingerprint (MACCS) alone reaches Spearman $0.443$ (sequence) and
$0.360$ (structure) while protein features alone stay at or below $0.13$, so affinity is primarily a
ligand driven task (Appendix~\ref{app:downstreams}).

\begin{figure}[htbp]
\centering
\includegraphics[width=\textwidth]{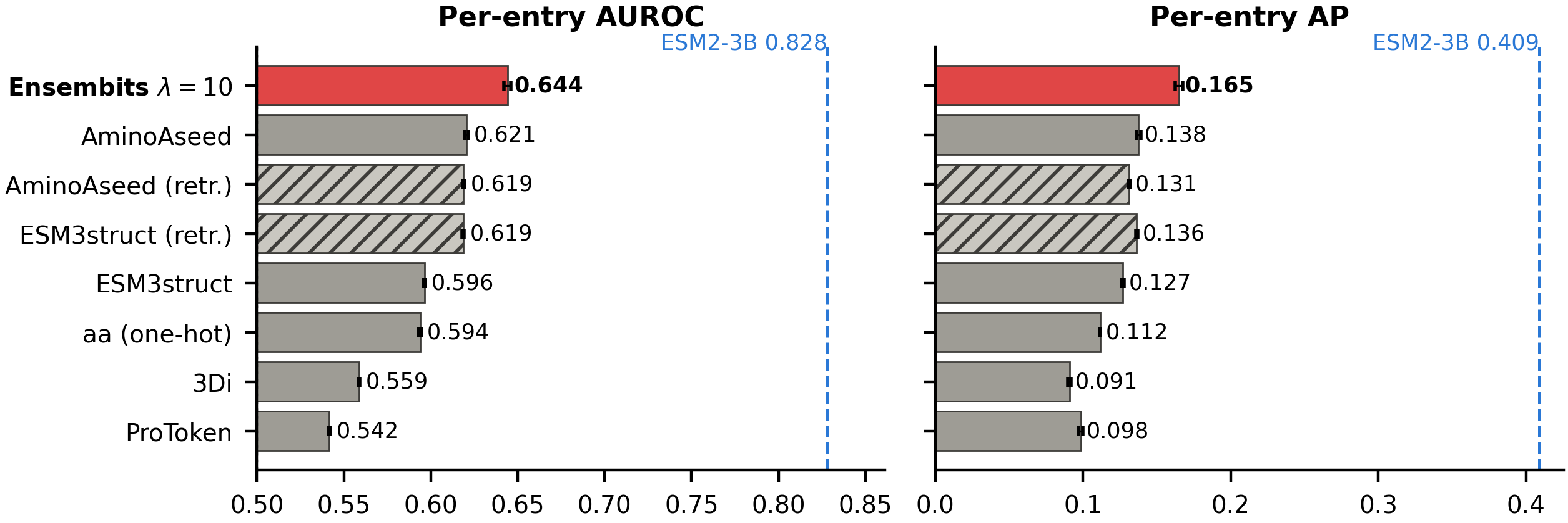}
\caption{CryptoBench MLP probe on the $222$ apo test entries: per-entry AUROC (left) and AP (right),
mean $\pm$ sample std over $5$ probe seeds. Hatched: AminoAseed and ESM3struct retrained on our corpus
(Appendix~\ref{app:static_retrain}). Dashed: ESM2-3B embeddings (continuous, descriptive only);
dotted: chance AUROC.}
\label{fig:cryptobench}
\end{figure}

We also test the tokenizers on cryptic pocket prediction with CryptoBench~\citep{CryptoBench}
(Appendix~\ref{sec:cryptobench}, Figure~\ref{fig:cryptobench}).
This benchmark matches what SFTD is designed to do. A cryptic pocket is absent from the apo
structure and appears only in a holo structure after a conformational change, so the pocket
residues cannot be read off the apo geometry alone. Every tokenizer sees only the single apo
structure, while the labels come from the holo partners. Because of SFTD, an \tool token
computed from the apo frame should carry information about the conformations that residue
can reach, which static tokenizers are not trained to capture. The results support this:
\tool tokens carry more cryptic pocket signal than any static tokenizer, published or
retrained on our corpus (per-entry AUROC $0.644$ vs $0.621$ for AminoAseed).

%% file: 3_conclusion.tex
\section{Discussion}
In this work we formalized the problem of protein-dynamics tokenization,
proposed a concrete architecture for solving it, verified that the learned tokens
faithfully capture ensemble-level dynamics, and demonstrated that the
quantized representation transfers to downstream functional tasks.
The main contribution is the algorithmic framework that allows us to tokenize
protein dynamics; optimization of the pipeline is left as future work.
\paragraph{Limitations.}
Token quality is bounded by the coverage and fidelity of the training ensembles.
The set encoder is permutation-invariant by design, so \tool applies to MD, PDB
and predicted ensembles without assuming temporal order; kinetic information
is outside the scope of these tokens.
We compare against every published structure tokenizer using their
released checkpoints; training data differs across methods.
Current
ensemble generators are either costly (MD) or not yet mature
enough (ML) to serve as a reliable source of augmentation.
\paragraph{Future directions.}
We highlight two directions. First,
given access to substantially larger and higher-quality ensemble
corpora, it would be valuable to scale the present architecture and
study how the token alphabet evolves with model and data size. Second,
because our tokens carry explicit dynamical information, they are a
natural input modality for protein language models and conditioning for
generative models (e.g.\ ESMDiff~\citep{ESMDiff}); we are especially
interested in whether augmenting PLMs with dynamics tokens yields
representations better suited to multi-frame structure generation.

%% file: appendix.tex
\appendix
\section{Descriptor specification}
\label{app:descriptors}

This appendix gives the full definition of the \tool descriptors
used as input to the RVQ-VAE. The descriptor follows the ESM3
relative-frame construction and is built from the backbone
$\mathrm{N/C}\alpha/\mathrm{C}$ atoms only; \textbf{it carries no sequence information.}

\subsection{Conventions}

Inputs are the per-frame coordinates $x^p \in \mathbb{R}^{L \times d}$
for $p \in \{1, \dots, P\}$, with $d = 3A$ encoding the $A$ atoms
retained per residue. The descriptor uses the real backbone
N, C$\alpha$ and C atoms of every structure in this paper. Where only C$\alpha$
is available, N and C would be reconstructed from consecutive C$\alpha$ positions
with ideal peptide geometry (canonical $|$N$-$CA$|=1.46$~\AA, $|$CA$-$C$|=1.52$~\AA, and the
N--CA--C angle $\approx 111^\circ$). We write
$\mathrm{C}\alpha_r$ for the C$\alpha$ position of residue $r$ in the
current frame.

Within a frame, the neighbours of residue $r$ are the residues
nearest to it in C$\alpha$ distance, with no minimum sequence
separation.

\subsection{ESM3-style relative-frame descriptor}
\label{app:descriptor-esm3}

The ESM3-style descriptor encodes a residue's local backbone
environment as the SE(3) transformations between its own backbone
frame and those of its $K$ spatial neighbours, following the
frame construction and neighbour search of ESM3~\citep{ESM3}
(no learnable parameters inside the descriptor itself).

\paragraph{Per-residue local frame.}
Given real backbone atoms $(\mathrm{N}_r, \mathrm{C}\alpha_r, \mathrm{C}_r)$
in frame $p$, we build a per-residue local frame
$T^{p}_{r} \in \mathrm{SE}(3)$ with origin $\mathrm{C}\alpha_r$ by
Gram--Schmidt orthonormalisation: the first axis points along
$\mathrm{C}\alpha_r - \mathrm{C}_r$, and the second axis is taken from
$\mathrm{N}_r - \mathrm{C}\alpha_r$ after removing its component along the first axis as in ESM3.

\paragraph{Per-neighbour block ($K \cdot 16$D).}
We select the $K = 16$ nearest residues to $r$ in C$\alpha$ distance via
the ESM3 neighbour search (with no sequence-gap filter, so
chain-adjacent residues are included), sorted from closest to farthest.
The list keeps $r$ itself as its first entry (distance $0$), followed by its
$15$ nearest neighbours. We keep this self slot because of its dihedrals:
$r$'s relative frame to itself is the identity and carries no information,
but the slot also carries $r$'s own backbone dihedrals $\phi_r, \psi_r$
(below), the most direct signal of $r$'s local conformation. Each entry's
frame is expressed in $r$'s local coordinates:
\[
T^{p,\mathrm{rel}}_{r \to j} \;=\; (T^{p}_{r})^{-1} \circ T^{p}_{j},
\qquad j \in \mathcal{N}^{p}_{r},\; |\mathcal{N}^{p}_{r}| = K,
\]
and flattened to its $12$-dimensional \texttt{Affine3D.tensor}
representation (a $3\times 3$ rotation matrix plus a $3$-vector translation,
all in $r$'s frame). To each entry we append its own backbone dihedrals
$\phi_j = \angle(\mathrm{C}_{j-1}, \mathrm{N}_j, \mathrm{C}\alpha_j, \mathrm{C}_j)$
and
$\psi_j = \angle(\mathrm{N}_j, \mathrm{C}\alpha_j, \mathrm{C}_j, \mathrm{N}_{j+1})$,
gathered along the same kNN edges and encoded as sine/cosine pairs; the undefined $\phi$ of the first residue and
$\psi$ of the last are set to $(0, 0)$, and so are $\phi_i$ and
$\psi_{i-1}$ wherever residues $i-1$ and $i$ are not bonded
($|\mathrm{C}_{i-1} - \mathrm{N}_i| > 2.5$~\AA), as at the junctions
between the chains of a MISATO complex or at residues missing from a
structure. Backbones reconstructed with ideal geometry (Conventions) are
treated as fully connected, because their reconstructed C and N atoms do
not form peptide bonds. Each entry therefore contributes
\[
\mathbf{g}^{p}_{r,j} \;=\; \bigl[\,T^{p,\mathrm{rel}}_{r \to j},\;
\sin\phi^{p}_{j},\, \cos\phi^{p}_{j},\, \sin\psi^{p}_{j},\, \cos\psi^{p}_{j}\,\bigr]
\;\in\; \mathbb{R}^{16},
\]
and the slate of $K = 16$ entries is concatenated into a
$K \cdot 16 = 256$-dimensional per-residue, per-frame descriptor:
\[
\mathbf{f}^{p}_{r} \;=\; \bigl[\,\mathbf{g}^{p}_{r,n^{p}_{r,1}},\ \ldots,\
\mathbf{g}^{p}_{r,n^{p}_{r,K}}\,\bigr] \;\in\; \mathbb{R}^{256},
\qquad n^{p}_{r,1} = r.
\]
The descriptor is SE(3)-invariant by construction: a global rigid-body
motion of the entire structure left-multiplies every $T^{p}_{j}$ by the same
group element, which then cancels under the
$(T^{p}_{r})^{-1} \circ T^{p}_{j}$ composition, and the dihedrals are
intrinsic. Real backbone N/C$\alpha$/C are required; the only
exception is the ideal geometry reconstruction described under
Conventions.

\subsection{Neighbour selection by mode}
\label{app:descriptors-modes}

The three modes below describe how the slate of $K$ neighbour slots is populated, or equivalently, how to choose the nearest neighbors. 

\paragraph{Fixed.}
For each residue $r$, we select a reference frame
\[
p^{\star}(r) \;=\; \argmax_{p \in \{1, \dots, P\}} R_g^{\mathrm{loc}}(r, p),
\]
where $R_g^{\mathrm{loc}}(r, p)$ denotes the local radius of gyration
of a small backbone window centered at $r$ in frame $p$ ---
equivalently, the conformation in which the local environment is most
expanded. This gives us a permutation-invariant ordering. We compute
the $K$ nearest neighbours of $r$ in this reference frame, obtaining
the neighbour set
\[
\mathcal{N}_r \;=\; \{n_{r,1}, \dots, n_{r,K}\} \;\subseteq\; \{1, \dots, L\}, \qquad n_{r,1} = r,
\]
and hold this set fixed across all frames. The per-frame feature
vector $\mathbf{f}^{p}_{r} \in \mathbb{R}^{D_f}$ is then computed from
the per-pair feature block (relative frame and dihedrals)
between $r$ and $\mathcal{N}_r$ evaluated in frame $p$. The
dimension is independent of $P$.

\paragraph{Dynamical.}
Rather than anchoring to a single frame, we recompute the
$K$-nearest-neighbour set
\[
\mathcal{N}^{p}_{r} \;=\; \{n^{p}_{r,1}, \dots, n^{p}_{r,K}\}
\;\subseteq\; \{1, \dots, L\}, \qquad n^{p}_{r,1} = r
\]
independently in every frame $p$. The per-frame feature vector
$\mathbf{f}^{p}_{r}$ is computed between $r$ and $\mathcal{N}^{p}_{r}$,
so neighbour identities are themselves frame-dependent. The descriptor
therefore captures contact formation and breakage along the ensemble,
a signal that is invisible to fixed-neighbour variants. The dimension
again matches the fixed mode and is independent of $P$. Our
tokenizer uses this mode with $K = 16$ and no sequence-gap filter.

\paragraph{Fused.}
The fused mode retains every neighbour that appears in any frame. For
each residue $r$, we sort the $P$ frames in decreasing order of
$R_g^{\mathrm{loc}}(r, p)$, then concatenate the per-frame
$k$-nearest-neighbour lists in this canonical order to form a fused
neighbour list of length $P \cdot k$. Duplicate occurrences are kept,
so every residue receives a slate of fixed cardinality. Per-frame
feature vectors $\mathbf{f}^{p}_{r}$ are computed against this fused
list in every frame. This mode retains the maximum information
available from the ensemble at the cost of a $P$-dependent dimension
(linear in $P$), and won't permit SFTD.

\paragraph{Choice of mode.}
\tool uses the dynamical mode. Its descriptor dimension does not grow
with $P$, unlike the fused mode. It needs no reference frame or
canonical frame ordering, which the fixed and fused modes take from the
local radius of gyration. And because the descriptor of a frame depends
on that frame alone, any sub-ensemble, down to a single frame, yields
descriptors of the same form, which single-frame-to-token distillation
(SFTD) requires. It also records contacts that form and break along the
ensemble, which a fixed neighbour set cannot.

\subsection{Resulting dimensions}

For the descriptor with $K = 16$:

\begin{center}
\begin{tabular}{lc}
\toprule
Mode & $D_f$ \\
\midrule
Fixed or dynamical & $256$ (independent of $P$) \\
Fused (never used) & $256 \cdot P$ (e.g.\ $2560$ at $P = 10$) \\
\bottomrule
\end{tabular}
\end{center}

\newpage
\section{Model Details}
\label{app:model}

\subsection{RVQ-VAE}
\label{app:RVQVAE}

RVQ-VAE~\citep{RVQVAE} is a multistage extension~\citep{VQVAEReview} in
which a continuous embedding is approximated by a sum of $N_q$ codebook
entries, one drawn from each of $N_q$ independently learned codebooks
$\mathcal{C}_1,\dots,\mathcal{C}_{N_q}$, with nearest neighbour
quantization operators
\[
Q_\ell(\rho) \;=\; \argmin_{e \in \mathcal{C}_\ell} \lVert \rho - e \rVert.
\]
Given the latent $z$ produced by the set encoder, the quantized embedding
\begin{equation}
q \;=\; \sum_{\ell=1}^{N_q} Q_\ell(\rho_{\ell-1}),
\qquad
\rho_0 = z,\quad \rho_\ell = \rho_{\ell-1} - Q_\ell(\rho_{\ell-1}),
\label{eq:rvq-recurrence}
\end{equation}
is built up additively. Each $Q_\ell$ returns the entry of
$\mathcal{C}_\ell$ nearest to the residual $\rho_{\ell-1}$, and the
indices $(c_1,\dots,c_{N_q})$ collected along the way form the discrete
token tuple (Algorithm~\ref{alg:rvq}). The straight through estimator is
applied to the summed embedding $q$, so the encoder receives a single
reconstruction gradient through the bottleneck, while the codebooks are
updated separately (below). Each level adds a commitment term, and
\[
\mathcal{L}_{\mathrm{commit}} \;=\; \sum_{\ell=1}^{N_q}
\frac{1}{D} \bigl\lVert \rho_{\ell-1} - \mathrm{sg}\bigl[Q_\ell(\rho_{\ell-1})\bigr] \bigr\rVert_2^2 ,
\]
weighted by $\beta = 0.5$ in Eq.~\ref{eq:sftd}. We use $N_q = 3$ levels
with $M_1 = 2048$ and $M_2 = M_3 = 128$ codes, in a latent space of
dimension $D = 128$.

\begin{algorithm}[htbp]
\caption{Residual Vector Quantization}
\label{alg:rvq}
\begin{algorithmic}[1]
\Require $z$, the latent produced by the set encoder; codebooks
$\mathcal{C}_\ell$ for $\ell = 1,\dots,N_q$
\Ensure the quantized embedding $q$ and the token tuple $(c_1,\dots,c_{N_q})$
\State $q \gets \mathbf{0}$
\State $\rho \gets z$ \Comment{residual}
\For{$\ell = 1$ \textbf{to} $N_q$}
    \State $c_\ell \gets \argmin_{i} \lVert \rho - C^{\ell}_{i} \rVert$
    \State $q \gets q + C^{\ell}_{c_\ell}$
    \State $\rho \gets \rho - C^{\ell}_{c_\ell}$
\EndFor
\State \Return $q,\ (c_1,\dots,c_{N_q})$
\end{algorithmic}
\end{algorithm}

The total addressable codebook grows multiplicatively: with
$M_\ell = |\mathcal{C}_\ell|$ codes at level $\ell$, the token tuple
$(c_1,\dots,c_{N_q})$ ranges over $\prod_{\ell=1}^{N_q} M_\ell$
possibilities, which explodes with even two or three levels. In practice
we found that training on all levels but using only the first level codes
is sufficient for good downstream representations and works better than a
plain single codebook VQ-VAE with the same primary vocabulary size
$M_1 = 2048$ (Appendix~\ref{app:ablation-vq}). We suspect that the later
levels absorb fine grained, residue specific variation, freeing
$\mathcal{C}_1$ to encode the dominant dynamical mode in a denoised form,
something a single codebook cannot do.

We also evaluated the basis reparameterization of
AminoAseed~\citep{AminoAseed}, which builds each codeword from a fixed
orthogonal basis composed with a learnable projection. It was not effective in our setting: the RMSF correlation dropped. \textit{We did not re-evaluate the functional tasks with this reparameterization.}

Several training choices proved essential for stable convergence. We
update each codebook $\mathcal{C}_\ell$ independently by an exponential
moving average (EMA) with decay $\gamma = 0.99$~\citep{VQVAE}. For each
code $i \in \{1, \dots, M_\ell\}$,
\begin{equation}
N_i^{(t)} = \gamma\, N_i^{(t-1)} + (1-\gamma)\, n_i^{(t)}, \qquad
m_i^{(t)} = \gamma\, m_i^{(t-1)} + (1-\gamma)\!\!\sum_{\rho' \in \mathcal{A}_i^{(t)}} \!\! \rho', \qquad
e_i^{(t)} = \frac{m_i^{(t)}}{\tilde{N}_i^{(t)}},
\label{eq:ema}
\end{equation}
where $\mathcal{A}_i^{(t)}$ is the multiset of level $\ell$ inputs
$\rho'$ assigned to code $i$ at training step $t$,
$n_i^{(t)} = |\mathcal{A}_i^{(t)}|$, and $e_i^{(t)}$ is the updated
codeword. The count is Laplace smoothed,
$\tilde{N}_i = (N_i + \epsilon)\, \sum_j N_j \,/\, \bigl(\sum_j N_j + M_\ell\, \epsilon\bigr)$
with $\epsilon = 10^{-5}$, so that rarely used codes do not divide by
zero. EMA avoids the slow, gradient driven oscillations we observed when
the codebook was learned through an auxiliary loss, and lets the codebook
track the encoder distribution directly. Before training, every codebook
is initialised by $k$-means (20 Lloyd iterations) on $16{,}384$ sampled
training residues, level by level: the first codebook on the encoder
outputs and each later one on the residuals left by the levels before it.

To guard against codebook collapse, where a few codes capture all
assignments and the rest stop being updated, we apply a windowed
dead-code revival rule inspired by~\citep{DeadRevival}.
A code that receives no assignment over a window of full batches
($100$ batches in the first two epochs, $400$ afterwards) is reseeded
to a distinct, randomly chosen input of its level from the current batch
(an encoder output at the first level, a residual at later levels)
plus small Gaussian noise, and its EMA count is reset to one.
The window restarts after each check.
To prevent interference with convergence, no code is reseeded in the
last $50$ steps of an epoch, and the last partial batch of each epoch
is dropped from the window count.
No checkpoint in any run contains duplicate codes.
During training we also add
Gaussian noise with standard deviation $0.05$ to the encoder output $z$
before quantization.

We optimise with AdamW~\citep{adamW} (learning rate $10^{-3}$, weight decay $10^{-5}$,
batches of $4096$ residues), with a linear warmup over the first $1000$
steps followed by cosine annealing~\citep{CosineAnnealing} towards
$10^{-6}$ over a horizon of $1000$ epochs. Training stops when the
validation reconstruction loss has not improved for $40$ epochs, and we
keep the checkpoint with the lowest validation loss. Without the cosine
decay, late updates were large enough to perturb a converged codebook and
make utilization oscillate. With these choices we observed a stable loss
descent and high code utilization at every level
($2{,}043/2{,}048 = 99.76\%$ of the primary codes at $P = 1$ on the
validation set).
Further model details are in Appendix~\ref{app:model-impl}.

\subsection{Dataset}
\label{app:dataset}

\tool is trained on a combined corpus of two all atom MD datasets,
BioEmu-CATH and MISATO, processed by a single descriptor pipeline and a
single training recipe. A third dataset, mdCATH-div, is used for evaluation
only.

\paragraph{BioEmu-CATH.}
The MD dataset of CATH domains released with BioEmu~\citep{bioEmu}
contains $1{,}093$ systems in two deposits; $1{,}090$ have
trajectories and cover $1{,}074$ distinct domains, $16$ of which were
simulated in both deposits. Each system is simulated at $300$~K in
many independent runs. For every system we concatenate the trajectories of
all runs, subsample them uniformly to at most $2{,}000$ frames, and select
$10$ frames by farthest point sampling (FPS) on C$\alpha$ RMSD after
optimal superposition to maximize structural heterogeneity. We remove the $17$ domains that appear in an
mdCATH-div test set (below), which leaves $1{,}072$ ensembles of
$1{,}057$ distinct domains; for the $15$ remaining domains simulated in
both deposits, both simulations are kept as separate ensembles.

\paragraph{MISATO.}
MISATO~\citep{misato} contains MD of $16{,}972$ protein--ligand complexes
derived from PDBbind, each simulated for $10$~ns in explicit water after
semi-empirical QM refinement of the starting PDB structure. For every
complex we select $10$ of the $100$ saved frames by FPS on C$\alpha$ RMSD
after optimal superposition, as for BioEmu-CATH, starting from the first
frame. The $8$ frames used by the MISATO downstream tasks are
selected by the same procedure; since the selection is greedy and starts
from the same frame, they are a subset of these $10$: the first $8$ greedy picks.

\paragraph{Combined training corpus.}
The union holds $18{,}044$ ensembles of $18{,}029$ distinct domains
and complexes. We split the domains (BioEmu-CATH) and complexes (MISATO)
at random, $80/10/10$, into training, validation and test sets for the
tokenizer, so that both simulations of a domain fall into the same
set ($14{,}423$ / $1{,}802$ / $1{,}804$ domains and complexes; $5.73$M /
$0.68$M / $0.70$M residues). This split only monitors reconstruction during training;
generalization is assessed by the downstream tasks, which use their own
splits and own training.

\paragraph{mdCATH-div (evaluation).}
For the flexibility probe we use mdCATH-div, a homology deduplicated
derivative of the mdCATH all atom MD corpus~\citep{mdCATH}, which spans
about $5{,}400$ CATH domains, $5$ temperatures and $5$ replicas per
domain. Training a probe on every (domain, temperature, replica) ensemble
would leak homology between train and test proteins, since CATH groups many
sequence redundant domains under the same family. We therefore keep one
representative domain per H superfamily, the finest level of the CATH
classification, giving $2{,}442$ ensembles from $2{,}447$ H groups. For
every representative we take the $320$~K trajectories of all five
replicas, keep every $10$th saved frame, and select maximally diverse
subsets of $P_s \in \{3, 5, 10\}$ frames by FPS on C$\alpha$ RMSD. Splits are
$80/10/10$ at the H superfamily level ($1{,}953$ train / $244$ val / $245$
test). The $675$ domains in any mdCATH-div test set (sequence, structure or
random split) are excluded from BioEmu-CATH training as described above.
The removal is per domain, not per superfamily: $61$ of the $853$ BioEmu-CATH training
domains share a CATH H superfamily with an mdCATH-div test domain.
The RMSF label uses every saved $320$\,K frame of the five replicas, without
the stride.

\paragraph{BioEmu-CATH (evaluation).} The RMSF probe on BioEmu-CATH uses all
$1{,}090$ systems with trajectories, split $80/10/10$ by CATH H superfamily with seed
$42$, as for the mdCATH-div structure split ($884$ / $91$ / $99$ domains from $470$
superfamilies, taken from the CATH domain list, where a domain missing from the list
would form its own group; both simulations of a domain fall into the same set). The label is the
C$\alpha$ RMSF over every frame of every run. This split differs from the tokenizer's
split: $79$ of the $99$ test domains are in the tokenizer's training set, so this column
measures how well the tokens encode the flexibility of proteins whose trajectories the
tokenizer has seen, not generalization to new folds. Generalization is tested on the mdCATH-div eval dataset, which training never sees.

\paragraph{Augmentation.}
When MD is unavailable, one can use generative models such as
BioEmu~\citep{bioEmu} or CFRandom~\citep{CFRandom} to generate multiple
frames for the same protein and treat them as an ensemble; the quality of
the learned tokens then depends on the generation quality. All reported results use the MD corpus above only.

\subsection{Implementation details}
\label{app:model-impl}

We summarize the architecture and training hyperparameters of \tool,
trained on the BioEmu-CATH and MISATO corpus with the ESM3 style
descriptor ($K = 16$, $P_{\max} = 10$, $D_f = 256$).

\paragraph{Architecture.}
\begin{itemize}
    \item \textbf{Set encoder} (in the style of PerceiverIO): a shared two
    layer MLP (hidden size $256$, ReLU) embeds each descriptor, one cross
    attention block lets $n_q = 8$ learnable queries attend to the $P$
    embeddings, and a stack of $n_{\mathrm{enc}} = 4$ self attention plus
    FFN blocks (FFN hidden size $512$, GELU) operates on the queries. All
    attention layers use $h = 4$ heads. The queries are concatenated and
    linearly projected to a latent of dimension $D = 128$.
    \item \textbf{Residual quantizer}: $N_q = 3$ levels with codebook sizes
    $[M_1, M_2, M_3] = [2048, 128, 128]$, an addressable token space of
    $M_1 M_2 M_3 \approx 3.4\times 10^{7}$; only the first level tokens
    ($M_1 = 2048$) are used downstream. Codebooks are initialised by
    $k$-means and updated by EMA with decay $\gamma = 0.99$. Dead-code
    revival follows a windowed rule: a code that receives no assignment
    over a window of full batches ($100$ in the first two epochs, $400$
    afterwards) is reseeded to a randomly chosen input of its level from
    the current batch plus small Gaussian noise (see
    Appendix~\ref{app:RVQVAE} for the full rule). The EMA update and the
    reseeding use only the full-ensemble branch of SFTD (all $10$ frames);
    the sub-ensemble branch reads the codebooks without updating them.
    During training, Gaussian
    noise with standard deviation $0.05$ is added to the encoder output;
    it applies to every encoder output during training, so it also enters
    the SFTD target $z_1$, the commitment term and the samples used for the
    $k$-means initialisation, and it is off at inference.
    \item \textbf{Decoder}: an MLP with three hidden layers of size $256$
    and GELU activations, mapping $q \in \mathbb{R}^{128}$ to
    $P_{\max} = 10$ descriptor vectors.
    \item Total trainable parameters: $3.59$M.
\end{itemize}

\paragraph{Loss.}
The training objective is \eqref{eq:sftd},
$\mathcal{L}_{\mathrm{total}} = \mathcal{L}_{\mathrm{recon}}
+ \beta\,\mathcal{L}_{\mathrm{commit}}
+ \tfrac{\lambda}{D}\,\lVert z_2 - \mathrm{sg}[z_1] \rVert^{2}$,
with commitment weight $\beta = 0.5$ and SFTD weight $\lambda = 10$, where
$\mathcal{L}_{\mathrm{recon}}$ is the Hungarian matched loss of
\eqref{eq:fape} ($w_{\mathrm{dih}} = 3$, FAPE clamp $10$~\AA) averaged
over the two SFTD branches.

\paragraph{FAPE on relative frames.}
\label{app:fape}
A descriptor $\mathbf{f}$ holds $K$ relative frames
$g_j = (R_j, \mathbf{t}_j)$, $j = 1, \dots, K$, all expressed in the frame of
the central residue (the predicted rotations $\hat{R}_j$ are first made
orthonormal by Gram--Schmidt). FAPE compares two rigid motions by how
differently they move a fixed set of probe points. We use three probe points,
$O = \{(1.458, 0, 0),\ (0, 0, 0),\ (-0.545, 1.421, 0)\}$~\AA, and for every
ordered pair of frames $(k, l)$ we move $O$ by the predicted and by the true
transform from frame $l$ into frame $k$ and measure the displacement:
\[
\mathrm{FAPE}(\hat{\mathbf{f}}, \mathbf{f}) \;=\;
\frac{1}{3K^{2}} \sum_{k=1}^{K} \sum_{l=1}^{K} \sum_{\mathbf{o} \in O}
\min\Bigl(10~\text{\AA},\;
\bigl\lVert \hat{g}_k^{-1} \hat{g}_l\, \mathbf{o} \;-\; g_k^{-1} g_l\, \mathbf{o} \bigr\rVert\Bigr).
\]
This is the FAPE of \citet{AF} or at least its equivalent: there, the backbone atoms of residue $l$ are
placed by applying $g_l$ to fixed ideal offsets, and expressing them in frame
$k$ gives exactly $g_k^{-1} g_l\, \mathbf{o}$. Because the three probe points
are not collinear, a term vanishes only when the predicted and true relative
transforms agree. The probe points are otherwise arbitrary; they are AF2's
ideal N, C$\alpha$ and C offsets, which do not coincide with the real N and C
positions in the ESM3 frame used here, hence ``pseudo atoms''. Their spread
sets the weight of rotation errors relative to translation errors (a lever
arm of about $1.5$~\AA). As in AF2, the distance is not squared and is
clamped at $10$~\AA.

\begin{figure}[htbp]
    \centering
    \includegraphics[width=1\linewidth]{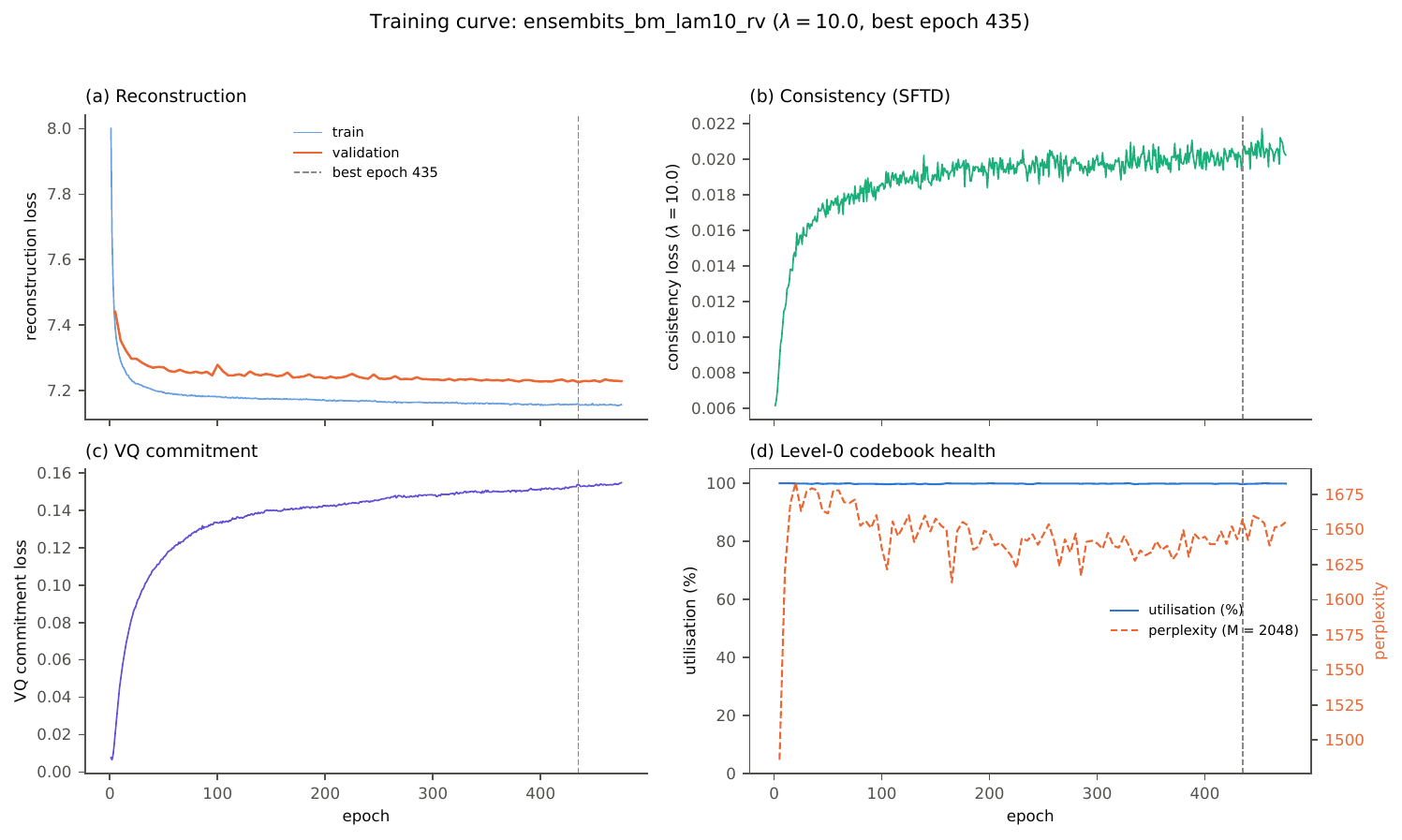}
    \caption{Training curve of the $\lambda = 10$ model.
    (a)~Reconstruction loss (train and validation);
    (b)~SFTD consistency loss;
    (c)~VQ commitment loss;
    (d)~level-0 codebook utilisation and perplexity ($M = 2048$).
    Best epoch: 435 (early stopping on validation reconstruction, patience
    40 epochs). 475 epochs trained.}
    \label{fig:training_curve}
\end{figure}

\paragraph{Optimization (Fig.~\ref{fig:training_curve}).}
AdamW~\citep{adamW} with initial learning rate $10^{-3}$, weight decay
$10^{-5}$, a $1000$-step linear warmup, and cosine annealing towards
$10^{-6}$ over a horizon of $1000$ epochs. Batch size $4096$ residues,
gradient clipping at norm $1.0$. Early stopping is evaluated every $5$
epochs; training stops when the validation reconstruction loss has not
decreased by more than $10^{-5}$ for $40$ epochs, and we keep the
checkpoint with the lowest validation loss.
The reported model ($\lambda = 10$) stopped at epoch $475$ with the
best checkpoint at epoch $435$.
Descriptors are used without standardization. In the
sub-ensemble branch of SFTD, the number of input frames is sampled at each
step as $P' \sim \mathcal{U}\{1, \dots, 10\}$, while the full-ensemble
branch always uses all $10$ frames, so the encoder sees every
$P \in \{1, \dots, 10\}$.

\paragraph{Hardware and runtime.}
Training was performed on a single NVIDIA H200 GPU.
The reported $\lambda = 10$ model trained for $24.1$ hours ($475$ epochs, best
at epoch $435$, over $5{,}734{,}822$ training residues / $684{,}485$ validation
residues).

\paragraph{$\lambda$ selection.}
$\lambda = 10$ was selected on the basis of validation RMSF on
mdCATH-div and validation reconstruction loss.

\paragraph{Codebook utilization.}
At convergence, the $M_1 = 2048$ primary codebook reaches
$2{,}043 / 2{,}048 = 99.8\%$ utilization on the validation set at
$P = 1$ (perplexity $\approx 1{,}640$; at $P = 10$: $2{,}042$ ids, perplexity $1{,}658$).
The codebook sizes were chosen to keep code utilization high at
the current data scale.

\newpage
\section{Token visualizations}
\label{app:viz}

\subsection{Tokens}
For each token, exemplars are picked by ranking all of its
assigned residues by Euclidean distance $d_z$ between the encoder's
pre-quantization latent and the token's codebook centroid; panels are
arranged left-to-right by increasing $d_z$ so the leftmost exemplar is
the most token-central and the rightmost the most peripheral. For each
exemplar residue (anchor), we identify the sixteen canonical $k$-nearest
neighbours (defined as the residues most frequently chosen as
descriptor neighbours across the $P = 10$ ensemble frames, matching
the $K = 16$ kNN that the ESM3-style descriptor uses at training time;
no sequence-gap filter is applied) and highlight the local 3-mer
$(j{-}1,\,j,\,j{+}1)$ around each. To reveal token-level dynamics
rather than global rigid-body motion, we Kabsch-align~\citep{kabsch}
all $P$ frames using \emph{all} C$_\alpha$ atoms of the protein with
the 3-mer residues excluded, applying the same rigid-body transform to
the all-atom coordinates from the source trajectory (BioEmu-CATH or MISATO MD; five of the ten descriptor frames).

\begin{figure}[htbp]
    \centering
    \includegraphics[width=1\linewidth]{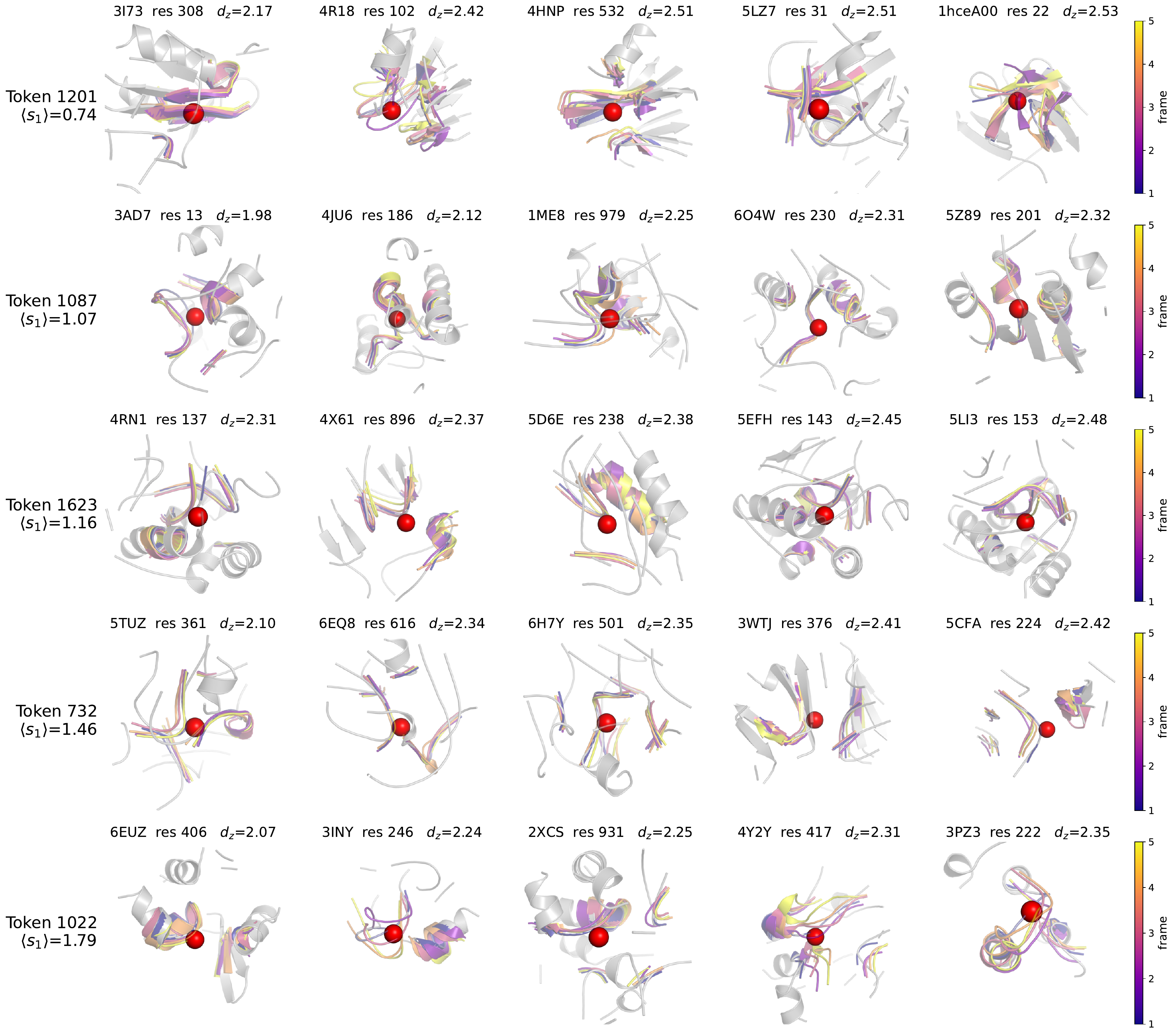}
    \caption{Five representative tokens from the codebook (rows),
    ordered top to bottom by increasing average residue motion
    amplitude $\langle s_1 \rangle$ (compact-stationary $\to$
    dynamical), and within each row by increasing encoder-latent
    distance $d_z$ to the codebook centroid (most $\to$ least
    token-central, left to right). Each cell shows one
    distinct-protein exemplar of the row's token: the anchor
    C$_\alpha$ (red sphere), its sixteen canonical kNN 3-mers
    (cartoons coloured by frame index along the \texttt{plasma}
    colormap, five of the ten descriptor frames from the source trajectory, BioEmu-CATH or MISATO MD), and the
    surrounding local backbone (gray, single frame). The same local
    geometry recurs consistently across proteins drawn from the training corpus; moving down the rows, the 3-mers spread out
    across the overlaid frames, indicating progressively larger
    conformational fluctuations encoded by the token.}
    \label{fig:5x5tokens}
\end{figure}

We here show representative tokens from the learned codebook, each
illustrated with five distinct-protein exemplars drawn from the
training split of the BioEmu-CATH\,+\,MISATO corpus on which the tokenizer was
trained. Rows are ordered from top to bottom by increasing average
per-residue motion amplitude
$\langle s_1 \rangle$\footnote{See Appendix~\ref{app:token_anova} for a
more detailed definition of $s_1$.}, the leading singular value of the
locally-aligned C$_\alpha$ frame matrix, so the top rows correspond to
compact, near-stationary local motifs and the bottom rows to more
dynamical motifs whose 3-mer fragments visibly fan out across the five
overlaid frames. Within each row, columns are ordered left to right by
increasing encoder-latent distance $d_z$ from the token's codebook
centroid, so the leftmost panel is the most ``token-central'' exemplar
and the rightmost the most peripheral. The same local geometry recurs
across unrelated proteins drawn from both source corpora: the anchor
(red sphere) sits at a consistent position relative to its sixteen
canonical $k$-NN 3-mers, demonstrating that the codebook captures
protein-agnostic structural motifs rather than protein-specific or
dataset-specific peculiarities. About half of the fifteen non-self neighbours fall within a sequence local window of $|i - j| \le 15$ ($41\%$ to $68\%$ per row of Fig.~\ref{fig:5x5tokens}); the others are long-range contacts from other secondary structure elements packed against the
anchor's local environment. Selection, alignment, and rendering follow
the procedure described above.

\newpage
\subsection{Codebook structure}
\label{sec:codebook_tsne}

\begin{figure}[htbp]
    \centering
    \includegraphics[width=1\linewidth]{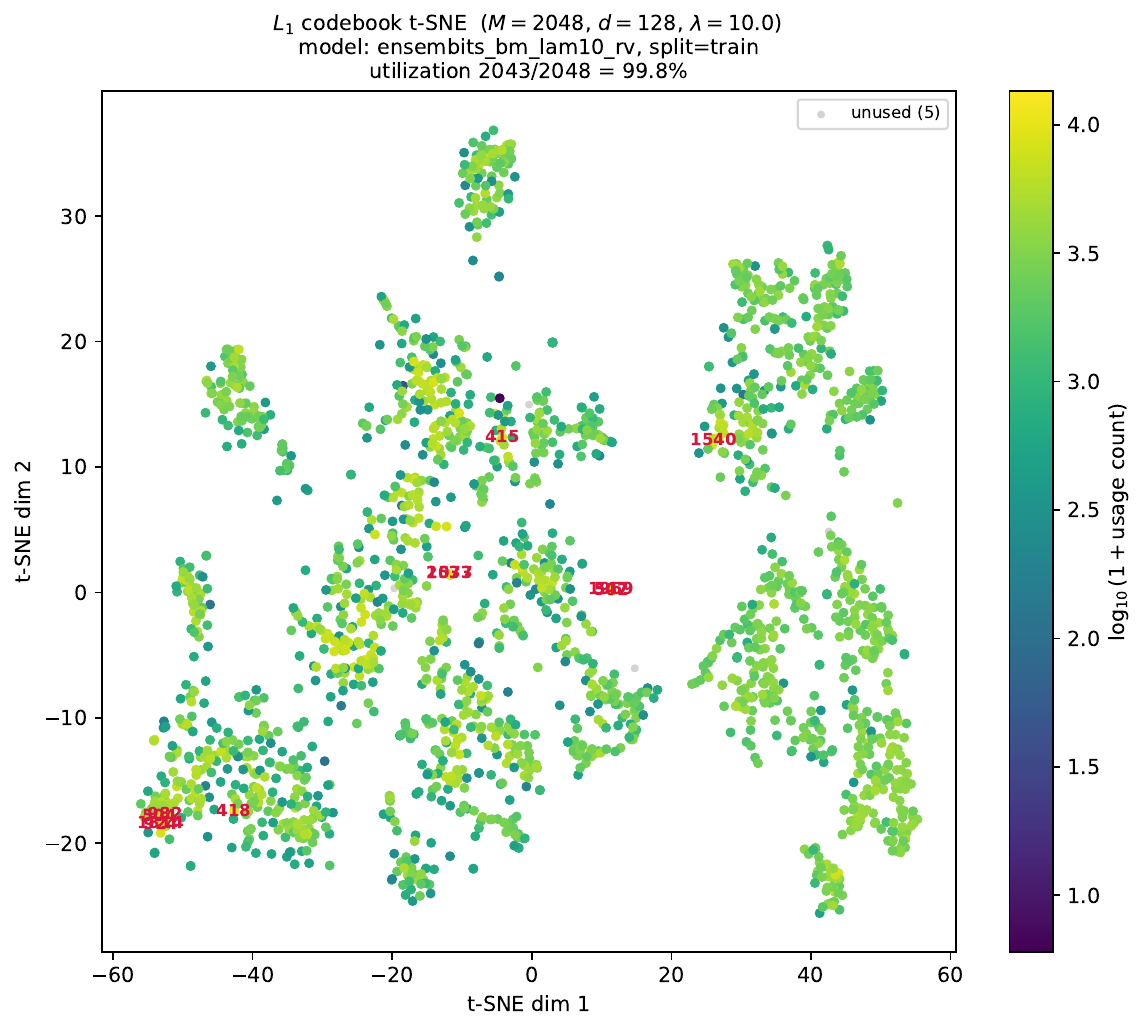}
    \caption{t-SNE projection of the $L_1$ primary codebook ($M = 2048$
    entries, $D = 128$) for the reported tokenizer (BioEmu-CATH + MISATO training split, $K = 16$, $P = 10$). Each point is one codebook entry;
    the color of used codes encodes
    $\log_{10}(1 + \mathrm{usage\ count})$ across the combined corpus,
    while the $5$ unused codes are drawn in light gray. The twelve
    most-used token IDs are annotated in red. Codebook utilization is
    $99.76\%$ with a near-uniform usage distribution (top-to-mean
    ratio $\approx 4.8\times$), and the t-SNE shows no separation
    between high- and low-usage codes --- the codebook tiles the latent
    manifold approximately uniformly.}
    \label{fig:codebook_tsne}
\end{figure}

To inspect the geometry of the learned token vocabulary, we project the
$L_1$ primary codebook ($M = 2048$ entries, each in $\mathbb{R}^{128}$)
to two dimensions via t-SNE on the cosine distance, and color each
point by how often the code is used as a per-residue token assignment
across the training split of the BioEmu-CATH\,+\,MISATO corpus ($5{,}734{,}822$ residue token assignments over $14{,}436$ ensembles: $866$ BioEmu-CATH systems and $13{,}570$ MISATO complexes); see Fig.~\ref{fig:codebook_tsne}.

Two observations are notable. First, the codebook achieves near-complete
utilization: $2{,}043 / 2{,}048 = 99.76\%$ of codes are assigned at least one
residue across the corpus, with no dead-code clusters --- the five unused codes (gray) appear as isolated points scattered among the live
ones, not as a connected region. This confirms that the EMA codebook
update combined with random-restart dead-code revival succeeds at
populating the entire vocabulary. Second, the per-code usage
distribution remains close to uniform: the most-used token receives
$13{,}515$ residues while the mean over used codes is $\approx 2{,}807$,
a top-to-mean ratio of only $4.8\times$ (and a p99/p50 ratio of
$3.11\times$). In typical VQ-VAE training, this ratio can exceed
$10\times$ when the codebook is over-capacity --- a few attractor
codes absorb most assignments and the rest atrophy. The fact that
high- and low-usage codes are visually intermixed in the t-SNE further
reinforces this: there is no pocket of attractor codes carving out a
high-density region. Together these properties indicate that the
codebook is neither collapsed nor over-allocated; it tiles the
encoder's latent manifold approximately uniformly, with each code
contributing meaningfully to the tokenization.

\newpage
\section{Baselines}
\label{app:baselines}
We describe the implementations of each baseline and \tool here. When
N/C atoms are unavailable in the source structure, missing backbone
atoms are reconstructed with ideal geometry (canonical $|$N$-$CA$|=1.46$~\AA,
$|$CA$-$C$|=1.52$~\AA, and the N--CA--C angle $\approx 111^\circ$). No reported result uses this reconstruction: every mdCATH-div, BioEmu-CATH and MISATO structure is read with its real backbone atoms. On ProteinGym, every method reads the ESMFold predicted structure (N/C$\alpha$/C, plus O for ProToken and C$\beta$ for 3Di).

\paragraph{Residues and inputs of the RMSF probe.} On each dataset every
method is trained, validated and tested on the same residues, fixed before any
probe is run: on mdCATH-div all $308{,}293$ residues of the $2{,}442$ domains, on BioEmu-CATH all residues of the $1{,}090$ systems, on MISATO all residues
of the $4{,}894$ complexes. A method that cannot featurize a residue of this set is
an error, not a smaller test set. Single-frame methods, \tool at $P{=}1$ included,
read the first frame of the ensemble: on mdCATH-div and BioEmu-CATH the frame where
farthest point sampling starts (drawn at random with a fixed seed from the pooled
frames of all replicas or runs), on MISATO the first saved MD frame. The RMSF label
is computed from all MD frames, of which the probe inputs are a small subset
($10$ on mdCATH-div and BioEmu-CATH, $8$ of $100$ on MISATO). Each baseline tokenizes one polypeptide chain at a time. A MISATO molecule can hold
several chains, and one chain can be split over several molecules at missing residues,
so on MISATO the chains are recovered from the deposited PDB entry: every backbone break
(C--N $>2.5$~\AA) and every change of molecule is located in the deposited structure and
cut only where its two sides lie in different chains, are farther apart there than the
missing residues can span, or cannot be located. A chain is thus kept whole across missing
residues, with residue indices following the SEQRES positions, as our descriptor also
reads it whole. mdCATH-div and BioEmu-CATH proteins (single chains without gaps) are encoded whole;
the joint multi-chain modes of ESM3 and ProToken are not used, and baseline features enter the probe without normalisation.

\paragraph{Single-frame methods:}
\begin{itemize}
    \item \textit{aa}: 21-dimensional one-hot encoding of the amino-acid
    identity at each residue (20 amino acids and one class for nonstandard
residues; force-field residue names, such as the CHARMM histidine names in mdCATH,
are mapped to their amino acid).
    \item \textit{3Di\_tokens}: the Foldseek 3Di alphabet~\citep{Foldseek}
    applied to the first frame of each protein.
    Foldseek's pretrained encoder maps each residue's $10$-D descriptor (from
    N, C$\alpha$, C and C$\beta$) to a $2$-D latent and assigns the nearest of $20$
    centroids, the 3Di states learned with a VQ-VAE; the probe receives that
    \textit{frozen} $2$-D centroid, computed with
    \href{https://github.com/althonos/mini3di}{mini3di}. Residues without
    a 3Di descriptor, such as chain ends, take Foldseek's default state.
    \item \textit{Random}: a sanity-check baseline in which every residue
    receives a fixed random $2$-D Gaussian vector, seeded by the protein
    and the residue position so that it does not change between probe seeds. This isolates
    the contribution of the learned vocabulary from the downstream
    classifier capacity.
    \item \textit{\tool, P=1}: the single-frame \tool tokenizer. Tokens
    are obtained by computing the same per-residue descriptor used at
    training time -- the 256-D ESM3-style $K{=}16$ nearest-neighbor relative
    SE(3) features with backbone dihedrals -- from a \emph{single} conformation, and passing it
    through the multi-frame set encoder at $P{=}1$. Because the encoder
    is permutation-invariant over the frame axis, the $P{=}1$ inference
    mode requires no architectural change. The vocabulary is the
    primary RVQ codebook ($M_1 = 2048$).
    \item \textit{AminoAseed}: a single-frame structural VQ tokenizer
    from StructTokenBench~\citep{AminoAseed}; the
    \texttt{codebook\_512x1024-1e+19-linear-fixed-last} checkpoint trained on
    PDB. Per residue, the encoder (consuming only N/C$\alpha$/C atoms)
    produces a VQ index in $[0, 512)$ which we look up in the frozen
    $1024$-D codebook after the quantizer's linear projection.
    \item \textit{ProToken}: ProToken-1.0~\citep{ProToken}, a
    single-frame structural VQ tokenizer trained on PDB. Per residue,
    the encoder (consuming a minimal N/C$\alpha$/C/O backbone; it does not read the residue identity) emits a
    VQ index in $[0, 512)$, which we look up in its frozen $32$-D
    codebook. A ProToken token does not describe its own residue: the
    encoder attends over the whole chain and is trained only so that the decoder,
    which reads all tokens together, rebuilds the structure, so the information
    about a residue is spread over neighbouring tokens.
    \item \textit{ESM3struct}: the structure-track VQ tokenizer of ESM3~\citep{ESM3}, specifically the \texttt{StructureTokenEncoder} of
    \texttt{esm3-sm-open-v1}. Per residue, the encoder (consuming
    N/C$\alpha$/C atoms) emits a VQ index in $[0, 4096)$, which we look
    up in the frozen $128$-D EMA codebook.
\end{itemize}

\paragraph{Multi-frame methods:}
\begin{itemize}
    \item \textit{ProtProfileMD\_K}: per-residue profile over the
    20 3Di structural states, built with the ProtProfileMD
    recipe~\citep{ProtProfileMD}: Foldseek assigns a 3Di state to each of the $K$
    MD frames, and MMseqs2 \texttt{result2profile} (pseudocounts, masking and
    composition correction off) turns the $K$ state sequences into a per-residue
    profile, so each bin is a sequence-weighted frequency rather than a raw frame
    count; chains shorter than four residues receive an all-zero profile;
    $K{=}8$ on MISATO and $K{=}10$ on mdCATH-div and BioEmu-CATH. We additionally interleave each
    bin's probability with the fixed 2-D mini3di centroid of that
    state, yielding a $60$-D per-residue feature.
    \item \textit{Vote\_3di}: majority-vote 3Di tokens across the
    same $K$ frames as ProtProfileMD\_K per residue. A state wins if it occurs in at least two
    frames and more often than any other; otherwise (a tie, or all frames
    different) the state whose centroid is nearest to the mean of the frames'
    centroids is kept. The feature is the winning state's $2$-D mini3di centroid.
    Captures the most-frequent local structural state.
    \item \textit{\tool, P=full}: the multi-frame \tool tokenizer run on
    the full $P$-frame ensemble of each protein. Identical model and
    codebook as \textit{\tool, P=1}; differs only in the number of
    conformations fed to the permutation-invariant set encoder
    ($P{=}10$ on mdCATH-div and BioEmu-CATH, $P{=}8$ on MISATO).
\end{itemize}

\newpage
\section{Tokens encode distinguishable local dynamics: an ANOVA test}
\label{app:token_anova}

Throughout the paper we claim that \tool tokens capture
\emph{ensemble dynamics} rather than purely static structure. The clean
way to test this claim is to ask: \textit{does knowing a residue's token
identity reduce uncertainty about its local fluctuation amplitude in MD?}
We answer this with a one-way analysis of variance (ANOVA) on the
per-residue C$\alpha$ motion amplitude conditioned on the primary token.

\paragraph{Setup.}
For every residue $r$ in the mdCATH-div corpus we have a $P{=}10$
frame C$\alpha$ ensemble and a primary token assignment
$t(r) \in \{1, \dots, M\}$ with $M = 2048$. To extract a scalar measure
of local fluctuation amplitude that is independent of global rigid-body
motion, we Kabsch-align the $10$ frames using the local C$\alpha$ ball
of radius $10\,\text{\AA}$ around residue $r$ in frame $0$, then compute
the top singular value $s_1(r)$ of the resulting $(10, 3)$
aligned-coordinate matrix. Intuitively, $s_1(r)$ measures how much the
central C$\alpha$ atom fluctuates along its dominant local motion
direction relative to its neighbourhood; small $s_1$ means a rigid
residue, large $s_1$ means a flexible one. No tokens are pooled and no residue is dropped: every tokenizer is evaluated on the same $308{,}240$ residues of the $2{,}442$ domains (the RMSF residue set without the $53$ residues whose $s_1$ is undefined).

\paragraph{One-way ANOVA: variance decomposition.}
ANOVA decomposes the total variance of $s_1$ into
``token-explained'' and ``within-token-residual'' components:
\begin{equation}
\underbrace{\sum_r \bigl(s_1(r) - \bar{s}_1\bigr)^2}_{\mathrm{SS}_{\mathrm{total}}}
\;=\;
\underbrace{\sum_t n_t \bigl(\bar{s}_1^{(t)} - \bar{s}_1\bigr)^2}_{\mathrm{SS}_{\mathrm{between}}}
\;+\;
\underbrace{\sum_t \sum_{r \in t} \bigl(s_1(r) - \bar{s}_1^{(t)}\bigr)^2}_{\mathrm{SS}_{\mathrm{within}}},
\end{equation}
where $\bar{s}_1$ is the corpus mean of $s_1$, $\bar{s}_1^{(t)}$ is the
mean within token $t$, and $n_t$ is the count of residues assigned to
$t$. The decomposition is exact and makes no parametric assumption on
the distribution of $s_1$. Replacing each residue's value by its
token's mean would leave $\mathrm{SS}_{\mathrm{between}}$ as the
remaining variance --- this is the variance that ``token id alone''
recovers.

The headline summary is the proportion of variance explained,
\begin{equation}
\eta^2 \;=\; \frac{\mathrm{SS}_{\mathrm{between}}}{\mathrm{SS}_{\mathrm{total}}} \in [0, 1].
\label{eq:eta2}
\end{equation}
Equivalently, $\eta^2$ is the $R^2$ of the one-hot-token regression
$s_1 \sim \mathrm{onehot}(t)$. Two extreme cases anchor the
interpretation: if all tokens have identical mean $s_1$ then
$\eta^2 = 0$ (token id is uninformative about motion amplitude); if
every residue with the same token has the same $s_1$ then $\eta^2 = 1$
(token id deterministically predicts amplitude).

\paragraph{F-statistic and parametric $p$-value.}
The classical inferential question --- is $\eta^2$ distinguishable from
zero? --- is answered by the $F$-statistic
\begin{equation}
F \;=\; \frac{\mathrm{SS}_{\mathrm{between}} / (M - 1)}{\mathrm{SS}_{\mathrm{within}} / (N - M)} \;\sim\; F(M-1, N-M)
\quad\text{under } H_0,
\end{equation}
where $H_0$ asserts that all tokens have the same true mean. Under
$H_0$, both numerator and denominator estimate the same residual
variance and one expects $F \approx 1$. Values much larger than $1$
indicate that the between-token variance exceeds what residual scatter
alone could produce.

Residues of the same protein are not independent, so the parametric $p$ value of this test is not valid here; we report $F$ as a descriptive statistic and test significance with the circular shift null below.

\paragraph{Permutation null.} Shuffling token labels over all residues would ignore that neighbouring residues of a protein share both tokens and motion. We therefore rotate the token sequence of each protein by a random offset between $5$ and $L-5$ residues, which keeps every protein's token composition and local autocorrelation but breaks the alignment of tokens and motion, recompute $\eta^2$, and repeat $1{,}000$ times; $p = (b+1)/(n+1)$. This null also calibrates the finite sample bias of $\eta^2$: with $1{,}796$ groups the rotated labels still absorb some variance.

\paragraph{Results.}
Table~\ref{tab:token_anova} reports the test for two motion features:
the dominant amplitude $s_1$ (above) and the anisotropy ratio
$s_1 / s_2$ (a scalar measure of how directional the local motion is).

\begin{table}[htbp]
\centering\small
\begin{tabular}{lccc}
\toprule
Feature & $\eta^2$ (observed) & $F$-test & circular shift null mean \\
\midrule
motion amplitude $s_1$ & $0.393$ & $F(1{,}795,\ 306{,}444) = 110.5$ & $0.044$ \\
motion shape $s_1/s_2$ & $0.044$ & $F(1{,}795,\ 306{,}444) = 7.9$ & $0.007$ \\
\bottomrule
\end{tabular}
\caption{One-way ANOVA on per-residue motion features grouped by the primary token of \tool ($P{=}10$, $\lambda{=}10$), $n = 308{,}240$ residues of $2{,}442$ mdCATH-div domains, $M = 1{,}796$ tokens (no pooling). For both features none of $1{,}000$ circular shifts reaches the observed $\eta^2$ ($p < 10^{-3}$).}
\label{tab:token_anova}
\end{table}

Token identity explains \textbf{$\eta^2 = 0.393$} of the variance in motion amplitude ($\omega^2 = 0.389$ after correcting for the number of groups): knowing the token reduces the uncertainty about a residue's local fluctuation. None of the $1{,}000$ circular shifts comes close: the shifted labels give $\eta^2 \approx 0.044$ on average, so the observed value is about $9\times$ the null mean (and about $68\times$ the mean of a plain shuffle of all labels, $0.006$, which ignores the dependence within proteins), $p < 10^{-3}$.

The motion shape result $\eta^2 = 0.044$ is more modest but still
highly significant. This makes sense: motion direction (encoded in
$s_1/s_2$) is geometrically harder to align across residues with
different local frames, while motion amplitude is a scalar quantity
that the codebook can carve into clean strata.

\paragraph{Interpretation.}
The codebook is not interchangeable on dynamics: each token corresponds
to a statistically distinguishable distribution of local motion. A
useful comparison: a perfect oracle that mapped each residue to its
own state would have $\eta^2 = 1$, the same tokens permuted over the residues give $\eta^2 \approx 0.006$, and \tool sits at $\eta^2 = 0.393$, about two fifths of the way to the oracle. The remaining $61\%$ of variance is residue-specific noise the token does
not capture (sequence context, position in the fold, side-chain
identity), which is consistent with the codebook being a deliberately
compact summary of geometry rather than a per-residue identifier.

\begin{figure}[htbp]
    \centering
    \includegraphics[width=1\linewidth]{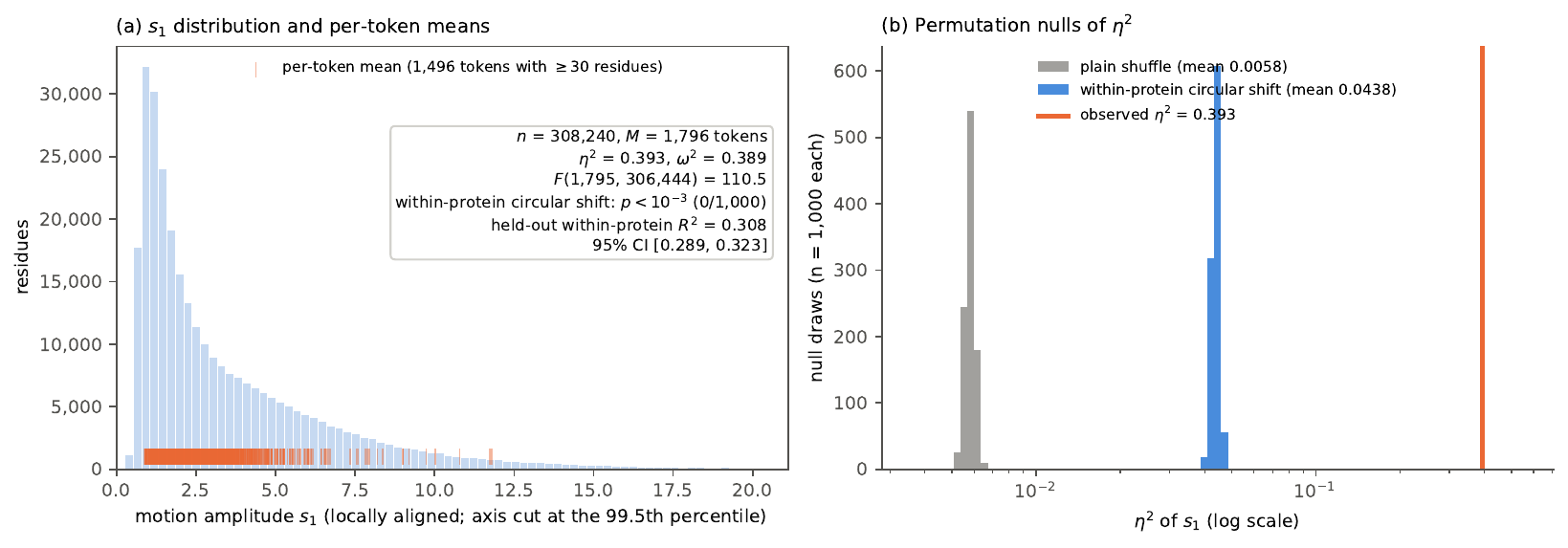}
    \caption{ANOVA test that different tokens encode different dynamics. Per-residue motion amplitude $s_1$ is the top singular value of the locally Kabsch-aligned $P{=}10$ C$\alpha$-frame matrix around each residue, on $n = 308{,}240$ residues and $M = 1{,}796$ tokens. \textbf{(a)} Histogram of $s_1$ (blue) with the mean of every token with at least $30$ residues as a red tick (display only; the statistics use every token). Inset: $\eta^2 = 0.393$, $\omega^2 = 0.389$, $F(1{,}795, 306{,}444) = 110.5$, circular shift $p < 10^{-3}$, held-out $R^2 = 0.308$ $[0.289, 0.323]$. \textbf{(b)} Null distributions of $\eta^2$ from $1{,}000$ within-protein circular shifts and, for reference, $1{,}000$ plain shuffles (log axis), with the observed $\eta^2$ in red.}
    \label{fig:anova_test}
\end{figure}

\paragraph{Negative controls.}
The above $\eta^2$ value rules out the null hypothesis that token id is
uninformative about motion amplitude, but it leaves open a second
concern: are the tokens merely recovering a coarse confounder such as
fold class, chain position, or protein size? To rule this out we
re-run the same one-way ANOVA on $s_1$ with three alternative
grouping variables that should explain at most a small share of the
amplitude variance: CATH structural class (4 broad fold types plus special bucket),
residue position quintile within the chain ($5$ groups), and
protein-length quintile ($5$ groups). Table~\ref{tab:token_anova_controls}
reports the resulting $\eta^2$. None of the controls comes close to
the token signal: CATH class and protein length each capture
$\eta^2 \approx 0.054$ and residue position only $\eta^2 = 0.020$,
each about $7\times$ to $20\times$ smaller than the token result. The
codebook is therefore distinguishing residues at a much finer
granularity than fold class or chain position, and is not reducible
to either confounder.

\begin{table}[htbp]
\centering\small
\begin{tabular}{lcc}
\toprule
Grouping variable & \# groups $M$ & $\eta^2$ on $s_1$ \\
\midrule
\textbf{\tool} & $1{,}796$ & $\mathbf{0.393}$ \\
\midrule
CATH structural class                   & $5$       & $0.054$ \\
Protein-length quintile                  & $5$       & $0.054$ \\
Residue position quintile (within chain) & $5$       & $0.020$ \\
\bottomrule
\end{tabular}
\caption{One-way ANOVA on per-residue motion amplitude $s_1$ across
$308{,}240$ residues, with grouping by \tool primary token and
three negative controls. Tokens explain $\sim 7\times$ more variance
than the strongest non-token baseline, ruling out fold-class or
position-along-chain as alternative explanations of the result in
Fig.~\ref{fig:anova_test}.}
\label{tab:token_anova_controls}
\end{table}

\paragraph{Comparison to existing structural tokenizers.}
We run the same $\eta^2$ test using each baseline tokenizer's per-residue
token assignment as the grouping variable, on the same ${\sim}308$k
mdCATH-div residues (Table~\ref{tab:token_anova_baselines}).
\tool has the strongest amplitude conditioning by a wide margin:
$\eta^2 = 0.393$ against $\le 0.125$ for every other tokenizer. The
gap separates into four regimes: (i) ProToken sits at $\eta^2 = 0.007$
(a ProToken token does not describe its own residue; App.~\ref{app:baselines}); (ii) the
single-frame structure tokenizers 3Di\_tokens $0.051$ and ESM3struct $0.066$ ($6\times$ to $8\times$ less dynamics signal than
\tool), despite ESM3struct's PDB-scale pretraining; (iii) AminoAseed,
a single-frame structural VQ-VAE with a learned $512$-code vocabulary,
sits at $\eta^2 = 0.080$ ($\sim 5\times$ below \tool),
suggesting its codebook has organically discovered some
flexibility-correlated structural classes beyond the fixed-vocabulary
3Di alphabet; (iv) multi-frame aggregations of mini3di tokens
(Vote\_3Di $0.125$; ProtProfileMD\_K10 profile $R^2$ $0.122$), still $\sim 3\times$ below \tool.
The closest baseline is within $\sim 3\times$ of \tool and the
next tier within $\sim 5\times$, so the gap is not ``an order of
magnitude'' once the multi-frame mini3di aggregations are included;
nevertheless every non-\tool row sits well below the half-height of
the \tool result.

\begin{table}[htbp]
\centering\small
\begin{tabular}{lcccc}
\toprule
Method & Regime & Held-out $R^2$ $s_1$ [95\% CI] & Held-out $R^2$ RMSF [95\% CI] & Gain over confounds $s_1$ [CI] \\
\midrule
\textbf{\tool{} $P{=}10$} ($\lambda{=}10$) & 10-frame & \textbf{0.308} [0.289, 0.323] & \textbf{0.528} [0.515, 0.540] & \textbf{0.224} [0.215, 0.233] \\
\tool{} $P{=}1$ ($\lambda{=}10$)           & single   & 0.182 [0.173, 0.190] & 0.327 [0.314, 0.341] & 0.074 [0.070, 0.079] \\
\midrule
Vote\_3Di                                   & 10-frame & 0.097 [0.089, 0.104] & 0.177 [0.169, 0.184] & 0.049 [0.046, 0.052] \\
ProtProfileMD\_K10 ($R^2$)                  & 10-frame & 0.102 [0.094, 0.110] & 0.191 [0.182, 0.199] & 0.049 [0.046, 0.052] \\
AminoAseed                                  & single   & 0.087 [0.082, 0.092] & 0.141 [0.136, 0.146] & 0.033 [0.031, 0.035] \\
ESM3struct                                  & single   & 0.041 [0.036, 0.046] & 0.091 [0.087, 0.094] & 0.015 [0.012, 0.018] \\
3Di (mini3di, single-frame)                 & single   & 0.044 [0.040, 0.047] & 0.106 [0.101, 0.110] & 0.010 [0.009, 0.012] \\
ProToken                                    & single   & 0.005 [0.003, 0.006] & 0.003 [0.002, 0.004] & 0.002 [0.001, 0.003] \\
Amino acid                                  & ---      & 0.009 [0.008, 0.011] & 0.021 [0.020, 0.023] & (in confounds) \\
\bottomrule
\end{tabular}
\caption{Token-conditioned ANOVA on per-residue motion amplitude $s_1$
(from locally Kabsch-aligned $P{=}10$ C$\alpha$ frames) and all-frames C$\alpha$ RMSF across
$308{,}240$ residues in $2{,}442$ mdCATH-div proteins.
All circular-shift $p < 10^{-3}$.
Confounds-only $R^2$: $s_1$ $0.180$; RMSF $0.374$.
Tokens with at least $30$ residues are shown in the rug of Fig.~\ref{fig:anova_test}.\tool ($P{=}10$) explains $3.5\times$ more $s_1$ variance than the best static
tokenizer (AminoAseed $0.087$), with the gain persisting after controlling
for confounds ($0.224$ vs $0.033$).}
\label{tab:token_anova_baselines}
\end{table}

\newpage
\section{Downstream task experiments}
\label{app:downstreams}

This appendix expands the downstream summary in
Section~\ref{sec:downstreams}. We give the full task setup, evaluation
protocol, and per-task tables for EC classification, GO term
prediction, binding-site identification, and binding-affinity
regression on the misato structure split, the most pertinent of the
sequence, structure, and random splits. The complete experiments on the other splits are documented in Appendix~\ref{app:experiments}. Seed standard deviations of the primary metrics (mAP, AUROC) are at most $0.03$; affinity varies more across seeds (Spearman standard deviation up to $0.16$ on the structure split).

\begin{figure}[htbp]
    \centering
    \includegraphics[width=1\linewidth]{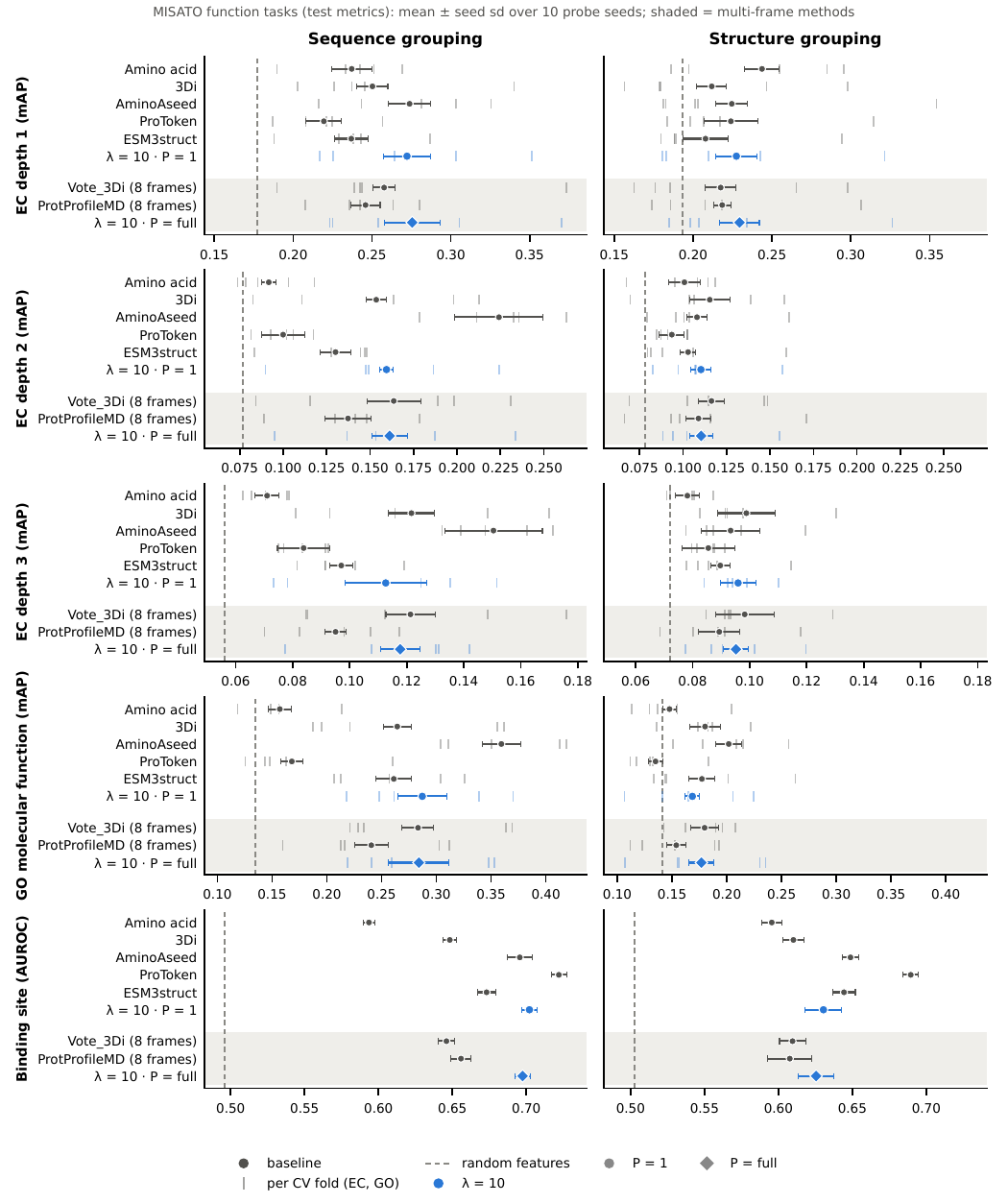}
    \caption{Function task results (test split) across MISATO downstream tasks.
    Each panel shows one task and metric; dots are methods, error bars are
    $\pm 1$ sample standard deviation over seeds. \tool ($P{=}1$ and $P{=}\text{full}$)
    is competitive with static tokenizers on most tasks; AminoAseed leads EC/GO
    on the sequence split while ProToken leads binding AUROC on the structure split.}
    \label{fig:downstream_radar}
\end{figure}

\paragraph{Probe training.}
All heads read the frozen level-1 codeword of each residue (for the
baselines, their frozen token embedding or feature). The EC, GO,
binding-site and binding-affinity heads are trained with AdamW
(learning rate $10^{-3}$, weight decay $10^{-4}$, batch size $32$
proteins), a $5$ epoch linear warm-up followed by cosine decay, and early
stopping on the validation loss with patience $12$ after at least $20$
epochs, for at most $40$ epochs ($80$ for binding affinity); the test
metrics are those of the best validation epoch. No seed is rerun or excluded.
Inputs are standardised with statistics of the training data (embedding tables over their rows, continuous features over the training residues; MACCS bits are left raw).
The RMSF probe is an
MLP with two hidden layers of size $128$ (GELU, dropout $0.1$) trained with MSE,
AdamW (learning rate $3\times10^{-4}$, weight decay $10^{-4}$, batch size
$2{,}048$ residues), stopped when the pooled validation Spearman correlation has
not improved by more than $10^{-4}$ for $20$ epochs (at most $500$); the weights of
the best validation epoch are evaluated once on the test residues. Probe seeds $0$
to $9$ change the initialization, dropout and batch order; tokens and features are
fixed. Besides the Spearman correlation pooled over all test residues, we report the
Spearman correlation within each test protein, averaged over the test proteins with
at least $10$ residues and non-constant labels; a method whose predictions are
constant within such a protein scores $0$ on it. A protein is a domain on
mdCATH-div, a simulated system on BioEmu-CATH (a domain simulated in both deposits
counts twice) and a complex on MISATO.

\begin{table}[htbp]
\centering\small
\caption{RMSF Spearman correlation within each test protein (proteins with at least $10$ residues and
non-constant labels), averaged over proteins; mean $\pm$ sample std over $10$ probe seeds. Same probes,
splits and frames as Table~\ref{tab:rmsf}. Bold: best, underline: second best, within each block.}
\label{tab:rmsf_per_protein}
\begin{tabular}{lcccc}
\toprule
Model & mdCATH-div & BioEmu-CATH & MISATO (seq) & MISATO (struct) \\
\midrule
AA         & 0.255 $\pm$ 0.001 & 0.259 $\pm$ 0.002 & 0.263 $\pm$ 0.001 & 0.284 $\pm$ 0.002 \\
3Di        & 0.398 $\pm$ 0.001 & 0.415 $\pm$ 0.004 & 0.412 $\pm$ 0.000 & 0.443 $\pm$ 0.001 \\
AminoAseed & \underline{0.524 $\pm$ 0.003} & \underline{0.536 $\pm$ 0.005} & \underline{0.588 $\pm$ 0.001} & \underline{0.613 $\pm$ 0.001} \\
ProToken   & 0.006 $\pm$ 0.002 & 0.031 $\pm$ 0.010 & 0.053 $\pm$ 0.002 & 0.057 $\pm$ 0.006 \\
ESM3struct & 0.460 $\pm$ 0.003 & 0.481 $\pm$ 0.003 & 0.525 $\pm$ 0.001 & 0.560 $\pm$ 0.001 \\
\tool, $P{=}1$ & \textbf{0.642 $\pm$ 0.003} & \textbf{0.678 $\pm$ 0.002} & \textbf{0.654 $\pm$ 0.001} & \textbf{0.680 $\pm$ 0.001} \\
\midrule
ProtProfileMD\_K & \underline{0.601 $\pm$ 0.001} & \underline{0.640 $\pm$ 0.008} & \underline{0.488 $\pm$ 0.001} & \underline{0.513 $\pm$ 0.001} \\
Vote\_3Di        & 0.474 $\pm$ 0.000 & 0.516 $\pm$ 0.000 & 0.428 $\pm$ 0.000 & 0.456 $\pm$ 0.001 \\
\tool, $P{=}\text{full}$ & \textbf{0.716 $\pm$ 0.001} & \textbf{0.775 $\pm$ 0.001} & \textbf{0.658 $\pm$ 0.001} & \textbf{0.684 $\pm$ 0.002} \\
\bottomrule
\end{tabular}
\end{table}

\paragraph{EC Function Prediction}
We evaluate \tool tokens on Enzyme Commission (EC) class
prediction~\citep{DeepFRI,Proteinshake} on distinct proteins (one representative complex per set of UniProt accessions) with grouped, label stratified 5-fold cross-validation: in rotation $r$ fold $r$ is the test fold, fold $r{+}1$ the validation fold and the other three train the head; each seed's value is the mean over the five test folds, and we report mean $\pm$ sample std over $10$ seeds. mAP is the primary metric. Each protein
is annotated with one or more EC numbers; we collapse them to depth-1
(the top-level enzyme class) for the headline table and report depths
2 and 3 in Appendix~\ref{app:experiments}. Per-residue features are
obtained as token embeddings, and a 1D-CNN classifier is fit on top:
a stack of three Conv1D layers, masked mean$+$max pooling across
residues, and a small MLP head producing per-class logits trained
with multi-label binary cross-entropy. \footnote{In hindsight, such 1D Conv layers should be accompanied with positional embeddings of the tokens. We will leave it for the readers to investigate.} We report five metrics on the
test split: \textit{top-1 hit} --- the fraction of test proteins
whose highest-scored class is among the ground-truth labels;
\textit{mAP} (macro-averaged AP) and \textit{$\mu$AP} (micro-averaged
AP) --- average-precision computed per class then averaged, and over
all (protein, class) pairs pooled, respectively; and the macro /
micro F1 at the 0.5 decision threshold (\textit{F1@0.5},
\textit{$\mu$F1@0.5}). The label set is the classes seen in the
training proteins; validation and test proteins with none of these
classes are dropped, and all metrics are computed over the classes with
at least one positive test protein. Table~\ref{tab:ec_d1_structure} reports the
depth-1 results; full per-depth breakdowns are deferred to
Appendix~\ref{app:experiments}.

\textbf{Two caveats apply to every function table. First, ESM3struct and AminoAseed were pretrained on PDB scale structural corpora that almost certainly overlap with the MISATO test proteins. Second, ESM3struct's tokenizer is trained jointly with an inverse folding objective, so its tokens carry sequence information.}

On the structure split (Table~\ref{tab:ec_d1_structure}), amino acid identity has the best top-1 ($0.442$) and mAP ($0.244$) among single-frame methods and \tool, $P{=}1$ is second on both ($0.436$, $0.227$) and best on $\mu$AP and the two F1 scores. Among multi-frame methods, \tool, $P{=}\text{full}$ has the best mAP ($0.229$) and ProtProfileMD\_K8 the best top-1 ($0.448$), $\mu$AP and F1 scores. All methods are close to the majority class rate (top-1 $0.411$), and no difference on the structure split is significant (Appendix~\ref{app:significance}).

\begin{table}[htbp]
\centering\small
\caption{EC depth 1, MISATO structure split, grouped 5-fold cross-validation on distinct proteins (groups: CATH H superfamilies); per seed, the mean over the five test folds; mean $\pm$ sample std over $10$ seeds. Bold: best, underline: second best, separately for the single-frame (top) and multi-frame (bottom) blocks.}
\label{tab:ec_d1_structure}
\begin{tabular}{lccccc}
\toprule
Model & top-1 & mAP & $\mu$AP & F1@0.5 & $\mu$F1@0.5 \\
\midrule
aa & \textbf{0.442 $\pm$ 0.023} & \textbf{0.244 $\pm$ 0.011} & \underline{0.425 $\pm$ 0.033} & 0.079 $\pm$ 0.029 & 0.219 $\pm$ 0.085 \\
3Di\_tokens & 0.411 $\pm$ 0.012 & 0.212 $\pm$ 0.009 & 0.397 $\pm$ 0.006 & 0.081 $\pm$ 0.018 & 0.238 $\pm$ 0.057 \\
Random & 0.405 $\pm$ 0.010 & 0.193 $\pm$ 0.004 & 0.381 $\pm$ 0.006 & 0.008 $\pm$ 0.007 & 0.023 $\pm$ 0.021 \\
AminoAseed & 0.427 $\pm$ 0.024 & 0.225 $\pm$ 0.010 & 0.412 $\pm$ 0.023 & \underline{0.084 $\pm$ 0.016} & \underline{0.250 $\pm$ 0.054} \\
ProToken & 0.405 $\pm$ 0.026 & 0.224 $\pm$ 0.017 & 0.398 $\pm$ 0.013 & 0.052 $\pm$ 0.017 & 0.156 $\pm$ 0.058 \\
ESM3struct & 0.423 $\pm$ 0.021 & 0.208 $\pm$ 0.014 & 0.395 $\pm$ 0.016 & 0.079 $\pm$ 0.020 & 0.220 $\pm$ 0.069 \\
\tool, $P{=}1$ & \underline{0.436 $\pm$ 0.023} & \underline{0.227 $\pm$ 0.013} & \textbf{0.428 $\pm$ 0.019} & \textbf{0.106 $\pm$ 0.012} & \textbf{0.305 $\pm$ 0.037} \\
\midrule
ProtProfileMD\_K8 & \textbf{0.448 $\pm$ 0.018} & \underline{0.218 $\pm$ 0.005} & \textbf{0.429 $\pm$ 0.011} & \textbf{0.109 $\pm$ 0.016} & \textbf{0.308 $\pm$ 0.047} \\
Vote\_3Di & 0.412 $\pm$ 0.017 & 0.217 $\pm$ 0.010 & 0.394 $\pm$ 0.008 & 0.088 $\pm$ 0.014 & 0.260 $\pm$ 0.047 \\
\tool, $P{=}\text{full}$ & \underline{0.436 $\pm$ 0.036} & \textbf{0.229 $\pm$ 0.013} & \underline{0.427 $\pm$ 0.020} & \underline{0.097 $\pm$ 0.020} & \underline{0.276 $\pm$ 0.052} \\
\bottomrule
\end{tabular}
\end{table}

\paragraph{GO Label Prediction}
We evaluate top-50 Gene Ontology term
prediction~\citep{DeepFRI,Proteinshake} under the misato structure
split (Table~\ref{tab:go_structure}). Each protein is annotated with
its SIFTS GO molecular function terms, propagated to their ancestors, without the root, protein binding and do-not-annotate terms. The classes are the $50$ most frequent terms among those annotated to at least $10$ and to fewer than half of the training proteins of each fold. Architecture, optimizer, loss, metrics and cross-validation are as for EC.

On the structure split, AminoAseed has the best mAP among single-frame methods ($0.202$) and ESM3struct the best top-1 ($0.428$), with \tool, $P{=}1$ second on top-1 ($0.413$). Among multi-frame methods, \tool, $P{=}\text{full}$ is best on top-1 ($0.430$), $\mu$AP and both F1 scores, and Vote\_3Di on mAP ($0.180$ vs $0.177$). The micro averaged AP of every method is close to the frequency floor ($0.265$), so on this split it mostly reflects label frequency.

\begin{table}[htbp]
\centering\small
\caption{GO molecular function (top 50 terms), MISATO structure split, grouped 5-fold cross-validation on distinct proteins (groups: CATH H superfamilies); per seed, the mean over the five test folds; mean $\pm$ sample std over $10$ seeds. Bold: best, underline: second best, separately for the single-frame (top) and multi-frame (bottom) blocks.}
\label{tab:go_structure}
\begin{tabular}{lccccc}
\toprule
Model & top-1 & mAP & $\mu$AP & F1@0.5 & $\mu$F1@0.5 \\
\midrule
aa & 0.389 $\pm$ 0.047 & 0.148 $\pm$ 0.006 & 0.248 $\pm$ 0.011 & 0.012 $\pm$ 0.008 & 0.051 $\pm$ 0.032 \\
3Di\_tokens & 0.353 $\pm$ 0.023 & \underline{0.180 $\pm$ 0.014} & 0.252 $\pm$ 0.012 & \underline{0.036 $\pm$ 0.008} & \underline{0.106 $\pm$ 0.027} \\
Random & 0.368 $\pm$ 0.033 & 0.141 $\pm$ 0.003 & \textbf{0.260 $\pm$ 0.008} & 0.001 $\pm$ 0.001 & 0.007 $\pm$ 0.007 \\
AminoAseed & 0.376 $\pm$ 0.029 & \textbf{0.202 $\pm$ 0.012} & 0.248 $\pm$ 0.016 & \textbf{0.049 $\pm$ 0.015} & \textbf{0.116 $\pm$ 0.035} \\
ProToken & 0.383 $\pm$ 0.043 & 0.135 $\pm$ 0.007 & 0.235 $\pm$ 0.013 & 0.008 $\pm$ 0.003 & 0.040 $\pm$ 0.015 \\
ESM3struct & \textbf{0.428 $\pm$ 0.049} & 0.177 $\pm$ 0.012 & \underline{0.260 $\pm$ 0.016} & 0.031 $\pm$ 0.017 & 0.086 $\pm$ 0.030 \\
\tool, $P{=}1$ & \underline{0.413 $\pm$ 0.030} & 0.169 $\pm$ 0.007 & 0.257 $\pm$ 0.017 & 0.033 $\pm$ 0.009 & 0.078 $\pm$ 0.017 \\
\midrule
ProtProfileMD\_K8 & \underline{0.420 $\pm$ 0.041} & 0.154 $\pm$ 0.009 & 0.228 $\pm$ 0.013 & 0.023 $\pm$ 0.008 & 0.069 $\pm$ 0.025 \\
Vote\_3Di & 0.365 $\pm$ 0.043 & \textbf{0.180 $\pm$ 0.013} & \underline{0.250 $\pm$ 0.015} & \underline{0.035 $\pm$ 0.010} & \underline{0.100 $\pm$ 0.027} \\
\tool, $P{=}\text{full}$ & \textbf{0.430 $\pm$ 0.033} & \underline{0.177 $\pm$ 0.011} & \textbf{0.275 $\pm$ 0.017} & \textbf{0.040 $\pm$ 0.014} & \textbf{0.102 $\pm$ 0.031} \\
\bottomrule
\end{tabular}
\end{table}

\paragraph{Binding Site}
We evaluate per-residue binding-site prediction under the misato
structure split (Table~\ref{tab:binding_structure}). Each residue
carries a binary label indicating whether it lies within the PDBbind pocket (residues within approximately $10$\,\AA{} of any ligand heavy atom), aligned to the MISATO residue list; for the $130$ complexes whose ligand in the first MD frame lies more than $5$\,\AA{} from the crystal pose, the pocket is taken from the simulated ligand (C$\alpha$ within $10$\,\AA{} of any ligand heavy atom), and $4$ complexes with fewer than $5$ such residues are dropped. Per-residue features are
obtained as for EC and GO, and a per-residue 1D-CNN head with the
same conv stack (no pooling --- the prediction is per residue)
outputs a binding-site logit, trained with masked binary
cross-entropy. We report two metrics, both pooled over all test-set
residues: \textit{AUROC} (area under the ROC curve) measures ranking
quality independent of decision threshold, and \textit{AP} (average
precision = area under the precision-recall curve) is the
threshold-free metric of choice for this strongly imbalanced task
(positive rate $\approx 0.10$ on the sequence and structure test sets and $\approx 0.13$ on the random test set).

On the structure split, ProToken has the best AUROC ($0.689$) and AP ($0.197$), followed by AminoAseed ($0.649$, $0.168$) and ESM3struct ($0.644$, $0.165$); \tool, $P{=}1$ reaches $0.630$ and $0.150$, and \tool, $P{=}\text{full}$ is the best multi-frame method ($0.625$, $0.147$). ProToken's advantage over \tool on this split is significant (pooled AUROC $\Delta = -0.057$, Holm $p = 0.005$). On the sequence split \tool, $P{=}1$ is second to ProToken ($0.702$ vs $0.722$), and on the random split every method except 3Di\_tokens, Vote\_3Di and Random exceeds AUROC $0.88$.

\begin{table}[htbp]
\centering\small
\caption{Binding site, MISATO structure split, fixed split, mean $\pm$ sample std over $10$ seeds. Bold: best, underline: second best, separately for the single-frame (top) and multi-frame (bottom) blocks.}
\label{tab:binding_structure}
\begin{tabular}{lcc}
\toprule
Model & AUROC & AP \\
\midrule
aa & 0.595 $\pm$ 0.007 & 0.137 $\pm$ 0.003 \\
3Di\_tokens & 0.610 $\pm$ 0.007 & 0.140 $\pm$ 0.002 \\
Random & 0.503 $\pm$ 0.001 & 0.101 $\pm$ 0.000 \\
AminoAseed & \underline{0.649 $\pm$ 0.006} & \underline{0.168 $\pm$ 0.007} \\
ProToken & \textbf{0.689 $\pm$ 0.005} & \textbf{0.197 $\pm$ 0.005} \\
ESM3struct & 0.644 $\pm$ 0.008 & 0.165 $\pm$ 0.006 \\
\tool, $P{=}1$ & 0.630 $\pm$ 0.012 & 0.150 $\pm$ 0.007 \\
\midrule
ProtProfileMD\_K8 & 0.608 $\pm$ 0.015 & \underline{0.142 $\pm$ 0.005} \\
Vote\_3Di & \underline{0.610 $\pm$ 0.009} & 0.140 $\pm$ 0.002 \\
\tool, $P{=}\text{full}$ & \textbf{0.625 $\pm$ 0.012} & \textbf{0.147 $\pm$ 0.006} \\
\bottomrule
\end{tabular}
\end{table}

\paragraph{Binding Affinity}
We additionally evaluate ligand-affinity regression under the misato
structure split (Table~\ref{tab:affinity_structure}). The label is
$-\log K_d/K_i$ for the bound ligand of each complex, and the head
conditions on a $167$-bit MACCS structural-key fingerprint of the
ligand (resolved from the MISATO \texttt{ligand\_id} via the PDB Chemical Component Dictionary, or from the PDBbind ligand file for $31$ complexes) concatenated to a mean$+$max-pooled
protein representation, after projecting the fingerprint to $64$ dimensions (linear layer, GELU, dropout); the resulting joint vector is mapped to the
affinity scalar by a small MLP. We report mean $\pm$ std over 10
seeds on three metrics: $R^2$ (variance-explained, higher is better),
Spearman $\rho_S$ (rank correlation, higher is better), and
\textit{MSE} (mean squared error in log-affinity units,
\textbf{lower is better}).

Affinity on the structure split is the hardest task: $R^2$ ranges from $-0.057$ to $0.097$ and Spearman from $0.226$ to $0.390$, with seed standard deviations up to $0.16$, and no difference between tokenizers is significant (Appendix~\ref{app:significance}). The controls in Table~\ref{tab:affinity_controls} show where the signal comes from: the fingerprint alone reaches Spearman $0.360$ on the structure split and $0.443$ on the sequence split, while the same head without the fingerprint stays at or below $0.126$ and $0.048$; the best structure split Spearman ($0.390$) is reached with random protein features plus the fingerprint. On the structure split the ligand is the driving force. Only on the random split, where test complexes resemble training complexes, does a learned protein representation add to the fingerprint (Spearman $0.695$ to $0.728$ against $0.689$ for the fingerprint alone and $0.645$ with random protein features).

\begin{table}[htbp]
\centering\small
\caption{Binding affinity ($-\log K_d/K_i$), MISATO structure split, ligand aware head (MACCS-167), fixed split, mean $\pm$ sample std over $10$ seeds. Bold: best, underline: second best, separately for the single-frame (top) and multi-frame (bottom) blocks. \textbf{MSE: lower is better.}}
\label{tab:affinity_structure}
\begin{tabular}{lccc}
\toprule
Model & $R^2$ & Spearman & MSE \\
\midrule
aa & 0.032 $\pm$ 0.061 & 0.277 $\pm$ 0.141 & 2.925 $\pm$ 0.185 \\
3Di\_tokens & $-$0.057 $\pm$ 0.090 & 0.231 $\pm$ 0.149 & 3.193 $\pm$ 0.272 \\
Random & 0.074 $\pm$ 0.032 & \textbf{0.390 $\pm$ 0.007} & 2.799 $\pm$ 0.098 \\
AminoAseed & 0.009 $\pm$ 0.055 & 0.236 $\pm$ 0.137 & 2.994 $\pm$ 0.166 \\
ProToken & 0.048 $\pm$ 0.054 & 0.358 $\pm$ 0.032 & 2.875 $\pm$ 0.162 \\
ESM3struct & \textbf{0.097 $\pm$ 0.063} & \underline{0.382 $\pm$ 0.034} & \textbf{2.730 $\pm$ 0.190} \\
\tool, $P{=}1$ & \underline{0.093 $\pm$ 0.027} & 0.372 $\pm$ 0.021 & \underline{2.742 $\pm$ 0.083} \\
\midrule
ProtProfileMD\_K8 & \underline{$-$0.019 $\pm$ 0.072} & \underline{0.317 $\pm$ 0.028} & \underline{3.081 $\pm$ 0.219} \\
Vote\_3Di & $-$0.055 $\pm$ 0.101 & 0.226 $\pm$ 0.158 & 3.187 $\pm$ 0.306 \\
\tool, $P{=}\text{full}$ & \textbf{0.086 $\pm$ 0.035} & \textbf{0.375 $\pm$ 0.023} & \textbf{2.762 $\pm$ 0.105} \\
\bottomrule
\end{tabular}
\end{table}

\newpage
\section{All Experiments}
\label{app:experiments}

This appendix reports the full per-split results for each downstream
benchmark. EC and GO are evaluated on the structure and sequence splits by grouped 5-fold cross-validation; binding site and affinity on the structure, sequence and random splits. Within every table the upper
half lists the per-residue (single-frame) methods and the lower half
lists the per-protein (multi-frame) methods; bold marks the best and
underline the second-best result \emph{within each half} per column.
All entries are mean $\pm$ sample std over $10$ seeds; for EC and GO each seed's value is the mean over the five test folds.

Before reading the section, we would like to remind the reader that the purpose of \tool is to capture protein dynamics, not to improve downstream performance. We include the study here just to show that \tool performs on par with the other tokenizers. Curious readers should design more rigorous experiments to test the functional tasks.
\subsection{EC Function Prediction}

We report Enzyme Commission classification at three taxonomy depths
(top-level enzyme class, subclass, and full EC number) under the structure and sequence splits.

\paragraph{structure split}
The hardest setting: no CATH H superfamily is shared between training and test proteins.

\begin{table}[htbp]
\centering\small
\caption{EC depth 1, structure split (5-fold CV, CATH H superfamilies).}
\begin{tabular}{lccccc}
\toprule
Model & top-1 & mAP & $\mu$AP & F1@0.5 & $\mu$F1@0.5 \\
\midrule
aa & \textbf{0.442 $\pm$ 0.023} & \textbf{0.244 $\pm$ 0.011} & \underline{0.425 $\pm$ 0.033} & 0.079 $\pm$ 0.029 & 0.219 $\pm$ 0.085 \\
3Di\_tokens & 0.411 $\pm$ 0.012 & 0.212 $\pm$ 0.009 & 0.397 $\pm$ 0.006 & 0.081 $\pm$ 0.018 & 0.238 $\pm$ 0.057 \\
Random & 0.405 $\pm$ 0.010 & 0.193 $\pm$ 0.004 & 0.381 $\pm$ 0.006 & 0.008 $\pm$ 0.007 & 0.023 $\pm$ 0.021 \\
AminoAseed & 0.427 $\pm$ 0.024 & 0.225 $\pm$ 0.010 & 0.412 $\pm$ 0.023 & \underline{0.084 $\pm$ 0.016} & \underline{0.250 $\pm$ 0.054} \\
ProToken & 0.405 $\pm$ 0.026 & 0.224 $\pm$ 0.017 & 0.398 $\pm$ 0.013 & 0.052 $\pm$ 0.017 & 0.156 $\pm$ 0.058 \\
ESM3struct & 0.423 $\pm$ 0.021 & 0.208 $\pm$ 0.014 & 0.395 $\pm$ 0.016 & 0.079 $\pm$ 0.020 & 0.220 $\pm$ 0.069 \\
\tool, $P{=}1$ & \underline{0.436 $\pm$ 0.023} & \underline{0.227 $\pm$ 0.013} & \textbf{0.428 $\pm$ 0.019} & \textbf{0.106 $\pm$ 0.012} & \textbf{0.305 $\pm$ 0.037} \\
\midrule
ProtProfileMD\_K8 & \textbf{0.448 $\pm$ 0.018} & \underline{0.218 $\pm$ 0.005} & \textbf{0.429 $\pm$ 0.011} & \textbf{0.109 $\pm$ 0.016} & \textbf{0.308 $\pm$ 0.047} \\
Vote\_3Di & 0.412 $\pm$ 0.017 & 0.217 $\pm$ 0.010 & 0.394 $\pm$ 0.008 & 0.088 $\pm$ 0.014 & 0.260 $\pm$ 0.047 \\
\tool, $P{=}\text{full}$ & \underline{0.436 $\pm$ 0.036} & \textbf{0.229 $\pm$ 0.013} & \underline{0.427 $\pm$ 0.020} & \underline{0.097 $\pm$ 0.020} & \underline{0.276 $\pm$ 0.052} \\
\bottomrule
\end{tabular}
\end{table}

\begin{table}[htbp]
\centering\small
\caption{EC depth 2, structure split (5-fold CV, CATH H superfamilies).}
\begin{tabular}{lccccc}
\toprule
Model & top-1 & mAP & $\mu$AP & F1@0.5 & $\mu$F1@0.5 \\
\midrule
aa & 0.128 $\pm$ 0.024 & 0.101 $\pm$ 0.009 & 0.113 $\pm$ 0.019 & 0.004 $\pm$ 0.003 & 0.026 $\pm$ 0.032 \\
3Di\_tokens & 0.148 $\pm$ 0.010 & \textbf{0.115 $\pm$ 0.012} & \underline{0.149 $\pm$ 0.008} & \underline{0.011 $\pm$ 0.003} & \underline{0.085 $\pm$ 0.011} \\
Random & 0.110 $\pm$ 0.012 & 0.078 $\pm$ 0.004 & 0.094 $\pm$ 0.004 & 0.000 $\pm$ 0.000 & 0.000 $\pm$ 0.000 \\
AminoAseed & \underline{0.149 $\pm$ 0.015} & 0.108 $\pm$ 0.006 & 0.147 $\pm$ 0.011 & \underline{0.011 $\pm$ 0.004} & 0.066 $\pm$ 0.036 \\
ProToken & 0.120 $\pm$ 0.036 & 0.094 $\pm$ 0.007 & 0.110 $\pm$ 0.021 & 0.004 $\pm$ 0.005 & 0.031 $\pm$ 0.037 \\
ESM3struct & \underline{0.149 $\pm$ 0.014} & 0.103 $\pm$ 0.004 & 0.134 $\pm$ 0.016 & 0.009 $\pm$ 0.004 & 0.053 $\pm$ 0.038 \\
\tool, $P{=}1$ & \textbf{0.169 $\pm$ 0.016} & \underline{0.110 $\pm$ 0.006} & \textbf{0.161 $\pm$ 0.007} & \textbf{0.015 $\pm$ 0.002} & \textbf{0.099 $\pm$ 0.011} \\
\midrule
ProtProfileMD\_K8 & \textbf{0.173 $\pm$ 0.011} & 0.109 $\pm$ 0.007 & \underline{0.152 $\pm$ 0.010} & \textbf{0.016 $\pm$ 0.008} & \underline{0.089 $\pm$ 0.034} \\
Vote\_3Di & 0.143 $\pm$ 0.013 & \textbf{0.116 $\pm$ 0.007} & 0.147 $\pm$ 0.009 & 0.011 $\pm$ 0.004 & 0.081 $\pm$ 0.019 \\
\tool, $P{=}\text{full}$ & \underline{0.167 $\pm$ 0.018} & \underline{0.110 $\pm$ 0.007} & \textbf{0.160 $\pm$ 0.009} & \underline{0.015 $\pm$ 0.002} & \textbf{0.105 $\pm$ 0.014} \\
\bottomrule
\end{tabular}
\end{table}

\begin{table}[htbp]
\centering\small
\caption{EC depth 3, structure split (5-fold CV, CATH H superfamilies).}
\begin{tabular}{lccccc}
\toprule
Model & top-1 & mAP & $\mu$AP & F1@0.5 & $\mu$F1@0.5 \\
\midrule
aa & 0.027 $\pm$ 0.008 & 0.078 $\pm$ 0.004 & 0.038 $\pm$ 0.002 & 0.000 $\pm$ 0.000 & 0.000 $\pm$ 0.000 \\
3Di\_tokens & 0.037 $\pm$ 0.008 & \textbf{0.099 $\pm$ 0.010} & 0.058 $\pm$ 0.011 & \textbf{0.006 $\pm$ 0.003} & \underline{0.013 $\pm$ 0.007} \\
Random & 0.024 $\pm$ 0.010 & 0.072 $\pm$ 0.004 & 0.038 $\pm$ 0.001 & 0.000 $\pm$ 0.000 & 0.000 $\pm$ 0.000 \\
AminoAseed & \textbf{0.061 $\pm$ 0.031} & 0.093 $\pm$ 0.010 & \underline{0.078 $\pm$ 0.029} & 0.002 $\pm$ 0.002 & \underline{0.013 $\pm$ 0.026} \\
ProToken & 0.043 $\pm$ 0.025 & 0.086 $\pm$ 0.009 & 0.057 $\pm$ 0.024 & 0.001 $\pm$ 0.002 & 0.008 $\pm$ 0.026 \\
ESM3struct & 0.032 $\pm$ 0.010 & 0.090 $\pm$ 0.003 & 0.053 $\pm$ 0.018 & 0.003 $\pm$ 0.003 & 0.003 $\pm$ 0.003 \\
\tool, $P{=}1$ & \underline{0.056 $\pm$ 0.027} & \underline{0.096 $\pm$ 0.006} & \textbf{0.081 $\pm$ 0.022} & \underline{0.004 $\pm$ 0.003} & \textbf{0.027 $\pm$ 0.037} \\
\midrule
ProtProfileMD\_K8 & \underline{0.049 $\pm$ 0.026} & 0.089 $\pm$ 0.007 & \underline{0.074 $\pm$ 0.021} & 0.002 $\pm$ 0.003 & \underline{0.019 $\pm$ 0.033} \\
Vote\_3Di & 0.044 $\pm$ 0.009 & \textbf{0.098 $\pm$ 0.010} & 0.065 $\pm$ 0.016 & \underline{0.005 $\pm$ 0.003} & 0.011 $\pm$ 0.007 \\
\tool, $P{=}\text{full}$ & \textbf{0.074 $\pm$ 0.016} & \underline{0.095 $\pm$ 0.004} & \textbf{0.098 $\pm$ 0.015} & \textbf{0.007 $\pm$ 0.004} & \textbf{0.067 $\pm$ 0.035} \\
\bottomrule
\end{tabular}
\end{table}

\newpage
\paragraph{sequence split}
Train and test proteins are sequence disjoint (no training and test chains share $30\%$ identity over half of the shorter chain, MMseqs2 graph clustering~\citep{MMseqs2}, with chains of one UniProt entry kept together) but
may share structural class.

\begin{table}[htbp]
\centering\small
\caption{EC depth 1, sequence split (5-fold CV, MMseqs2 clusters at 30\% identity).}
\begin{tabular}{lccccc}
\toprule
Model & top-1 & mAP & $\mu$AP & F1@0.5 & $\mu$F1@0.5 \\
\midrule
aa & 0.490 $\pm$ 0.021 & 0.237 $\pm$ 0.013 & 0.495 $\pm$ 0.021 & 0.112 $\pm$ 0.015 & 0.337 $\pm$ 0.047 \\
3Di\_tokens & 0.501 $\pm$ 0.014 & 0.250 $\pm$ 0.010 & 0.527 $\pm$ 0.011 & 0.132 $\pm$ 0.012 & 0.397 $\pm$ 0.039 \\
Random & 0.431 $\pm$ 0.008 & 0.177 $\pm$ 0.003 & 0.408 $\pm$ 0.007 & 0.020 $\pm$ 0.016 & 0.077 $\pm$ 0.064 \\
AminoAseed & 0.518 $\pm$ 0.024 & \textbf{0.274 $\pm$ 0.013} & \underline{0.535 $\pm$ 0.016} & \underline{0.143 $\pm$ 0.015} & \underline{0.431 $\pm$ 0.032} \\
ProToken & 0.442 $\pm$ 0.015 & 0.219 $\pm$ 0.011 & 0.457 $\pm$ 0.020 & 0.079 $\pm$ 0.024 & 0.265 $\pm$ 0.079 \\
ESM3struct & \underline{0.528 $\pm$ 0.024} & 0.237 $\pm$ 0.011 & 0.520 $\pm$ 0.016 & 0.137 $\pm$ 0.012 & 0.413 $\pm$ 0.036 \\
\tool, $P{=}1$ & \textbf{0.541 $\pm$ 0.021} & \underline{0.272 $\pm$ 0.015} & \textbf{0.563 $\pm$ 0.025} & \textbf{0.155 $\pm$ 0.015} & \textbf{0.457 $\pm$ 0.036} \\
\midrule
ProtProfileMD\_K8 & \underline{0.514 $\pm$ 0.027} & 0.246 $\pm$ 0.009 & 0.523 $\pm$ 0.020 & \underline{0.145 $\pm$ 0.016} & \underline{0.432 $\pm$ 0.039} \\
Vote\_3Di & 0.505 $\pm$ 0.023 & \underline{0.258 $\pm$ 0.007} & \underline{0.530 $\pm$ 0.014} & 0.140 $\pm$ 0.013 & 0.414 $\pm$ 0.037 \\
\tool, $P{=}\text{full}$ & \textbf{0.534 $\pm$ 0.033} & \textbf{0.275 $\pm$ 0.018} & \textbf{0.560 $\pm$ 0.030} & \textbf{0.153 $\pm$ 0.020} & \textbf{0.457 $\pm$ 0.044} \\
\bottomrule
\end{tabular}
\end{table}

\begin{table}[htbp]
\centering\small
\caption{EC depth 2, sequence split (5-fold CV, MMseqs2 clusters at 30\% identity).}
\begin{tabular}{lccccc}
\toprule
Model & top-1 & mAP & $\mu$AP & F1@0.5 & $\mu$F1@0.5 \\
\midrule
aa & 0.156 $\pm$ 0.020 & 0.092 $\pm$ 0.004 & 0.154 $\pm$ 0.017 & 0.007 $\pm$ 0.004 & 0.040 $\pm$ 0.029 \\
3Di\_tokens & 0.245 $\pm$ 0.031 & 0.154 $\pm$ 0.006 & 0.230 $\pm$ 0.020 & 0.026 $\pm$ 0.004 & 0.122 $\pm$ 0.015 \\
Random & 0.127 $\pm$ 0.008 & 0.077 $\pm$ 0.004 & 0.123 $\pm$ 0.004 & 0.000 $\pm$ 0.000 & 0.000 $\pm$ 0.000 \\
AminoAseed & \textbf{0.350 $\pm$ 0.031} & \textbf{0.224 $\pm$ 0.025} & \textbf{0.324 $\pm$ 0.031} & \textbf{0.075 $\pm$ 0.017} & \textbf{0.230 $\pm$ 0.049} \\
ProToken & 0.181 $\pm$ 0.035 & 0.100 $\pm$ 0.012 & 0.155 $\pm$ 0.021 & 0.005 $\pm$ 0.004 & 0.023 $\pm$ 0.026 \\
ESM3struct & 0.237 $\pm$ 0.027 & 0.130 $\pm$ 0.009 & 0.215 $\pm$ 0.012 & 0.022 $\pm$ 0.005 & 0.096 $\pm$ 0.025 \\
\tool, $P{=}1$ & \underline{0.296 $\pm$ 0.030} & \underline{0.159 $\pm$ 0.004} & \underline{0.279 $\pm$ 0.024} & \underline{0.036 $\pm$ 0.005} & \underline{0.172 $\pm$ 0.019} \\
\midrule
ProtProfileMD\_K8 & 0.247 $\pm$ 0.021 & 0.137 $\pm$ 0.013 & 0.203 $\pm$ 0.014 & 0.023 $\pm$ 0.007 & 0.106 $\pm$ 0.029 \\
Vote\_3Di & \underline{0.251 $\pm$ 0.039} & \textbf{0.164 $\pm$ 0.015} & \underline{0.240 $\pm$ 0.028} & \underline{0.029 $\pm$ 0.005} & \underline{0.123 $\pm$ 0.013} \\
\tool, $P{=}\text{full}$ & \textbf{0.285 $\pm$ 0.027} & \underline{0.161 $\pm$ 0.010} & \textbf{0.276 $\pm$ 0.024} & \textbf{0.036 $\pm$ 0.007} & \textbf{0.180 $\pm$ 0.036} \\
\bottomrule
\end{tabular}
\end{table}

\begin{table}[htbp]
\centering\small
\caption{EC depth 3, sequence split (5-fold CV, MMseqs2 clusters at 30\% identity).}
\begin{tabular}{lccccc}
\toprule
Model & top-1 & mAP & $\mu$AP & F1@0.5 & $\mu$F1@0.5 \\
\midrule
aa & 0.097 $\pm$ 0.024 & 0.071 $\pm$ 0.004 & 0.069 $\pm$ 0.010 & 0.001 $\pm$ 0.001 & 0.002 $\pm$ 0.003 \\
3Di\_tokens & 0.166 $\pm$ 0.013 & \underline{0.122 $\pm$ 0.008} & 0.137 $\pm$ 0.013 & 0.008 $\pm$ 0.002 & 0.064 $\pm$ 0.011 \\
Random & 0.062 $\pm$ 0.014 & 0.056 $\pm$ 0.003 & 0.056 $\pm$ 0.003 & 0.000 $\pm$ 0.000 & 0.000 $\pm$ 0.000 \\
AminoAseed & \textbf{0.222 $\pm$ 0.036} & \textbf{0.150 $\pm$ 0.017} & \textbf{0.192 $\pm$ 0.036} & \textbf{0.027 $\pm$ 0.007} & \textbf{0.121 $\pm$ 0.039} \\
ProToken & 0.126 $\pm$ 0.023 & 0.084 $\pm$ 0.009 & 0.105 $\pm$ 0.019 & 0.004 $\pm$ 0.003 & 0.034 $\pm$ 0.031 \\
ESM3struct & 0.133 $\pm$ 0.022 & 0.097 $\pm$ 0.004 & 0.112 $\pm$ 0.018 & 0.005 $\pm$ 0.003 & 0.034 $\pm$ 0.029 \\
\tool, $P{=}1$ & \underline{0.180 $\pm$ 0.031} & 0.113 $\pm$ 0.014 & \underline{0.153 $\pm$ 0.029} & \underline{0.011 $\pm$ 0.001} & \underline{0.078 $\pm$ 0.005} \\
\midrule
ProtProfileMD\_K8 & 0.139 $\pm$ 0.023 & 0.095 $\pm$ 0.004 & 0.110 $\pm$ 0.015 & 0.005 $\pm$ 0.002 & 0.040 $\pm$ 0.029 \\
Vote\_3Di & \underline{0.158 $\pm$ 0.016} & \textbf{0.121 $\pm$ 0.009} & \underline{0.132 $\pm$ 0.013} & \underline{0.007 $\pm$ 0.002} & \underline{0.052 $\pm$ 0.020} \\
\tool, $P{=}\text{full}$ & \textbf{0.176 $\pm$ 0.030} & \underline{0.118 $\pm$ 0.007} & \textbf{0.140 $\pm$ 0.014} & \textbf{0.012 $\pm$ 0.003} & \textbf{0.074 $\pm$ 0.021} \\
\bottomrule
\end{tabular}
\end{table}

\newpage
\subsection{GO Label Prediction}

We report GO molecular function prediction (top 50 terms) under the structure and sequence splits.
\paragraph{structure split}\mbox{}\\

\begin{table}[htbp]
\centering\small
\caption{GO molecular function (top 50), structure split (5-fold CV, CATH H superfamilies).}
\begin{tabular}{lccccc}
\toprule
Model & top-1 & mAP & $\mu$AP & F1@0.5 & $\mu$F1@0.5 \\
\midrule
aa & 0.389 $\pm$ 0.047 & 0.148 $\pm$ 0.006 & 0.248 $\pm$ 0.011 & 0.012 $\pm$ 0.008 & 0.051 $\pm$ 0.032 \\
3Di\_tokens & 0.353 $\pm$ 0.023 & \underline{0.180 $\pm$ 0.014} & 0.252 $\pm$ 0.012 & \underline{0.036 $\pm$ 0.008} & \underline{0.106 $\pm$ 0.027} \\
Random & 0.368 $\pm$ 0.033 & 0.141 $\pm$ 0.003 & \textbf{0.260 $\pm$ 0.008} & 0.001 $\pm$ 0.001 & 0.007 $\pm$ 0.007 \\
AminoAseed & 0.376 $\pm$ 0.029 & \textbf{0.202 $\pm$ 0.012} & 0.248 $\pm$ 0.016 & \textbf{0.049 $\pm$ 0.015} & \textbf{0.116 $\pm$ 0.035} \\
ProToken & 0.383 $\pm$ 0.043 & 0.135 $\pm$ 0.007 & 0.235 $\pm$ 0.013 & 0.008 $\pm$ 0.003 & 0.040 $\pm$ 0.015 \\
ESM3struct & \textbf{0.428 $\pm$ 0.049} & 0.177 $\pm$ 0.012 & \underline{0.260 $\pm$ 0.016} & 0.031 $\pm$ 0.017 & 0.086 $\pm$ 0.030 \\
\tool, $P{=}1$ & \underline{0.413 $\pm$ 0.030} & 0.169 $\pm$ 0.007 & 0.257 $\pm$ 0.017 & 0.033 $\pm$ 0.009 & 0.078 $\pm$ 0.017 \\
\midrule
ProtProfileMD\_K8 & \underline{0.420 $\pm$ 0.041} & 0.154 $\pm$ 0.009 & 0.228 $\pm$ 0.013 & 0.023 $\pm$ 0.008 & 0.069 $\pm$ 0.025 \\
Vote\_3Di & 0.365 $\pm$ 0.043 & \textbf{0.180 $\pm$ 0.013} & \underline{0.250 $\pm$ 0.015} & \underline{0.035 $\pm$ 0.010} & \underline{0.100 $\pm$ 0.027} \\
\tool, $P{=}\text{full}$ & \textbf{0.430 $\pm$ 0.033} & \underline{0.177 $\pm$ 0.011} & \textbf{0.275 $\pm$ 0.017} & \textbf{0.040 $\pm$ 0.014} & \textbf{0.102 $\pm$ 0.031} \\
\bottomrule
\end{tabular}
\end{table}

\paragraph{sequence split}\mbox{}\\

\begin{table}[htbp]
\centering\small
\caption{GO molecular function (top 50), sequence split (5-fold CV, MMseqs2 clusters at 30\% identity).}
\begin{tabular}{lccccc}
\toprule
Model & top-1 & mAP & $\mu$AP & F1@0.5 & $\mu$F1@0.5 \\
\midrule
aa & 0.441 $\pm$ 0.020 & 0.157 $\pm$ 0.010 & 0.242 $\pm$ 0.010 & 0.018 $\pm$ 0.009 & 0.054 $\pm$ 0.021 \\
3Di\_tokens & 0.487 $\pm$ 0.019 & 0.264 $\pm$ 0.013 & 0.345 $\pm$ 0.018 & 0.092 $\pm$ 0.020 & 0.178 $\pm$ 0.036 \\
Random & 0.399 $\pm$ 0.026 & 0.134 $\pm$ 0.003 & 0.239 $\pm$ 0.005 & 0.002 $\pm$ 0.001 & 0.007 $\pm$ 0.004 \\
AminoAseed & \textbf{0.538 $\pm$ 0.017} & \textbf{0.359 $\pm$ 0.018} & \textbf{0.431 $\pm$ 0.025} & \textbf{0.200 $\pm$ 0.019} & \textbf{0.314 $\pm$ 0.035} \\
ProToken & 0.417 $\pm$ 0.019 & 0.168 $\pm$ 0.010 & 0.243 $\pm$ 0.011 & 0.028 $\pm$ 0.010 & 0.057 $\pm$ 0.020 \\
ESM3struct & 0.474 $\pm$ 0.020 & 0.261 $\pm$ 0.016 & 0.324 $\pm$ 0.017 & 0.099 $\pm$ 0.017 & 0.177 $\pm$ 0.028 \\
\tool, $P{=}1$ & \underline{0.518 $\pm$ 0.032} & \underline{0.287 $\pm$ 0.022} & \underline{0.370 $\pm$ 0.032} & \underline{0.127 $\pm$ 0.021} & \underline{0.218 $\pm$ 0.033} \\
\midrule
ProtProfileMD\_K8 & 0.479 $\pm$ 0.018 & 0.241 $\pm$ 0.015 & 0.304 $\pm$ 0.021 & 0.072 $\pm$ 0.016 & 0.145 $\pm$ 0.028 \\
Vote\_3Di & \underline{0.502 $\pm$ 0.021} & \underline{0.283 $\pm$ 0.014} & \underline{0.364 $\pm$ 0.011} & \underline{0.109 $\pm$ 0.020} & \underline{0.204 $\pm$ 0.036} \\
\tool, $P{=}\text{full}$ & \textbf{0.520 $\pm$ 0.026} & \textbf{0.284 $\pm$ 0.028} & \textbf{0.381 $\pm$ 0.046} & \textbf{0.120 $\pm$ 0.038} & \textbf{0.224 $\pm$ 0.064} \\
\bottomrule
\end{tabular}
\end{table}

\newpage
\subsection{Binding Site}

We report residue-level binding-site classification under all three
splits.

\paragraph{structure split}\mbox{}\\
\begin{table}[htbp]
\centering\small
\caption{Binding site, structure split.}
\begin{tabular}{lcc}
\toprule
Model & AUROC & AP \\
\midrule
aa & 0.595 $\pm$ 0.007 & 0.137 $\pm$ 0.003 \\
3Di\_tokens & 0.610 $\pm$ 0.007 & 0.140 $\pm$ 0.002 \\
Random & 0.503 $\pm$ 0.001 & 0.101 $\pm$ 0.000 \\
AminoAseed & \underline{0.649 $\pm$ 0.006} & \underline{0.168 $\pm$ 0.007} \\
ProToken & \textbf{0.689 $\pm$ 0.005} & \textbf{0.197 $\pm$ 0.005} \\
ESM3struct & 0.644 $\pm$ 0.008 & 0.165 $\pm$ 0.006 \\
\tool, $P{=}1$ & 0.630 $\pm$ 0.012 & 0.150 $\pm$ 0.007 \\
\midrule
ProtProfileMD\_K8 & 0.608 $\pm$ 0.015 & \underline{0.142 $\pm$ 0.005} \\
Vote\_3Di & \underline{0.610 $\pm$ 0.009} & 0.140 $\pm$ 0.002 \\
\tool, $P{=}\text{full}$ & \textbf{0.625 $\pm$ 0.012} & \textbf{0.147 $\pm$ 0.006} \\
\bottomrule
\end{tabular}
\end{table}

\paragraph{sequence split}\mbox{}\\
\begin{table}[htbp]
\centering\small
\caption{Binding site, sequence split.}
\begin{tabular}{lcc}
\toprule
Model & AUROC & AP \\
\midrule
aa & 0.594 $\pm$ 0.004 & 0.134 $\pm$ 0.002 \\
3Di\_tokens & 0.648 $\pm$ 0.004 & 0.161 $\pm$ 0.003 \\
Random & 0.496 $\pm$ 0.001 & 0.097 $\pm$ 0.000 \\
AminoAseed & 0.696 $\pm$ 0.008 & \underline{0.203 $\pm$ 0.009} \\
ProToken & \textbf{0.722 $\pm$ 0.005} & \textbf{0.229 $\pm$ 0.005} \\
ESM3struct & 0.673 $\pm$ 0.006 & 0.180 $\pm$ 0.007 \\
\tool, $P{=}1$ & \underline{0.702 $\pm$ 0.005} & 0.197 $\pm$ 0.005 \\
\midrule
ProtProfileMD\_K8 & \underline{0.656 $\pm$ 0.007} & \underline{0.164 $\pm$ 0.006} \\
Vote\_3Di & 0.646 $\pm$ 0.006 & 0.159 $\pm$ 0.004 \\
\tool, $P{=}\text{full}$ & \textbf{0.697 $\pm$ 0.005} & \textbf{0.193 $\pm$ 0.008} \\
\bottomrule
\end{tabular}
\end{table}

\paragraph{random split}\mbox{}\\
\begin{table}[htbp]
\centering\small
\caption{Binding site, random split.}
\begin{tabular}{lcc}
\toprule
Model & AUROC & AP \\
\midrule
aa & \textbf{0.909 $\pm$ 0.002} & \textbf{0.739 $\pm$ 0.003} \\
3Di\_tokens & 0.821 $\pm$ 0.002 & 0.561 $\pm$ 0.003 \\
Random & 0.505 $\pm$ 0.002 & 0.135 $\pm$ 0.001 \\
AminoAseed & \underline{0.908 $\pm$ 0.001} & \underline{0.727 $\pm$ 0.001} \\
ProToken & 0.898 $\pm$ 0.001 & 0.707 $\pm$ 0.002 \\
ESM3struct & 0.904 $\pm$ 0.002 & 0.721 $\pm$ 0.003 \\
\tool, $P{=}1$ & 0.902 $\pm$ 0.002 & 0.709 $\pm$ 0.003 \\
\midrule
ProtProfileMD\_K8 & \underline{0.889 $\pm$ 0.002} & \underline{0.687 $\pm$ 0.003} \\
Vote\_3Di & 0.843 $\pm$ 0.001 & 0.602 $\pm$ 0.003 \\
\tool, $P{=}\text{full}$ & \textbf{0.912 $\pm$ 0.001} & \textbf{0.733 $\pm$ 0.003} \\
\bottomrule
\end{tabular}
\end{table}

\newpage
\subsection{Binding Affinity}

We report binding-affinity regression on misato (label: $-\log K_d/K_i$)
under all three splits. Each prediction conditions on a $167$-bit MACCS structural-key
fingerprint of the bound ligand (resolved from the MISATO \texttt{ligand\_id} via the PDB Chemical Component Dictionary, or from the PDBbind ligand file for $31$ complexes), in
addition to the per-residue protein representation. All numbers below are in the format of mean $\pm$ sample std over $10$ seeds. Bold = best,
underline = second-best, computed within each half per column.
Higher is better for $R^2$ and Spearman; \textbf{lower is better for
MSE}.

\paragraph{structure split}\mbox{}\\
\begin{table}[htbp]
\centering\small
\caption{Binding affinity ($-\log K_d/K_i$), structure split, ligand aware head (MACCS-167), mean $\pm$ sample std over $10$ seeds.}
\begin{tabular}{lccc}
\toprule
Model & $R^2$ & Spearman & MSE \\
\midrule
aa & 0.032 $\pm$ 0.061 & 0.277 $\pm$ 0.141 & 2.925 $\pm$ 0.185 \\
3Di\_tokens & $-$0.057 $\pm$ 0.090 & 0.231 $\pm$ 0.149 & 3.193 $\pm$ 0.272 \\
Random & 0.074 $\pm$ 0.032 & \textbf{0.390 $\pm$ 0.007} & 2.799 $\pm$ 0.098 \\
AminoAseed & 0.009 $\pm$ 0.055 & 0.236 $\pm$ 0.137 & 2.994 $\pm$ 0.166 \\
ProToken & 0.048 $\pm$ 0.054 & 0.358 $\pm$ 0.032 & 2.875 $\pm$ 0.162 \\
ESM3struct & \textbf{0.097 $\pm$ 0.063} & \underline{0.382 $\pm$ 0.034} & \textbf{2.730 $\pm$ 0.190} \\
\tool, $P{=}1$ & \underline{0.093 $\pm$ 0.027} & 0.372 $\pm$ 0.021 & \underline{2.742 $\pm$ 0.083} \\
\midrule
ProtProfileMD\_K8 & \underline{$-$0.019 $\pm$ 0.072} & \underline{0.317 $\pm$ 0.028} & \underline{3.081 $\pm$ 0.219} \\
Vote\_3Di & $-$0.055 $\pm$ 0.101 & 0.226 $\pm$ 0.158 & 3.187 $\pm$ 0.306 \\
\tool, $P{=}\text{full}$ & \textbf{0.086 $\pm$ 0.035} & \textbf{0.375 $\pm$ 0.023} & \textbf{2.762 $\pm$ 0.105} \\
\bottomrule
\end{tabular}
\end{table}

\newpage
\paragraph{sequence split}\mbox{}\\
\begin{table}[htbp]
\centering\small
\caption{Binding affinity ($-\log K_d/K_i$), sequence split, ligand aware head (MACCS-167), mean $\pm$ sample std over $10$ seeds.}
\begin{tabular}{lccc}
\toprule
Model & $R^2$ & Spearman & MSE \\
\midrule
aa & 0.124 $\pm$ 0.075 & 0.376 $\pm$ 0.095 & 2.757 $\pm$ 0.237 \\
3Di\_tokens & \textbf{0.179 $\pm$ 0.048} & \textbf{0.457 $\pm$ 0.012} & \textbf{2.584 $\pm$ 0.151} \\
Random & 0.124 $\pm$ 0.071 & \underline{0.455 $\pm$ 0.015} & 2.758 $\pm$ 0.223 \\
AminoAseed & 0.130 $\pm$ 0.051 & 0.425 $\pm$ 0.031 & 2.739 $\pm$ 0.162 \\
ProToken & 0.124 $\pm$ 0.078 & 0.391 $\pm$ 0.069 & 2.757 $\pm$ 0.244 \\
ESM3struct & \underline{0.156 $\pm$ 0.033} & 0.434 $\pm$ 0.020 & \underline{2.656 $\pm$ 0.102} \\
\tool, $P{=}1$ & 0.142 $\pm$ 0.065 & 0.416 $\pm$ 0.027 & 2.700 $\pm$ 0.206 \\
\midrule
ProtProfileMD\_K8 & \underline{0.175 $\pm$ 0.057} & \underline{0.451 $\pm$ 0.028} & \underline{2.599 $\pm$ 0.180} \\
Vote\_3Di & \textbf{0.178 $\pm$ 0.033} & \textbf{0.455 $\pm$ 0.007} & \textbf{2.587 $\pm$ 0.105} \\
\tool, $P{=}\text{full}$ & 0.141 $\pm$ 0.075 & 0.415 $\pm$ 0.033 & 2.706 $\pm$ 0.237 \\
\bottomrule
\end{tabular}
\end{table}

\paragraph{random split}\mbox{}\\
\begin{table}[htbp]
\centering\small
\caption{Binding affinity ($-\log K_d/K_i$), random split, ligand aware head (MACCS-167), mean $\pm$ sample std over $10$ seeds.}
\begin{tabular}{lccc}
\toprule
Model & $R^2$ & Spearman & MSE \\
\midrule
aa & \underline{0.516 $\pm$ 0.020} & \textbf{0.728 $\pm$ 0.009} & \underline{1.932 $\pm$ 0.080} \\
3Di\_tokens & 0.496 $\pm$ 0.019 & 0.699 $\pm$ 0.013 & 2.011 $\pm$ 0.074 \\
Random & 0.420 $\pm$ 0.014 & 0.645 $\pm$ 0.006 & 2.314 $\pm$ 0.055 \\
AminoAseed & 0.501 $\pm$ 0.017 & 0.706 $\pm$ 0.010 & 1.990 $\pm$ 0.067 \\
ProToken & 0.486 $\pm$ 0.017 & 0.695 $\pm$ 0.012 & 2.051 $\pm$ 0.069 \\
ESM3struct & 0.500 $\pm$ 0.021 & 0.699 $\pm$ 0.011 & 1.997 $\pm$ 0.084 \\
\tool, $P{=}1$ & \textbf{0.521 $\pm$ 0.011} & \underline{0.724 $\pm$ 0.012} & \textbf{1.913 $\pm$ 0.046} \\
\midrule
ProtProfileMD\_K8 & \underline{0.507 $\pm$ 0.019} & \underline{0.710 $\pm$ 0.012} & \underline{1.965 $\pm$ 0.076} \\
Vote\_3Di & 0.497 $\pm$ 0.013 & 0.706 $\pm$ 0.010 & 2.008 $\pm$ 0.051 \\
\tool, $P{=}\text{full}$ & \textbf{0.510 $\pm$ 0.021} & \textbf{0.715 $\pm$ 0.010} & \textbf{1.955 $\pm$ 0.086} \\
\bottomrule
\end{tabular}
\end{table}

\begin{table}[htbp]
\centering\small
\caption{Binding affinity controls, mean $\pm$ sample std over $10$ seeds. Ligand only: MLP on the MACCS-167 fingerprint; protein only: the affinity head without the fingerprint; protein + MACCS: the rows of the affinity tables. Not ranked.}
\label{tab:affinity_controls}
\begin{tabular}{llccc}
\toprule
Split & Input & $R^2$ & Spearman & MSE \\
\midrule
sequence & ligand only (MACCS) & 0.091 $\pm$ 0.043 & 0.443 $\pm$ 0.012 & 2.861 $\pm$ 0.136 \\
sequence & ESM3struct, protein only & $-$0.090 $\pm$ 0.088 & 0.043 $\pm$ 0.037 & 3.432 $\pm$ 0.278 \\
sequence & ESM3struct, protein + MACCS & 0.156 $\pm$ 0.033 & 0.434 $\pm$ 0.020 & 2.656 $\pm$ 0.102 \\
sequence & aa, protein only & $-$0.053 $\pm$ 0.037 & 0.048 $\pm$ 0.078 & 3.315 $\pm$ 0.115 \\
sequence & aa, protein + MACCS & 0.124 $\pm$ 0.075 & 0.376 $\pm$ 0.095 & 2.757 $\pm$ 0.237 \\
sequence & \tool, $P{=}1$, protein only & $-$0.087 $\pm$ 0.053 & 0.035 $\pm$ 0.048 & 3.423 $\pm$ 0.168 \\
sequence & \tool, $P{=}1$, protein + MACCS & 0.142 $\pm$ 0.065 & 0.416 $\pm$ 0.027 & 2.700 $\pm$ 0.206 \\
sequence & \tool, $P{=}\text{full}$, protein only & $-$0.104 $\pm$ 0.033 & 0.042 $\pm$ 0.047 & 3.477 $\pm$ 0.103 \\
sequence & \tool, $P{=}\text{full}$, protein + MACCS & 0.141 $\pm$ 0.075 & 0.415 $\pm$ 0.033 & 2.706 $\pm$ 0.237 \\
\midrule
structure & ligand only (MACCS) & 0.072 $\pm$ 0.021 & 0.360 $\pm$ 0.014 & 2.805 $\pm$ 0.062 \\
structure & ESM3struct, protein only & $-$0.065 $\pm$ 0.065 & 0.126 $\pm$ 0.060 & 3.217 $\pm$ 0.195 \\
structure & ESM3struct, protein + MACCS & 0.097 $\pm$ 0.063 & 0.382 $\pm$ 0.034 & 2.730 $\pm$ 0.190 \\
structure & aa, protein only & $-$0.049 $\pm$ 0.054 & 0.063 $\pm$ 0.154 & 3.170 $\pm$ 0.164 \\
structure & aa, protein + MACCS & 0.032 $\pm$ 0.061 & 0.277 $\pm$ 0.141 & 2.925 $\pm$ 0.185 \\
structure & \tool, $P{=}1$, protein only & $-$0.066 $\pm$ 0.045 & 0.083 $\pm$ 0.064 & 3.220 $\pm$ 0.136 \\
structure & \tool, $P{=}1$, protein + MACCS & 0.093 $\pm$ 0.027 & 0.372 $\pm$ 0.021 & 2.742 $\pm$ 0.083 \\
structure & \tool, $P{=}\text{full}$, protein only & $-$0.102 $\pm$ 0.056 & 0.026 $\pm$ 0.052 & 3.329 $\pm$ 0.169 \\
structure & \tool, $P{=}\text{full}$, protein + MACCS & 0.086 $\pm$ 0.035 & 0.375 $\pm$ 0.023 & 2.762 $\pm$ 0.105 \\
\midrule
random & ligand only (MACCS) & 0.468 $\pm$ 0.016 & 0.689 $\pm$ 0.008 & 2.121 $\pm$ 0.065 \\
random & ESM3struct, protein only & 0.285 $\pm$ 0.015 & 0.532 $\pm$ 0.012 & 2.852 $\pm$ 0.061 \\
random & ESM3struct, protein + MACCS & 0.500 $\pm$ 0.021 & 0.699 $\pm$ 0.011 & 1.997 $\pm$ 0.084 \\
random & aa, protein only & 0.344 $\pm$ 0.018 & 0.600 $\pm$ 0.009 & 2.618 $\pm$ 0.071 \\
random & aa, protein + MACCS & 0.516 $\pm$ 0.020 & 0.728 $\pm$ 0.009 & 1.932 $\pm$ 0.080 \\
random & \tool, $P{=}1$, protein only & 0.329 $\pm$ 0.019 & 0.568 $\pm$ 0.015 & 2.677 $\pm$ 0.077 \\
random & \tool, $P{=}1$, protein + MACCS & 0.521 $\pm$ 0.011 & 0.724 $\pm$ 0.012 & 1.913 $\pm$ 0.046 \\
random & \tool, $P{=}\text{full}$, protein only & 0.322 $\pm$ 0.021 & 0.568 $\pm$ 0.019 & 2.703 $\pm$ 0.086 \\
random & \tool, $P{=}\text{full}$, protein + MACCS & 0.510 $\pm$ 0.021 & 0.715 $\pm$ 0.010 & 1.955 $\pm$ 0.086 \\
\bottomrule
\end{tabular}
\end{table}

\newpage
\section{Ablations}
\label{app:ablations}

\subsection{Quantizer: RVQ versus VQ}
\label{app:ablation-vq}
RVQ (three levels, $2048{\times}128{\times}128$) versus a single $2048$-entry codebook,
both trained with $\lambda = 0.1$ on the same corpus and schedule. Only
first-level tokens are used downstream in both cases. The RVQ reconstructs
from the sum of its three quantizer levels, so the reconstruction columns
are not directly comparable.

\begin{table}[htbp]
\centering\small
\begin{tabular}{lccccc}
\toprule
Quantizer & Recon $P{=}1$ & Recon $P{=}10$ & L1 ids (util) & L1 ppl & mdCATH $P{=}\text{full}$ \\
\midrule
\multicolumn{6}{l}{\textit{$P = \text{full}$ (10 frames)}} \\
RVQ $2048{\times}128{\times}128$ & --- & --- & --- & --- & \textbf{0.628 $\pm$ 0.001} \\
VQ-2048              & --- & --- & --- & --- & 0.613 $\pm$ 0.001 \\
\midrule
\multicolumn{6}{l}{\textit{$P = 1$ (single frame)}} \\
RVQ $2048{\times}128{\times}128$ & \textbf{6.796} & \textbf{7.254} & 2{,}048 (100\%) & \textbf{1{,}653} & \textbf{0.521 $\pm$ 0.001} \\
VQ-2048              & 7.137 & 7.425 & 2{,}048 (100\%) & 1{,}585 & 0.515 $\pm$ 0.001 \\
\bottomrule
\end{tabular}
\caption{RVQ vs VQ ablation ($\lambda = 0.1$).
Recon: validation reconstruction loss (lower is better).
RMSF: test Spearman (pooled), mean $\pm$ sample std over $10$ seeds.
Bold: best per column.}
\label{tab:rvq_vs_vq}
\end{table}

\subsection{Single frame to token distillation}
\label{app:ablation-sftd}
The full SFTD ablation (distillation off, SFTD off, and a sweep over $\lambda$) is given in Table~\ref{tab:sftd_ablation_table}.
Distillation off ($\lambda = 0$) keeps both branches of
\eqref{eq:sftd} and only drops the distillation term; SFTD off trains a single
branch on all $10$ frames. We choose among trained tokenizers, including $\lambda$, on
validation numbers only: the validation reconstruction loss and the RMSF
Spearman on mdCATH-div and BioEMU CATH(pooled and per protein, at $P{=}1$ and $P{=}\text{full}$); functional signals play no part in the choice.
We use $\lambda = 10$ as a balance between reconstruction and dynamics.
Raising $\lambda$ strengthens the RMSF signal, but from $\lambda = 50$ on it degrades
reconstruction and collapses the codebook; $\lambda = 10$ keeps the reconstruction
loss close to the best runs and the codebook almost fully used ($99.8\%$),
while giving the strongest RMSF signal among the settings without collapse.

\begin{table}[htbp]
\centering\small
\resizebox{\linewidth}{!}{%
\begin{tabular}{lccccccc}
\toprule
$\lambda$ & Best/stop ep & Recon $P{=}1$ & Recon $P{=}10$ & L1 ids at $P{=}1$ (util) & L1 ppl ($P{=}1$) & mdCATH $P{=}\text{full}$ & BioEmu $P{=}\text{full}$ \\
\midrule
SFTD off & 120/160 & 8.975 & 7.239 & 51 (2\%) & 5.1 & 0.633 & 0.683 \\
0        & 205/245 & 6.774 & 7.243 & \textbf{2{,}048 (100\%)} & 1{,}643 & 0.632 & 0.682 \\
0.05     & 125/165 & 6.799 & 7.257 & \textbf{2{,}048 (100\%)} & 1{,}630 & 0.608 & 0.661 \\
0.1      & 110/150 & 6.796 & 7.254 & \textbf{2{,}048 (100\%)} & 1{,}653 & 0.628 & 0.676 \\
0.25     & 85/125  & 6.804 & 7.256 & \textbf{2{,}048 (100\%)} & \textbf{1{,}775} & 0.625 & 0.673 \\
0.5      & 320/360 & \textbf{6.719} & \textbf{7.209} & \textbf{2{,}048 (100\%)} & 1{,}637 & 0.619 & 0.668 \\
1        & 355/395 & 6.767 & 7.222 & \textbf{2{,}048 (100\%)} & 1{,}628 & 0.620 & 0.668 \\
5        & 150/190 & 6.864 & 7.237 & 2{,}047 (100\%) & 1{,}635 & 0.648 & 0.698 \\
10       & 435/475 & 6.882 & 7.225 & 2{,}043 (99.8\%) & 1{,}640 & 0.653 & 0.711 \\
50       & 40/80   & 7.133 & 7.411 & 1{,}900 (93\%)  & 967  & \textbf{0.687} & \textbf{0.738} \\
100      & 35/75   & 7.202 & 7.451 & 966 (47\%)      & 430  & 0.666 & 0.712 \\
500      & 10/50   & 8.212 & 8.346 & 1 (0\%)         & 1.0  & --- & --- \\
\bottomrule
\end{tabular}}
\caption{SFTD ablation ($\lambda$ sweep). Recon and codebook columns: validation set; RMSF column:
probe Spearman at $P{=}\text{full}$ on the held-out mdCATH-div test set, shown for reference (not used for selection).
Bold: best in each column (Recon: lower is better; ties all bold).
SFTD enables $P{=}1$ tokenization: SFTD off gives mdCATH-div $P{=}1$ Spearman $0.214$; $\lambda = 0$ (SFTD
training with two branches but no consistency loss) gives $0.521$.
$P{=}\text{full}$ RMSF dips slightly for $0 < \lambda \le 1$, then rises for $\lambda \ge 5$ and peaks at
$\lambda = 50$; $\lambda \ge 50$ causes codebook collapse (93\% utilisation at $\lambda = 50$,
47\% at $\lambda = 100$, 0\% at $\lambda = 500$).
We use $\lambda = 10$ (near-full utilisation, $2{,}043/2{,}048 = 99.8\%$; see text).}

\label{tab:sftd_ablation_table}
\end{table}

\newpage
\section{Cryptic Pocket Prediction (CryptoBench)}
\label{sec:cryptobench}

Cryptic pockets are binding sites that are absent in the apo crystal
structure but become accessible through conformational change.
CryptoBench~\citep{CryptoBench} provides apo structures of proteins with
known cryptic pockets and per-residue binary labels indicating whether a
residue lines a pocket that opens in at least one holo binding partner.
We use it to probe how much information about cryptic pocket residues each
representation carries.
This is a probe of representational content, not a proposal for a
cryptic-site predictor:
an improvement at a single input frame shows the tokens encode more of the
signal relevant to pocket formation, not that training on dynamics is the
sole cause of the gain (no model trained on a single frame of the same
corpus is available for comparison).

\paragraph{Benchmark and labels.}
We use all 1{,}276 chains of the CryptoBench release,
excluding 5 chains with fewer than 16 residues (the
minimum length of the structural descriptor).
Modified polymer residues (MLY, LLP, SEP and others) and residues without a complete backbone (N, C$\alpha$, C) are dropped.
Per-residue binary labels mark the union of apo pocket residues across
all holo binding partners for each chain, matched on author residue
number and insertion code.
The test set comprises 222 apo entries (261 chains).

\paragraph{Probe.}
A three-layer MLP (input dimension $\!\to\!256\!\to\!256\!\to\!2$, ReLU,
dropout 0.3) with cross-entropy loss (class weights $[1, n_\text{neg}/n_\text{pos}]$),
Adam lr $10^{-4}$, wd $10^{-5}$, batch 4{,}096 residues, $\le 80$ epochs,
patience 15 on the validation AUROC averaged over entries.
Inputs standardised with training mean/sd. Five seeds (0--4).
This probe deviates from the upstream pLM-NN protocol of CryptoBench in
several ways (training folds, epoch selection criterion, regularisation,
batch size, standardisation; see the code release for details).

\paragraph{Results.}
The primary readout for \tool is the discrete level-0 token
(Table~\ref{tab:cryptobench}).

\begin{table}[htbp]
\centering\small
\begin{tabular}{lcc}
\toprule
Method & MLP AUROC (per-entry) & MLP AP (per-entry) \\
\midrule
aa                    & 0.594 $\pm$ 0.001 & 0.112 $\pm$ 0.000 \\
3Di                   & 0.559 $\pm$ 0.000 & 0.091 $\pm$ 0.001 \\
AminoAseed            & 0.621 $\pm$ 0.001 & 0.138 $\pm$ 0.002 \\
ESM3struct            & 0.596 $\pm$ 0.001 & 0.127 $\pm$ 0.001 \\
ProToken              & 0.542 $\pm$ 0.000 & 0.098 $\pm$ 0.001 \\
\textbf{\tool{} $\lambda{=}10$} & \textbf{0.644 $\pm$ 0.002} & \textbf{0.165 $\pm$ 0.002} \\
\midrule
ESM2-3B (descriptive) & 0.828 $\pm$ 0.001 & 0.409 $\pm$ 0.002 \\
\bottomrule
\end{tabular}
\caption{CryptoBench MLP probe on 222 apo test entries.
\tool $\lambda{=}10$ tokens carry more cryptic-site signal than any
static tokenizer (0.644 vs AminoAseed 0.621).
The token-only MLP probe is not competitive with continuous ESM2 embeddings
(0.644 vs 0.828), as expected for a discrete representation.
ESM2-3B is descriptive only (last hidden layer, not in any significance family).}
\label{tab:cryptobench}
\end{table}

\newpage
\section{Same-fold-different-dynamics test}
\label{sec:fold_pairs}

\paragraph{Data.}
We mine pairs from BioEmu-CATH ($1{,}090$ systems) that share a fold but differ in dynamics.
All $141{,}630$ length-compatible pairs (length tolerance $15\%$, excluding same-domain pairs)
are scored by TM-align on the medoid frame (the frame closest to the run-1 mean C$\alpha$).
A pair passes the fold gate if $\mathrm{TM}_\mathrm{max} \ge 0.7$ over at least $50$ aligned
residues and sequence identity $\le 30\%$ over the aligned positions.
For each protein the PC1 of its C$\alpha$ covariance across MD frames must reproduce across
disjoint groups of independent runs (cosine similarity $\ge 0.6$).

\paragraph{Sets.}
The \emph{divergent} set (top $25$ candidates by ascending PC1 cosine)
requires $\cos(\mathrm{PC1}_A, \mathrm{PC1}_B) \le 0.30$; $19$ of $25$ pass verification.
The \emph{similar} control (top $50$ candidates by descending cosine) requires
$\cos(\mathrm{PC1}_A, \mathrm{PC1}_B) \ge 0.70$ and covariance overlap over $10$ modes $\ge 0.5$;
$3$ of $50$ pass. Selection is tokenizer-free.

\paragraph{Test statistic.}
For each pair and tokenizer we measure, at the TM-aligned positions of the two medoid structures,
the chance-corrected agreement
$\kappa = (p_\text{obs} - p_\text{chance}) / (1 - p_\text{chance})$,
where $p_\text{chance} = \sum_t f(t)^2$ is the empirical collision probability estimated from
the marginal token frequencies over a floor population of $1{,}090$ BioEmu-CATH systems (not $1/M$).
The contrast is $\Delta = \overline{\kappa}_\text{similar} - \overline{\kappa}_\text{divergent}$.
A dynamics-faithful tokenizer gives a positive $\Delta$; a fold-only tokenizer gives $\Delta \approx 0$.

\paragraph{Null and $p$-value.}
We bootstrap ($B{=}10{,}000$, seed $0$) by resampling the similar and divergent sets independently
with replacement, computing $\Delta^*_b$ for each resample.
The $p$-value is the fraction of resamples with $\Delta^*_b \le 0$ (one-sided, dynamics-positive hypothesis).
The $95\%$ CI is the percentile interval of the bootstrap distribution.
No multiple-comparison correction is applied (single pre-specified primary analysis per tokenizer).

\paragraph{Result.}
\tool $P{=}\text{full}$ is the only tokenizer whose $\Delta$ is significantly positive
(Table~\ref{tab:fold_pairs}).

\begin{table}[htbp]
\centering\small
\begin{tabular}{lcccc}
\toprule
Method & $\overline{\kappa}$ (similar) & $\overline{\kappa}$ (divergent) & $\Delta$ [95\% CI] & $p$ \\
\midrule
\tool{} $P{=}\text{full}$ & 0.236 & 0.179 & 0.056 [0.015, 0.095] & 0.004 \\
3Di & 0.412 & 0.418 & $-$0.006 [$-$0.101, 0.076] & 0.538 \\
AminoAseed & 0.171 & 0.138 & 0.033 [$-$0.026, 0.084] & 0.136 \\
ESM3-struct & 0.016 & 0.014 & 0.002 [$-$0.014, 0.017] & 0.408 \\
ProToken & 0.037 & 0.042 & $-$0.004 [$-$0.045, 0.038] & 0.585 \\
\tool{} $P{=}1$ & 0.187 & 0.160 & 0.027 [$-$0.020, 0.070] & 0.122 \\
\bottomrule
\end{tabular}
\caption{Same-fold-different-dynamics test ($n_\text{similar}{=}3$, $n_\text{divergent}{=}19$;
$B{=}10{,}000$ bootstrap resamples).
$\Delta$ vs baselines (\tool $P{=}\text{full}$ minus each baseline's $\Delta$):
vs 3Di: $+0.062$ [$-0.021$, $+0.160$], $p = 0.096$;
vs AminoAseed: $+0.023$ [$-0.032$, $+0.087$], $p = 0.252$;
vs ESM3struct: $+0.054$ [$+0.013$, $+0.093$], $p = 0.006$;
vs ProToken: $+0.061$ [$+0.022$, $+0.096$], $p = 0.001$.}
\label{tab:fold_pairs}
\end{table}

\newpage

\section{Significance testing}
\label{app:significance}

\paragraph{Protocol.} Each comparison is the difference $\Delta$ of a metric between \tool and a
baseline on the same test items, computed from the predictions averaged over the $10$ probe seeds.
Confidence intervals are $95\%$ percentile intervals of a paired cluster bootstrap ($B = 100{,}000$,
seed $20261019$) in which a drawn cluster brings all its proteins. Clusters are CATH topologies on
mdCATH-div ($153$ clusters among $245$ test domains) and BioEmu-CATH ($38$ among $100$ test systems),
and the sequence or structure groups of the split on MISATO ($127$ and $91$ among $489$ test
complexes). For EC and GO, $\Delta$ is computed per cross-validation fold with the groups of the
split as clusters, and the mean over the five folds is tested. Two sided $p$ values come from the
centred bootstrap, $p = (1 + \#\{b : |\Delta^*_b - \Delta| \ge |\Delta|\})/(B + 1)$, and are Holm
adjusted within each pre-specified family.

\paragraph{T-RMSF ($56$ comparisons).} \tool, $P{=}1$ against aa, 3Di, AminoAseed, ESM3struct and
ProToken, and \tool, $P{=}\text{full}$ against ProtProfileMD\_K and Vote\_3Di, on the four datasets,
for the pooled and the per-protein Spearman correlation. All $56$ differences favour \tool with
Holm $p = 5.6\times10^{-4}$, the smallest value attainable with $B = 100{,}000$ and $56$ tests. The
smallest are $P{=}\text{full}$ against ProtProfileMD\_K on mdCATH-div (pooled $\Delta = +0.031$,
CI $[+0.017, +0.046]$) and $P{=}1$ against AminoAseed on MISATO (pooled $\Delta = +0.077$
$[+0.056, +0.100]$ on the sequence split and $+0.076$ $[+0.052, +0.104]$ on the structure split).

\paragraph{T-Function ($55$ comparisons).} \tool, $P{=}1$ against the same five single-frame
baselines on EC (depth $3$) and GO (sequence and structure splits; protein centric AP and macro
mAP), binding site (structure split; pooled and per-protein AUROC) and affinity (structure split;
Spearman). $13$ of the $55$ are significant at Holm $p < 0.05$. On the sequence split \tool is
better than aa on EC and GO, than ProToken on EC protein AP and on GO, than ESM3struct on GO and
than 3Di on GO mAP, and worse than AminoAseed on EC and GO mAP ($\Delta = -0.092$ and $-0.064$,
Holm $p = 5.5\times10^{-4}$ and $0.014$). On the structure split the only significant difference is
ProToken over \tool on binding (pooled AUROC $\Delta = -0.057$ $[-0.084, -0.029]$, Holm $p = 0.005$).
No affinity comparison is significant.

\newpage
\section{Static tokenizers retrained on our data}
\label{app:static_retrain}

The main-text results compare against published tokenizer checkpoints, each trained
on different data (ESM3: PDB-scale; AminoAseed: CATH; ProToken: PDB; \tool: BioEmu-CATH + MISATO).
To control for this confound, we retrained AminoAseed and ESM3-struct from scratch on the
same BioEmu-CATH + MISATO corpus ($18{,}044$ proteins; identical train/val/test split as \tool)
using their published training code and hyperparameters without modification,
selecting the best checkpoint by validation backbone RMSD
(AminoAseed: $2.717$\,\AA{} at epoch~9; ESM3-style: $2.115$\,\AA{} at epoch~9;
$10$ epochs, $360{,}900$ steps, $1$ GPU each, ${\sim}71$ GPU-hours total).

Tables~\ref{tab:retrained_rmsf} and \ref{tab:retrained_function} compare \tool $\lambda{=}10$ ($P{=}1$, single frame) against both retrained baselines on every downstream task.
All three models are evaluated with the same test proteins, labels, probes and seeds ($10$ probe seeds, $5$-fold CV where applicable).

\paragraph{Dynamics (RMSF).}
Even at $P{=}1$, \tool leads on all three RMSF evaluation corpora by a wide margin.
Retraining improves ESM3-struct substantially
(e.g.\ per-protein Spearman on mdCATH-div: $0.543$ vs $0.460$ published)
while AminoAseed stays comparable.
All $12$ pairwise comparisons (two baselines $\times$ three datasets $\times$ pooled and per-protein Spearman)
favour \tool with Holm-corrected $p < 0.01$ (per-protein cluster bootstrap, $B{=}10{,}000$;
family of $12$ comparisons; the smallest effect is $\Delta = {+}0.047$ on MISATO per-protein against ESM3-struct retrained).

\paragraph{Function (EC, GO, binding, affinity).}
On EC depths $2$ and $3$ and on GO, the retrained static tokenizers outperform \tool ($P{=}1$):
ESM3-struct (retrained) leads EC depth $2$ ($0.215$, with AminoAseed retrained at $0.214$) and GO macro-AP ($0.375$),
and AminoAseed (retrained) leads EC depth $3$ ($0.159$).
On EC depth $1$, \tool matches ESM3-struct retrained ($0.272$ each) and is ahead of AminoAseed retrained ($0.264$).
On binding-site AUROC the three methods are comparable (ESM3-struct retrained $0.704$ vs \tool $0.702$);
affinity Spearman is likewise close ($0.409$/$0.426$/$0.416$).

\paragraph{ProteinGym.}
On the $96$ assays ($L \le 200$), ESM3-struct retrained ($0.358$) leads the three-way comparison,
with \tool ($0.347$) second and AminoAseed retrained ($0.337$) third.

\paragraph{CryptoBench.}
\tool $\lambda{=}10$ leads the cryptic-pocket MLP probe on both per-entry AUROC ($0.644$) and AP ($0.165$),
followed by AminoAseed retrained ($0.619$, $0.131$) and ESM3-struct retrained ($0.619$, $0.136$).

\begin{table}[htbp]
\centering\small
\begin{tabular}{lcccccc}
\toprule
 & \multicolumn{3}{c}{Pooled Spearman} & \multicolumn{3}{c}{Per-protein Spearman} \\
\cmidrule(lr){2-4}\cmidrule(lr){5-7}
Method & mdCATH & MISATO & BioEmu & mdCATH & MISATO & BioEmu \\
\midrule
AminoAseed (retrained)  & 0.426  & 0.447  & 0.437  & 0.523  & 0.595  & 0.531 \\
ESM3-struct (retrained)  & 0.451  & 0.455  & 0.458  & 0.543  & 0.608  & 0.561 \\
\tool{} $\lambda{=}10$, $P{=}1$ & \textbf{0.539} & \textbf{0.513} & \textbf{0.574} & \textbf{0.642} & \textbf{0.654} & \textbf{0.678} \\
\midrule
\tool{} $\lambda{=}10$, $P{=}\text{full}$ & 0.653 & 0.517 & 0.711 & 0.716 & 0.658 & 0.775 \\
\bottomrule
\end{tabular}
\caption{RMSF Spearman: retrained static tokenizers and \tool.
The $P{=}\text{full}$ row is the full multi-frame comparison (not included in the statistical tests).
Bold: best among the three single-frame rows.
All $12$ \tool $P{=}1$ vs retrained comparisons are significant at Holm $p < 0.01$
(per-protein cluster bootstrap, $B{=}10{,}000$, family of $12$).}
\label{tab:retrained_rmsf}
\end{table}

\begin{table}[htbp]
\centering\small
\setlength{\tabcolsep}{3.5pt}
\begin{tabular}{l ccc c cc c c cc}
\toprule
 & \multicolumn{3}{c}{EC mAP$\uparrow$} & GO & \multicolumn{2}{c}{Binding} & Aff. & PG & \multicolumn{2}{c}{CryptoBench} \\
\cmidrule(lr){2-4}\cmidrule(lr){5-5}\cmidrule(lr){6-7}\cmidrule(lr){8-8}\cmidrule(lr){9-9}\cmidrule(lr){10-11}
Method & d1 & d2 & d3 & mAP$\uparrow$ & AUC$\uparrow$ & AP$\uparrow$ & Sp.$\uparrow$ & Sp.$\uparrow$ & AUC$\uparrow$ & AP$\uparrow$ \\
\midrule
AA (retr.)   & \underline{0.264} & \underline{0.214} & \textbf{0.159} & \underline{0.347} & 0.699 & \underline{0.208} & 0.409 & 0.337 & \underline{0.619} & 0.131 \\
ESM3 (retr.) & \textbf{0.272} & \textbf{0.215} & \underline{0.147} & \textbf{0.375} & \textbf{0.704} & \textbf{0.212} & \textbf{0.426} & \textbf{0.358} & \underline{0.619} & \underline{0.136} \\
\tool{} $P{=}1$ & \textbf{0.272} & 0.159 & 0.113 & 0.287 & \underline{0.702} & 0.197 & \underline{0.416} & \underline{0.347} & \textbf{0.644} & \textbf{0.165} \\
\bottomrule
\end{tabular}
\caption{Function, fitness and CryptoBench: retrained static tokenizers vs \tool $\lambda{=}10$ ($P{=}1$).
EC and GO: sequence-split macro-AP, mean over the $5$ cross-validation folds ($10$ seeds each), as in the main EC/GO table.
Binding: sequence-split AUROC and AP.
Aff.: sequence-split Spearman (ligand-aware MACCS head).
PG: mean per-assay Spearman ($96$ assays, $L \le 200$, level-0 scoring).
CryptoBench: per-entry AUROC and AP ($222$ apo test entries, MLP probe, $5$ seeds).
Bold: best per column (ties at printed precision both bold); underline: second best.
All three models use the same labels, test proteins and probe protocol.}
\label{tab:retrained_function}
\end{table}

\newpage
\section{Compute Requirements}\label{sec:compute}

All experiments were run on NVIDIA H200 GPUs (one training/eval job per
GPU). Numbers below are wall-clock times we measured during the runs
that produced this paper; per-seed-per-task figures generalize to a
single NVIDIA H200, and ``$N$ GPUs'' figures assume embarrassingly-parallel
scheduling across $N$ devices.

\paragraph{One-time caching.}
Built once, reused across all experiments.
\begin{table}[htbp]
\centering\small
\begin{tabular}{lc}
\toprule
Step & Wall-clock \\
\midrule
Real-backbone cache, mdCATH-div (308k residues, $K{=}10$)         & $\sim$8\,min (32 CPU) \\
Real-backbone cache, MISATO (16{,}972 entries, $K{=}10$)            & $\sim$11\,min (32 CPU) \\
ESM3 $K{=}16$ descriptor cache, combined ($7.1$\,M residues, $P{=}10$) & $\sim$2\,min (1 GPU) \\
Per-tokenizer baseline token cache (mdCATH, BioEmu-CATH or MISATO)       & 3--5\,min (1 GPU) \\
ESMFold atom14 cache for ProteinGym ($\sim$400k variants)     & $\sim$2 GPU-days (4 GPU) \\
\bottomrule
\end{tabular}
\end{table}

The ESMFold cache is the dominant cold-start cost. Once it exists,
all downstream PG scoring is fast (final paragraph).

\paragraph{Tokenizer training.}
The reported \tool checkpoint ($\lambda = 10$, \textsc{RVQVAETokenizer},
$2048{\times}128{\times}128$ codebook, $3.59$\,M parameters,
$256$-D ESM3 $K{=}16$ descriptor input) trains on combined
BioEmu-CATH\,+\,MISATO ($5.73$\,M training residues, $684$\,K validation,
$P{=}10$, varP $1$--$10$) and converges via patience-40 early stopping at
\textbf{epoch 475 (best at epoch 435) in $\sim$24.1\,h on a single H200 GPU}
(batch size 4096, AdamW $10^{-3}$, k-means init, 1000-step warm-up + cosine decay).
The $\lambda$ sweep ($\lambda \in \{0, 0.05, 0.1, 0.25, 0.5, 1, 5, 10, 50, 100, 500\}$ plus SFTD off and VQ-2048, $13$ runs) took $185$ GPU-hours of training time (eight runs shared GPUs in pairs); runs are independent and parallelize across GPUs.

\paragraph{Downstream evaluation (per seed, per split, per task).}
\begin{table}[htbp]
\centering\small
\begin{tabular}{lc}
\toprule
Task & Per-seed wall-clock \\
\midrule
EC depth-$\{$1,2,3$\}$ (one split, at most 40 ep, bs 32)               & 30--90\,s \\
GO top-50 (one split, same trunk as EC)                       & 30--60\,s \\
Binding-site (one split, per-residue head)                    & 30--60\,s \\
Binding-affinity (all 3 splits, MACCS-167 ligand)             & $\sim$40\,s \\
RMSF probe (one split, per-residue MLP, early stopping)                & 20--40\,s \\
ProteinGym tokenize per assay (cache warm)                    & 30\,s\,--\,2\,min \\
ProteinGym scoring (all 96 assays, one tokenizer)             & $\sim$40\,min CPU \\
First-time variant tokenization (16{,}972 MISATO entries)       & 5--10\,min \\
\bottomrule
\end{tabular}
\end{table}

\paragraph{Full reproduction budget.}
Compiling every cell in this paper from a fresh checkout (assuming
the ESMFold and Real-backbone caches are already built):
\begin{itemize}
\item \textbf{Downstream sweep ($10$ seeds, 5 tasks, 3 splits, 8
      tokenizer baselines $+$ \tool at $P{=}\text{full}$ and $P{=}1$).}
      $\sim$1{,}500 jobs averaging $\sim$1\,min each. Sequential
      $\sim$25\,GPU-h; \textbf{$\sim$6.5\,h wall on 4 GPUs}.
\item \textbf{Ligand-aware affinity sweep ($10$ seeds, 3 splits,
      8 baselines $+$ \tool at $P{=}\text{full}$/$P{=}1$,
      all splits in one call).} $\sim$100 jobs at $\sim$40\,s;
      $\sim$15\,min on 4 GPUs.
\item \textbf{ProteinGym scoring (5 tokenizers, 96 assays each).}
      Tokenization and scoring in one GPU pass: $\sim$2.5\,h per tokenizer ($\sim$2.5\,h on 4 GPUs).
\item \textbf{ANOVA suite ($\eta^2$ + 1000-permutation null).}
      $\sim$4 CPU-minutes per tokenizer; negligible.
\item \textbf{Total without ESMFold cache build:} $\sim$80\,GPU-hours,
      achievable in $\sim$10--12\,h wall on a single 8$\times$H200
      node.
\item \textbf{Including ESMFold cache build:} add $\sim$2 GPU-days
      one-time.
\end{itemize}

\paragraph{Memory footprint.}
Peak: tokenizer training $\sim$55\,GB on H200 (varP set encoder $+$
chunked eval of $\sim$684k val residues); conv1d/MLP downstream head
$\sim$2--5\,GB; ESM3 descriptor generation $\sim$1--2\,GB; the largest
model load (AminoAseed, $\sim$140\,M params, full encoder $+$ decoder
$+$ quantizer for PG scoring) $\sim$30\,GB. All experiments fit
on a single 80\,GB device; the reported tokenizer training does
not fit on 40\,GB at batch size 4096 with full-$P$ chunked eval.